\documentclass[10pt,journal,compsoc]{IEEEtran}
\ifCLASSOPTIONcompsoc
  \usepackage[nocompress]{cite}
\else
  \usepackage{cite}
\fi
\ifCLASSINFOpdf
    \usepackage[pdftex]{graphicx}
    \usepackage{tikz}
    \usepackage{tikz-network}
    \usepackage{fontawesome}
    \usepackage{dsfont}
    \usepackage{chronosys}
    \usepackage{stackengine}
    \usepackage{nicefrac}
    \usepackage{tikzit}
    \usepackage[edges]{forest}
    \usepackage{wrapfig}
    \usepackage{lipsum}
    \usetikzlibrary{arrows,calc}
    \usetikzlibrary{calc,patterns,angles,quotes}
    \usetikzlibrary{matrix,chains,decorations.pathreplacing}
    \usetikzlibrary{positioning,fit,arrows.meta,shapes,automata}
    \usetikzlibrary{shapes,arrows, backgrounds, positioning, fit, decorations.pathmorphing}
    
    \tikzstyle{sectionBlock} = [draw, fill=blue!20, rectangle, minimum height=3em, minimum width=3em]
    \tikzstyle{block} = [draw, fill=white, rectangle, minimum height=3em, minimum width=3em]
    \tikzstyle{sum} = [draw, fill=white, circle, node distance=1cm]
    \tikzstyle{input} = [coordinate]
    \tikzstyle{output} = [coordinate]
    \tikzstyle{pinstyle} = [pin edge={to-,thin,black}]
    \tikzstyle{mcCircle} = [draw, circle, line width=0.5mm, minimum width=3em, minimum height=3em, fill=white]
    \tikzstyle{mcInput} = [draw, rectangle, line width=0.5mm, minimum width=3em, minimum height=3em, fill=black!10]
    \tikzstyle{mcCircleX} = [draw=myBlue!80, circle, line width=0.75mm, minimum width=3em, minimum height=3em, fill=white]
    \tikzstyle{mcCircleY} = [draw=myBlue!80, circle, line width=0.75mm, minimum width=3em, minimum height=3em, fill=white]
    
    \tikzset{nnbasic/.style={draw,fill=blue!20,text width=1em,text badly centered}}
    \tikzset{nninput/.style={nnbasic,circle}}
    \tikzset{nnweights/.style={nnbasic,rectangle}}
    \tikzset{nnfunctions/.style={nnbasic,circle,fill=blue!10}}
\else
\fi
\renewcommand{\min}[1]{\underset{#1}{\text{min}}\,}
\renewcommand{\max}[1]{\underset{#1}{\text{max}}\,}
\renewcommand{\inf}[1]{\underset{#1}{\text{inf}}\,}
\renewcommand{\sup}[1]{\underset{#1}{\text{sup}}\,}

\newcommand{\argmin}[1]{\underset{#1}{\text{arg\,min}}\,}
\newcommand{\argmax}[1]{\underset{#1}{\text{arg\,max}}\,}
\newcommand{\arginf}[1]{\underset{#1}{\text{arg\,inf}}\,}

\newcommand{\expectation}[1]{\underset{#1}{\mathbb{E}}}

\newcommand{\iid}[0]{\overset{\text{i.i.d.}}{\sim}}

\stackMath
\newlength\matfield
\newlength\tmplength

\usepackage{amsmath, amsfonts, amsthm}
\usepackage{blkarray}
\usepackage{float}
\usepackage{booktabs}
\usepackage{multirow}
\usepackage{dblfloatfix}
\theoremstyle{definition}

\usepackage{algorithm, algorithmic}
\usepackage{array}
\usepackage{subfig}
\let\MYorigsubfloat\subfloat
\renewcommand{\subfloat}[2][\relax]{\MYorigsubfloat[]{#2}}
\usepackage[acronym]{glossaries}
\newacronym{pde}{PDE}{Partial Density Estimation}

\newacronym{kde}{KDE}{Kernel Density Estimation}
\newacronym{ot}{OT}{Optimal Transport}
\newacronym{rbm}{RBM}{Restricted Boltzmann Machine}
\newacronym{mle}{MLE}{Maximum Likelihood Estimation}

\newacronym{gpgpu}{GPGPU}{General Purpose Graphics Processing Unit}
\newacronym{emd}{EMD}{Earth Mover Distance}
\newacronym{wmd}{WMD}{Word Mover Distance}
\newacronym{swmd}{SWMD}{Supervised Word Mover Distance}
\newacronym{mmd}{MMD}{Maximum Mean Discrepancy}
\newacronym{lot}{LOT}{Linear Optimal Transport}
\newacronym{bow}{BoW}{Bag of Words}
\newacronym{tfidf}{TF-IDF}{Term Frequency-Inverse Term Frequency}
\newacronym{wcd}{WCD}{Word Centroid Distance}
\newacronym{rwmd}{RWMD}{Relaxed Word Mover Distance}
\newacronym{rbf}{RBF}{Radial Basis Function}
\newacronym{ad}{AD}{Automatic Differentiation}
\newacronym{relu}{ReLU}{Rectified Linear Unit}
\newacronym{nn}{NN}{Neural Network}

\newacronym{kl}{KL}{Kullback-Leibler}
\newacronym{js}{JS}{Jensen-Shannon}

\newacronym{rv}{RV}{Random Variable}
\newacronym{drl}{DRL}{Distributional Reinforcement Learning}

\newacronym{po}{PO}{Policy Optimization}

\newacronym{gan}{GAN}{Generative Adversarial Network}
\newacronym{wgan}{WGAN}{Wasserstein GAN}
\newacronym{vae}{VAE}{Variational Autoencoder}
\newacronym{aae}{AAE}{Adversarial Autoencoder}
\newacronym{wae}{WAE}{Wasserstein Autoencoder}
\newacronym{wvae}{WVAE}{Wasserstein Variational Autoencoder}
\newacronym{nf}{NF}{Normalizing Flow}
\newacronym{cnf}{CNF}{Continuous NF}
\newacronym{ode}{ODE}{Ordinary Differential Equation}
\newacronym{node}{NODE}{Neural ODE}
\newacronym{rnode}{RNODE}{Regularized Neural ODE}
\newacronym{ffjord}{FFJORD}{Free Form Jacobian of Reversible Dynamics}

\newacronym{nmf}{NMF}{Non-negative Matrix Factorization}
\newacronym{wnmf}{WNMF}{Wasserstein NMF}

\newacronym{mke}{MKE}{Minimum Kantorovich Estimator}
\newacronym{bcd}{BCD}{Block Coordinate Descent}

\newacronym{tl}{TL}{Transfer Learning}
\newacronym{da}{DA}{Domain Adaptation}
\newacronym{msda}{MSDA}{Multi-Source Domain Adaptation}
\newacronym{hda}{HDA}{Heterogeneous Domain Adaptation}
\newacronym{rl}{RL}{Reinforcement Learning}
\newacronym[longplural={Markov Decision Processes}]{mdp}{MDP}{Markov Decision Process}
\newacronym{otda}{OTDA}{Optimal Transport for Domain Adaptation}
\newacronym{jdot}{JDOT}{Joint Distribution Optimal Transport}
\newacronym{visda}{VisDA}{Visual Domain Adaptation}
\newacronym{uda}{UDA}{Unsupervised Domain Adaptation}
\newacronym{jcpot}{JCPOT}{Joint Class Proportion and Optimal Transport}
\newacronym{wbt}{WBT}{Wasserstein Barycenter Transport}
\newacronym{coot}{COOT}{Co-Optimal Transport}
\newacronym{mlp}{MLP}{Multilayer Perceptron}
\newacronym{dann}{DANN}{Domain Adversarial Neural Network}
\newacronym{wdgrl}{WDGRL}{Wasserstein Distance Guided Representation Learning}
\newacronym{cce}{CCE}{Categorical Cross Entropy}
\newacronym{bce}{BCE}{Binary Cross Entropy}
\newacronym{mse}{MSE}{Mean Squared Error}
\newacronym{elbo}{ELBO}{Evidence Lower Bound}
\newacronym{wdl}{WDL}{Wasserstein Dictionary Learning}
\newacronym{lp}{LP}{Linear Programming}
\newacronym{gw}{GW}{Gromov-Wasserstein}
\newacronym{fgw}{FGW}{Fused Gromov-Wasserstein}
\newacronym{ibp}{IBP}{Iterative Bregman Projections}
\newacronym{dwe}{DWE}{Deep Wasserstein Embedding}
\newacronym{nlp}{NLP}{Natural Language Processing}
\newacronym{sne}{SNE}{Stochastic Neighbor Embeddings}
\newacronym{nca}{NCA}{Neighborhood Component Analysis}
\newacronym{glove}{GloVE}{Global Vectors}
\newacronym{elmo}{ELMo}{Embeddings from Language Models}
\newacronym{auslan}{AUSLAN}{Australian Sign Language}
\newacronym{cstr}{CSTR}{Continuous Stirred Tank Reactor}

\newacronym{dtw}{DTW}{Dynamic Time Warping}
\newacronym{opw}{OPW}{Order-Preserving Wasserstein}
\newacronym{sta}{STA}{Spatio-Temporal Alignment}

\newacronym{knn}{k-NN}{k-Nearest Neighbors}
\newacronym{greenkhorn}{GREENKHORN}{Greedy Sinkhorn}
\newacronym{icnn}{ICNN}{Input Convex Neural Network}
\newacronym{jko}{JKO}{Jordan-Kinderlehrer-Otto}
\newacronym{brl}{BRL}{Bayesian Reinforcement Learning}
\newacronym{wql}{WQL}{Wasserstein Q-Learning}

\newacronym{svm}{SVM}{Support Vector Machine}
\newacronym{cnn}{CNN}{Convolutional Neural Network}
\newacronym{wegl}{WEGL}{Wasserstein Embedding for Graph Learning}
\newacronym{otdd}{OTDD}{Optimal Transport Dataset Distance}
\newacronym{hvs}{HVS}{Human Visual System}
\newacronym{ssim}{SSIM}{Structural Similarity Index}
\newacronym{ipm}{IPM}{Integral Probability Metric}
\newacronym{spk}{SPK}{Shortest-Path Kernel}
\newacronym{pot}{POT}{Python Optimal Transport}
\newacronym{ml}{ML}{Machine Learning}
\newacronym{otml}{OTML}{Optimal Transport for Machine Learning}

\newacronym{erm}{ERM}{Empirical Risk Minimization}
\newacronym{iid}{i.i.d.}{independently and identically distributed}

\newacronym{sw}{SW}{Sliced Wasserstein}
\newacronym{gsw}{GSW}{Generalized Sliced Wasserstein}
\newacronym{srw}{SRW}{Subspace Robust Wasserstein}
\newacronym{rkhs}{RKHS}{Reproducing Kernel Hilbert Space}

\newacronym{mk}{MK}{Monge-Kantorovich}
\newacronym{kr}{KR}{Kantorovich-Rubinstein}
\newacronym{bb}{BB}{Benamou-Brenier}

\newacronym{fda}{FDA}{Fisher Discriminant Analysis}
\newacronym{wda}{WDA}{Wasserstein Discriminant Analysis}

\newacronym{gdl}{GDL}{Graph Dictionary Learning}
\newacronym{gwf}{GWF}{Gromov-Wasserstein Factorization}
\newacronym{otce}{OTCE}{Optimal Transport Conditonal Entropy}
\newacronym{mbot}{MBOT}{Mini-batch OT}

\newacronym{ccot}{CCOT}{Co-Clustering OT}
\newacronym{gmm}{GMM}{Gaussian Mixture Model}
\newacronym{dsw}{DSW}{Distributional SW}
\newacronym{dro}{DRO}{Distributionally Robust Optimization}

\newacronym{hott}{HOTT}{Hierarchical Optimal Transport Topic}

\usepackage[hidelinks]{hyperref}

\begin{document}
\title{Recent Advances in Optimal Transport for Machine Learning}

\author{Eduardo Fernandes~Montesuma,%~\IEEEmembership{Member,~IEEE,}
        Fred Maurice~Ngol\`e Mboula,%~\IEEEmembership{Fellow,~OSA,}
        and~Antoine~Souloumiac,%~\IEEEmembership{Life~Fellow,~IEEE}% <-this % stops a space
\IEEEcompsocitemizethanks{\IEEEcompsocthanksitem The authors are with the Université Paris-Saclay, CEA, LIST, F-91120, Palaiseau, France. %\protect\\
 E-mail: eduardo.fernandesmontesuma@cea.fr
}%\IEEEcompsocthanksitem J. Doe and J. Doe are with Anonymous University.}% <-this % stops an unwanted space
% \thanks{Manuscript received ; revised }
}


% \markboth{IEEE TRANSACTIONS ON PATTERN ANALYSIS AND MACHINE INTELLIGENCE,~Vol.~-1, No.~-1, MONTH YEAR}%
\markboth{Recent Advances in Optimal Transport for Machine Learning}%
{Shell \MakeLowercase{\textit{et al.}}: Bare Demo of IEEEtran.cls for Computer Society Journals}



\IEEEtitleabstractindextext{%
\begin{abstract}
Recently, Optimal Transport has been proposed as a probabilistic framework in Machine Learning for comparing and manipulating probability distributions. This is rooted in its rich history and theory, and has offered new solutions to different problems in machine learning, such as generative modeling and transfer learning. In this survey we explore contributions of Optimal Transport for Machine Learning over the period 2012 -- 2023, focusing on four sub-fields of Machine Learning: supervised, unsupervised, transfer and reinforcement learning. We further highlight the recent development in computational Optimal Transport and its extensions, such as partial, unbalanced, Gromov and Neural Optimal Transport, and its interplay with Machine Learning practice.
\end{abstract}

}

\maketitle

\IEEEdisplaynontitleabstractindextext

% \fi
\IEEEpeerreviewmaketitle

\IEEEraisesectionheading{\section{Introduction}\label{sec:introduction}}

\IEEEPARstart{O}{ptimal} transport is a well-established field of mathematics founded by the works of Gaspard Monge~\cite{monge1781memoire} and Leonid Kantorovich~\cite{kantorovich1942transfer}. Since its genesis, this theory has made significant contributions to science~\cite{villani2009optimal,frisch2002reconstruction,rubner2000earth}. Here, we study how \gls{ot} contributes to different problems within \gls{ml}. \gls{otml} is a growing research subject in the \gls{ml} community. Indeed, \gls{ot} is useful for \gls{ml} through at least two viewpoints: (i) as a loss function and (ii) for manipulating probability distributions.

First, \gls{ot} defines a \emph{metric between distributions}, known by different names, such as Wasserstein distance, Dudley metric, Kantorovich metric, or \gls{emd}. Under certain conditions, this metric belongs to the family of \glspl{ipm} (see section~\ref{sec:background}). In many problems (e.g., generative modeling), the Wasserstein distance is preferable over other notions of dissimilarity between distributions, such as the \gls{kl} divergence, due to its topological, statistical, and geometrical properties. Second, \gls{ot} presents a \emph{toolkit} or framework for \gls{ml} practitioners to manipulate probability distributions. Hence, \gls{ot} is a principled tool to understand the space of probability distributions.

This survey provides an updated view of how \gls{otml} has evolved recently. Even though previous surveys exist~\cite{ambrosio2013user,kolouri2017optimal,solomon2018optimal,levy2018notions,peyre2019computational,flamary2019transport}, the rapid growth of the field justifies a closer look at \gls{otml}. This paper is organized as follows. Section~\ref{sec:background} presents an overview of \gls{ot} theory. Section~\ref{sec:computational_ot} reviews recent developments in \emph{computational optimal transport}. The further sections explore \gls{ot} for 4 \gls{ml} problems: supervised (section~\ref{sec:supervised-learning}), unsupervised (section~\ref{sec:unsupervised-learning}), transfer (section~\ref{sec:transfer-learning}), and reinforcement learning (section~\ref{sec:reinf_learning}). Section~\ref{sec:conclusion} concludes this paper with general remarks and future research directions.
\section{Background}\label{sec:background}

In the following, we present a condensed review of \gls{ot} on $\mathbb{R}^{d}$. For a more detailed overview of \gls{ot} theory, we refer readers to~\cite{santambrogio2015optimal}. The space of probability distributions is denoted by $\mathbb{P}(\mathbb{R}^{d})$. For a mapping $T:\mathbb{R}^{d} \rightarrow \mathbb{R}^{d}$, its associated \emph{pushforward operator}, $T_{\sharp}$, is,
\begin{align}
    (T_{\sharp}P)(A) = P(T^{-1}(A))\text{, for } A \subset \mathbb{R}^{d}.\label{eq:pushfoward}
\end{align}
We denote the set of 1-Lipschitz functions by Lip$_{1}$, and the set of convex functions with finite moments w.r.t. $P \in \mathbb{P}(\mathbb{R}^{d})$ by CVX$(P)$. For $f:\mathbb{R}^{d}\rightarrow\mathbb{R}$, the \emph{convex conjugate} is,
\begin{align}
    f^{*}(\mathbf{x}_{2}) = \sup{\mathbf{x}_{1} \in \mathbb{R}^{d}}\langle \mathbf{x}_{1}, \mathbf{x}_{2} \rangle - f(\mathbf{x}_{1}).\label{eq:convex_conjugate}
\end{align}

Let $P,Q\in\mathbb{P}(\mathbb{R}^{d})$. The Monge formulation~\cite{monge1784memoire} searches for an optimal transport map $T$ such that,
\begin{align}
    \inf{T_{\sharp}P=Q}\mathcal{L}_{M}(T) := \expectation{\mathbf{x}\sim P}[c(\mathbf{x},T(\mathbf{x}))],\label{eq:monge_form}
\end{align}
where $c:\mathbb{R}^{d}\times\mathbb{R}^{d}\rightarrow\mathbb{R}_{+}$ is called \emph{ground-cost}. While $\mathcal{L}_{M}$ defines the \emph{effort of transportation}, $T_{\sharp}P=Q$ specifies \emph{mass conservation}, i.e. $(T_{\sharp}P)(A) = Q(A)$ for $A \subset \mathbb{R}^{d}$. The Monge formulation is notoriously difficult to analyze, partly due to the constraint involving $T_{\sharp}$. A simpler formulation~\cite{kantorovich1942transfer} relies on an \gls{ot} plan $\gamma:\mathbb{R}^{d}\times\mathbb{R}^{d}\rightarrow[0,1]$ such that,
\begin{align}
    \inf{\gamma \in \Gamma(P, Q)}\mathcal{L}_{K}(\gamma) := \expectation{(\mathbf{x}_{1},\mathbf{x}_{2})\sim \gamma}[c(\mathbf{x}_{1},\mathbf{x}_{2})],\label{eq:kantorovich_form}
\end{align}
where $\Gamma(P, Q)$ is the set of \emph{mass preserving plans}, i.e., for $A, B \subset \mathbb{R}^{d}$, $\gamma(\mathbb{R}^{d},B)=Q(B),\text{ and}, \gamma(A,\mathbb{R}^{d})=P(A)$. This formulation is known as \gls{mk}.

The \gls{mk} formulation is easier to analyze because $\Gamma(P,Q)$ and $\mathcal{L}_{K}(\gamma)$ are linear w.r.t. $\gamma$, which characterizes it as a linear program. As such, the \gls{mk} formulation admits a dual problem~\cite[Section 1.2]{santambrogio2015optimal} in terms of \emph{Kantorovich potentials} $\varphi:\mathbb{R}^{d}\rightarrow\mathbb{R}$ and $\psi:\mathbb{R}^{d}\rightarrow\mathbb{R}$,
\begin{align}
    \sup{(\varphi,\psi)\in\Phi_{c}}\mathcal{L}_{K}^{*}(\varphi,\psi) := \expectation{\mathbf{x}\sim P}[\varphi(\mathbf{x})] + \expectation{\mathbf{x} \sim Q}[\psi(\mathbf{x})],\label{eq:dual_kantorovich}
\end{align}
where $\Phi_{c}=\{(\varphi,\psi):\varphi(\mathbf{x}_{1})+\psi(\mathbf{x}_{2})\leq c(\mathbf{x}_{1},\mathbf{x}_{2})\}$. For $c(\mathbf{x}_{1},\mathbf{x}_{2}) = \lVert \mathbf{x}_{1}-\mathbf{x}_{2} \rVert_{2}^{2}$, the celebrated Brenier theorem~\cite{brenier1991polar} establishes a connection between the eqs.~\ref{eq:monge_form} and~\ref{eq:kantorovich_form}: $T = \nabla \varphi$.

Furthermore, \gls{ot} assumes special forms when the ground-cost $c(\mathbf{x}_{1},\mathbf{x}_{2}) = d(\mathbf{x}_{1}, \mathbf{x}_{2})^{p}$, for a metric $d$ on $\mathcal{X}$, and $p \in [1,\infty)$. For $p=1$, one has the \gls{kr} formulation~\cite[Theorem 1.14]{villani2021topics},
\begin{align}
    \sup{\varphi \in \text{Lip}_{1}}\mathcal{L}_{KR}^{*}(\varphi) := \expectation{\mathbf{x} \sim P}[\varphi(\mathbf{x})] - \expectation{\mathbf{x}\sim Q}[\varphi(\mathbf{x})],\label{eq:kantorovich_rubinstein}
\end{align}

In parallel, for $c(\mathbf{x}_{1},\mathbf{x}_{2}) = \lVert \mathbf{x}_{1}-\mathbf{x}_{2} \rVert_{2}^{p}$, $p > 1$, one can consider a time-dependent version of \gls{ot} that reflects how mass is moved from $P$ to $Q$. This is known as dynamic \gls{ot}~\cite[Chapter 6]{santambrogio2015optimal}, and is formulated in terms of a time-dependent distribution $\rho(t,\mathbf{x})$ s.t. $\rho(0,\cdot) = P$ and $\rho(1,\cdot) = Q$, and a vector field $\mathbf{v}$ defining how mass is moved. In these terms, eq.~\ref{eq:kantorovich_form} becomes,
\begin{align}
\mathcal{L}_{B}(\rho, \mathbf{v}) := \int_{0}^{1}\int_{\mathbb{R}^{d}}\lVert \mathbf{v}(t,\mathbf{x}) \rVert^{p}_{2}\rho_{t}(\mathbf{x})d\mathbf{x}dt,\label{eq:benamou_brenier_formulation}
\end{align}
for $\rho_{t} = \rho(t,\cdot)$, under mass conservation constraints,
\begin{align}
    \dfrac{\partial \rho_{t}}{\partial t} + \nabla \cdot (\rho_{t}\mathbf{v}) = 0.\label{eq:continuity_eq}
\end{align}

\begin{figure}[ht]
    \centering
    \subfloat[Case I]{\includegraphics[width=0.27\linewidth]{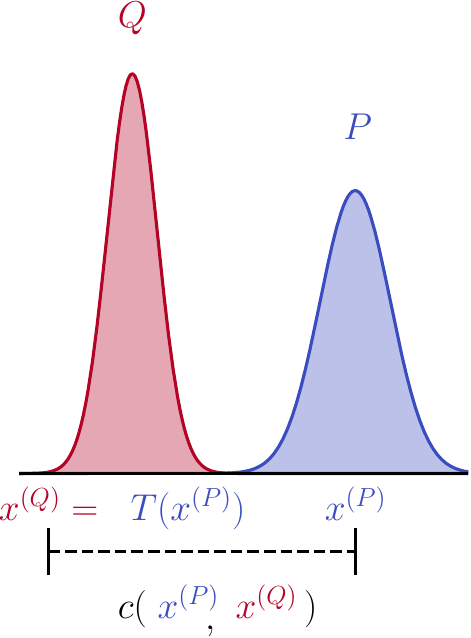}}\hfill
    \subfloat[Case II]{\includegraphics[width=0.33\linewidth]{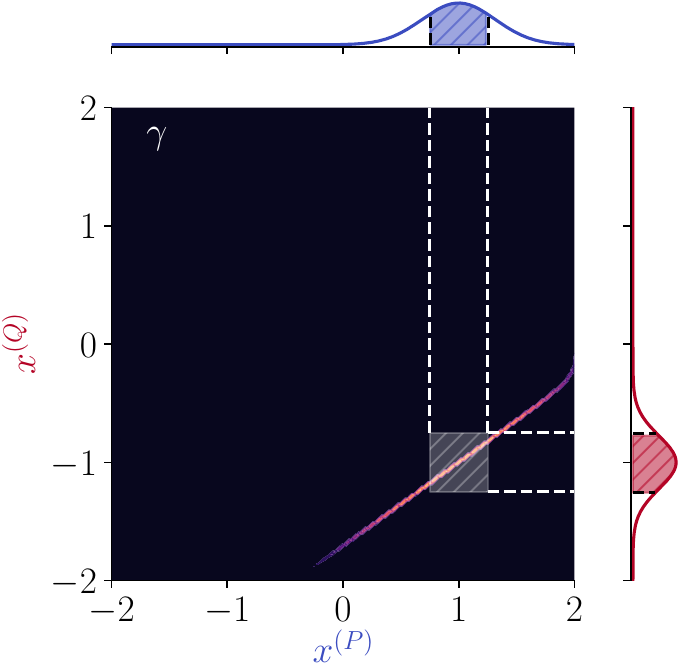}}\hfill
    \subfloat[Case III]{\includegraphics[width=0.33\linewidth]{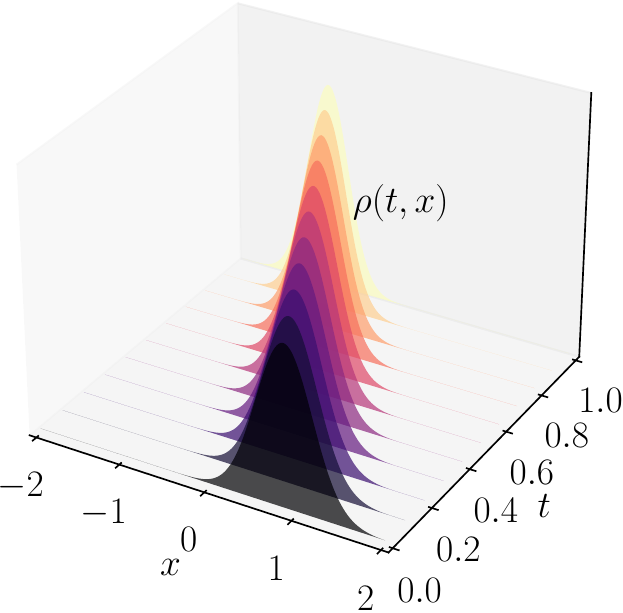}}
    \caption{Illustration of (a) Monge formulation, (b) Kantorovich formulation and (c) Benamou-Brenier formulation. While (a) focuses on transportation maps $T$, (b) relies on transport plans $\gamma$ and (c) revolves around interpolations $\rho(t,x)$.}
    \label{fig:IllustrationOT}
\end{figure}

We show a conceptual comparison of the Monge, Kantorovich and dynamic \gls{ot} formulations in figure~\ref{fig:IllustrationOT} (a), (b) and (c), respectively. Most importantly, due to its many formulations \gls{ot} theory is a quite flexible toolbox for analyzing probabilistic models, hence its popularity.
% The connection~\cite[Prop. 1.1.]{benamou2000computational} is stated through,
% \begin{align}
%    \begin{split}
%        \mathbf{v}(t,\mathbf{x}) = \nabla\phi(t,\mathbf{x})\text{ and }\dfrac{\partial \phi}{\partial t} + \dfrac{1}{2}\lVert \nabla\phi\rVert^{2} = 0.
%    \end{split}\label{eq:optimality_conditions_benamou}
%\end{align}
%where $\phi(1, \cdot) = \psi$ and $\phi(0,\cdot) = \varphi$, i.e., the Kantorovich potentials in eq.~\ref{eq:dual_kantorovich} (see e.g.,~\cite[Section 6.1]{santambrogio2015optimal}).

A central aspect of \gls{ot} theory is that one may define a loss between \textbf{distributions} based on its solutions,
\begin{align*}
    \mathcal{T}_{c}(P, Q) = \inf{\gamma\in\Gamma(P,Q)}\expectation{(\mathbf{x}_{1},\mathbf{x}_{2})\sim \gamma}[c(\mathbf{x}_{1},\mathbf{x}_{2})],
\end{align*}
which is a kind of distance between $P$ and $Q$. As it turns out, when $c$ comes from a metric over $\mathcal{X}$, $\mathcal{T}_{c}$ becomes a metric as well~\cite[Chapter 5]{santambrogio2015optimal}. For $p \in [1,\infty)$, one has the notion of \emph{Wasserstein distances},
\begin{align}
    W_{p}(P, Q) = (\mathcal{T}_{d^{p}}(P, Q))^{1/p},\label{eq:w_dist}
\end{align}
which are widely used in \gls{ml}. This definition has interesting consequences. For instance, the \gls{ot} provides a principled way of interpolating and averaging distributions, through Wasserstein geodesics~\cite{mccann1997convexity} and barycenters~\cite{agueh2011barycenters}.

Concerning geodesics, as shown in~\cite[Chapter 7]{villani2009optimal}, a geodesic between $P$ and $Q$ is a distribution is defined as $P_{t} = \pi_{t,\sharp}\gamma$, where $\pi_{t}(\mathbf{x}_{1},\mathbf{x}_{2}) = (1-t)\mathbf{x}_{1} + t\mathbf{x}_{2}$. Likewise, let $\mathcal{P} = \{P_{i}\}_{i=1}^{N}$, the Wasserstein barycenter~\cite{agueh2011barycenters} of $\mathcal{P}$, weighted by $\alpha \in \Delta_{N} = \{\mathbf{a} \in \mathbb{R}^{N}_{+}:\sum_{i=1}^{N}a_{i}=1\}$ is,
\begin{align}
    \mathcal{B}(\alpha;\mathcal{P}) = \arginf{Q}\sum_{i=1}^{N}\alpha_{i}W_{p}(P_{i},Q)^{p}.\label{eq:wbary}
\end{align}

\glsreset{ipm}

\emph{Probability metrics} are functionals that quantify how different two probability distributions are. These can be either proper metrics (e.g., the Wasserstein distance) or divergences (e.g., the \gls{kl} divergence). In probability theory, there are two prominent families of metrics, \glspl{ipm}~\cite{muller1997integral} and $f-$divergences~\cite{csiszar1967information}. For a family of functions $\mathcal{F}$, an \gls{ipm}~\cite{sriperumbudur2012empirical} is given by,
\begin{align*}
    d_{\mathcal{F}}(P,Q) = \sup{f\in\mathcal{F}}\biggr{|}\expectation{\mathbf{x}\sim P}[f(\mathbf{x})]-\expectation{\mathbf{x}\sim Q}[f(\mathbf{x})]\biggr{|}.
\end{align*}
As such, \glspl{ipm} measure the distance between distributions based on the difference $P - Q$. As a consequence of eq.~\ref{eq:kantorovich_rubinstein}, $W_{1}$ is an \gls{ipm} with $\mathcal{F}=\text{Lip}_{1}$ and $c(\mathbf{x}_{1},\mathbf{x}_{2}) = \lVert \mathbf{x}_{1} - \mathbf{x}_{2} \rVert_{2}$. Another important metric is the \gls{mmd}~\cite{gretton2007kernel}, defined for $\mathcal{F} = \{f\in\mathcal{H}_{k}:\lVert f \rVert_{\mathcal{H}_{k}} \leq 1\}$, where $\mathcal{H}_{k}$ is a \gls{rkhs} with kernel $k:\mathbb{R}^{d}\times\mathbb{R}^{d}\rightarrow\mathbb{R}$. These metrics play an important role in generative modeling and domain adaptation. Conversely, $f-$divergences measure the discrepancies based on the ratio between $P$ and $Q$. For a convex, lower semi-continuous function $f:\mathbb{R}_{+}\rightarrow\mathbb{R}$ with $f(1) = 0$,
\begin{align}
    D_{f}(P||Q) = \expectation{\mathbf{x}\sim Q}\biggr{[}f\biggr{(}\dfrac{P(\mathbf{x})}{Q(\mathbf{x})}\biggr{)}\biggr{]},\label{eq:f_div}
\end{align}
where, with an abuse of notation, $P(\mathbf{x})$ (resp. $Q$) denote the density of $P$. An example of $f-$divergence is the \gls{kl} divergence, with $f(u) = u\log u$.

\begin{figure}[ht]
\centering
\includegraphics[width=\linewidth]{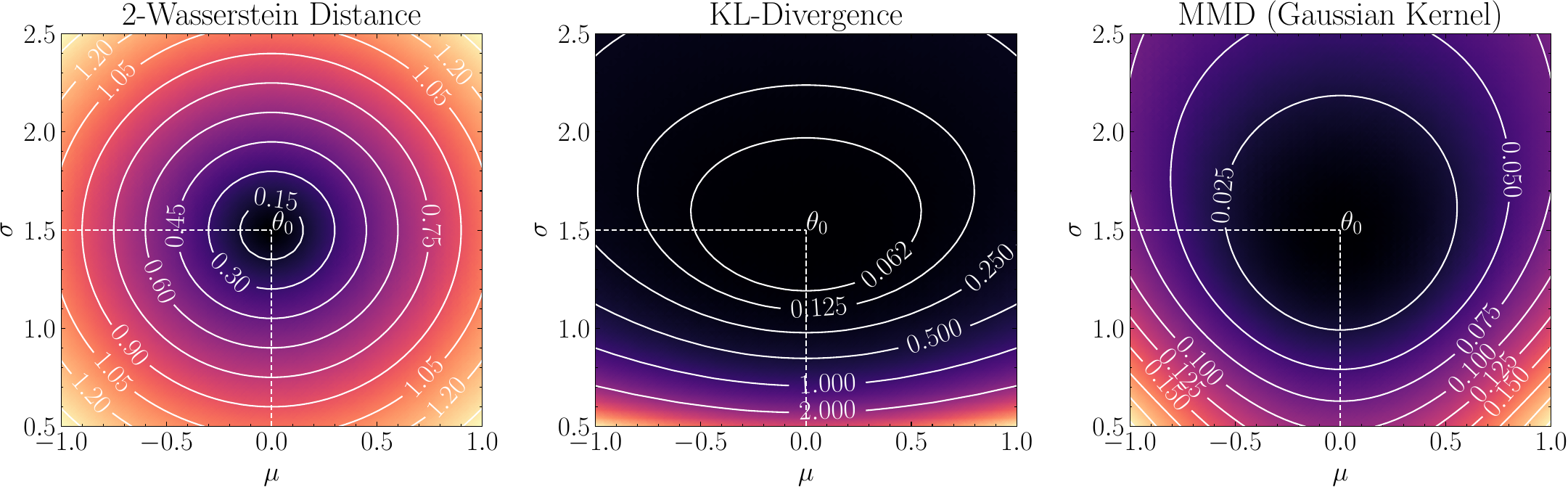}
\caption{Comparison on how different metrics and divergences calculate discrepancies on the manifold of Gaussian distributions. The geometry induced by the Wasserstein distance is simpler, and more intuitive than those given by other measures of discrepancy, which affects optimization procedures in machine learning.}
\label{fig:ipm_gaussian}
\end{figure}

As discussed in~\cite{sriperumbudur2012empirical}, two properties favor \glspl{ipm} over f-divergences. First, $d_{\mathcal{F}}$ is defined even when $P$ and $Q$ have disjoint supports. For instance, at the beginning of training, \glspl{gan} generate poor samples, so $P_{model}$ and $P_{data}$ have disjoint support. In this sense, \glspl{ipm} provide a meaningful metric, whereas $D_{f}(P_{model}||P_{data}) = +\infty$ irrespective of how bad $P_{model}$ is. Second, \glspl{ipm} account for the geometry of the space where the samples live. As an example, consider the manifold of Gaussian distributions $\mathcal{M} = \{\mathcal{N}(\mu,\sigma^{2}):\mu\in\mathbb{R},\sigma\in\mathbb{R}_{+}\}$. As discussed in~\cite[Remark 8.2]{peyre2019computational}, When restricted to $\mathcal{M}$, the Wasserstein distance with a Euclidean ground-cost is,
\begin{align*}
    W_{2}(P,Q) = \sqrt{(\mu_{P}-\mu_{Q})^{2} + (\sigma_{P}-\sigma_{Q})^{2}},
\end{align*}
whereas the \gls{kl} divergence is associated with an hyperbolic geometry. This is shown in Figure~\ref{fig:ipm_gaussian}. Overall, the choice of discrepancy between distributions heavily influences the success of learning algorithms (e.g., \glspl{gan}). Indeed, each choice of metric/divergence induces a different geometry in the space of probability distributions, thus changing the underlying optimization problems in \gls{ml}.

\section{Computational Optimal Transport}\label{sec:computational_ot}

Computational \gls{ot} is an active field of research within \gls{ml}. We refer readers to~\cite{levy2018notions,peyre2019computational} for its foundations, and to~\cite{flamary2021pot,cuturi2022optimal} for widely used software. In this survey, we focus on two discretization strategies: (i) discretizing the ambient space; (ii) approximating the distributions from samples. In both cases, let $\mathbf{x}_{i}^{(P)} \sim P$ with probability $p_{i} > 0$. The \emph{empirical approximation} $\hat{P}$ of $P$ is,
\begin{align}
    \hat{P}(\mathbf{x}) = \sum_{i=1}^{n}p_{i}\delta(\mathbf{x}-\mathbf{x}_{i}^{(P)}).\label{eq:empirical_approx}
\end{align}
Naturally, $\sum_{i=1}^{n}p_{i}=1$ or $\mathbf{p} \in \Delta_{n}$ in short. As follows, discretizing the ambient space is equivalent to binning it, thus assuming a fixed grid of points $\mathbf{x}_{i}^{(P)}$. The sample weights $p_{i}$ correspond to the number of points that are assigned to the $i$-th bin. In this sense, the parameters of $\hat{P}$ are the weights $p_{i}$. Conversely, one can sample $\mathbf{x}_{i}^{(P)} \iid P$ (resp. Q). In this case, $p_{i} = \nicefrac{1}{n}$, and the parameters of $\hat{P}$ are the locations $\mathbf{x}_{i}^{(P)}$. We now discuss discrete \gls{ot}.

Let $\{\mathbf{x}_{i}^{(P)}\}_{i=1}^{n}$ (resp. $\{\mathbf{x}_{j}^{(Q)}\}_{j=1}^{m}$) sampled from $P$ (resp. $Q$) with probability $p_{i}$ (resp. $q_{j}$). The Monge problem seeks a mapping $T$, that is the solution of,
\begin{equation}
T^{\star} = \argmin{T_{\sharp}\hat{P}=\hat{Q}} \mathcal{L}_{M}(T) = \sum_{i=1}^{n}c(\mathbf{x}_{i}^{(P)},T(\mathbf{x}_{i}^{(P)})),\label{eq:MongeProblem}
\end{equation}
where the constraint implies $\sum_{i\in\mathcal{I}}p_{i}=q_{j}$, for $\mathcal{I}=\{i:\mathbf{x}_{j}^{(Q)}=T(\mathbf{x}_{i}^{(P)})\}$. This formulation is non-linear w.r.t. $T$. In addition, for $m > n$, it does not have a solution. Conversely, the \gls{mk} formulation seeks an \gls{ot} \emph{plan} $\gamma \in \mathbb{R}^{n\times m}$, where $\gamma_{ij}$ denotes the amount of mass transported from sample $i$ to sample $j$. In this case $\gamma$ must minimize,
\begin{equation}
\hat{\gamma} = \argmin{\gamma\in \Gamma(\mathbf{p},\mathbf{q})} \mathcal{L}_{K}(\gamma) = \sum_{i=1}^{n}\sum_{j=1}^{m}\gamma_{ij}c(\mathbf{x}_{i}^{(P)},\mathbf{x}_{j}^{(Q)}),\label{eq:KantorovichProblem}
\end{equation}
where $\Gamma(\mathbf{p},\mathbf{q})=\{\gamma \in \mathbb{R}^{n\times m}:\sum_{i}\gamma_{ij}=q_{j}\text{ and }\sum_{j}\gamma_{ij}=p_{i}\}$. This is a linear program, which can be solved through the Simplex algorithm~\cite{dantzig1983reminiscences}, with time complexity $\mathcal{O}(n^{3}\log n)$. Alternatively, one can use the approximation introduced by~\cite{cuturi2013sinkhorn}, by solving a regularized problem,
\begin{equation}
\hat{\gamma}_{\epsilon} = \argmin{\gamma\in \Gamma(\mathbf{p},\mathbf{q})} \sum_{i=1}^{n}\sum_{j=1}^{m}\gamma_{ij}c(\mathbf{x}_{i}^{(P)},\mathbf{x}_{j}^{(Q)}) + \epsilon H(\gamma),\label{eq:sinkhorn}
\end{equation}
which provides a faster way to estimate $\gamma$. An additional advantage of the Sinkhorn algorithm is that
\begin{equation}
    \hat{\gamma}_{\epsilon} = \text{diag}(\mathbf{f})e^{-\mathbf{C}/\epsilon}\text{diag}(\mathbf{g}),\label{eq:sinkhorn_optimality}
\end{equation}
where, as in eq.~\ref{eq:dual_kantorovich}, $(\mathbf{f},\mathbf{g})$ are the Kantorovich potentials.

Solving \gls{ot} with finite samples provides an empirical estimator for $\mathcal{T}_{c}$ and $W_{p}$, i.e., $\mathcal{T}_{c}(\hat{P}, \hat{Q}) = \mathcal{L}_{K}(\hat{\gamma})$. Likewise, for $\gamma^{\star}_{\epsilon}$ one has $\mathcal{T}_{c,\epsilon}(\hat{P},\hat{Q})=\mathcal{L}_{K}(\gamma_{\epsilon}^{\star})$. This approximation motivates the Sinkhorn divergence~\cite{genevay2018learning},
\begin{align}
    \resizebox{0.85\linewidth}{!}{$S_{p,\epsilon}(\hat{P},\hat{Q}) = W_{p,\epsilon}(\hat{P},\hat{Q}) - \dfrac{W_{p,\epsilon}(\hat{P},\hat{P}) + W_{p,\epsilon}(\hat{Q},\hat{Q})}{2}$},\label{eq:sinkhorn_loss}
\end{align}
which has interesting properties, such as interpolating between the \gls{mmd} of~\cite{gretton2007kernel} and the Wasserstein distance. Overall, entropic \gls{ot} has two computational advantages w.r.t exact \gls{ot}. Indeed, its calculations are GPU-friendly, and for $L \geq 1$ iterations, its complexity is $\mathcal{O}(Ln^{2})$. In addition, $S_{c,\epsilon}$ is a smooth approximator of $W_{p}$~\cite{luise2018differential}, and it enjoys better sample complexity~\cite{genevay2019sample} (c.f., section~\ref{sec:challenges}).

In the following, we discuss recent innovations on computational \gls{ot}. Section~\ref{sec:projbased} present projection-based methods. Section~\ref{sec:structured_ot} discusses \gls{ot} formulations with prescribed structures. Section~\ref{sec:icnn_ot} presents \gls{ot} through \glspl{icnn}. Section~\ref{sec:minibatch_ot} explores how to compute \gls{ot} between mini-batches of data.

\subsection{Projection-based Optimal Transport}\label{sec:projbased}

\begin{figure}[ht]
    \centering
    \subfloat[Case I]{\includegraphics[width=0.3\linewidth]{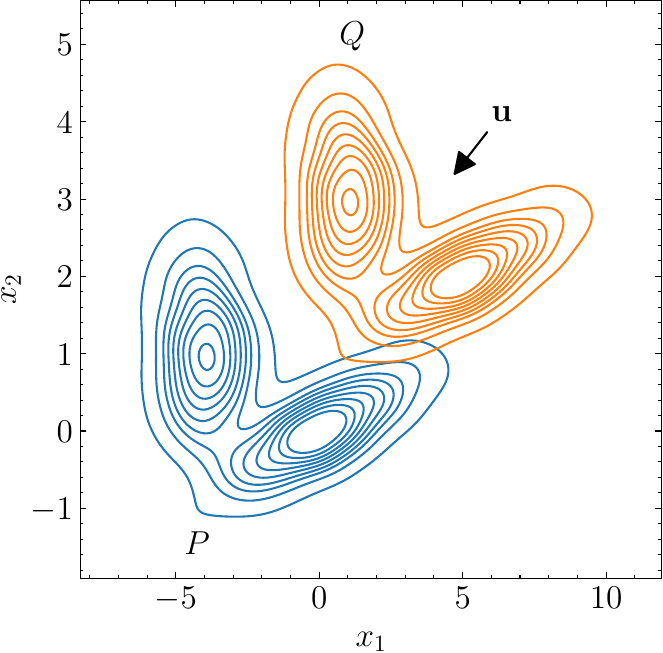}}\hfill
    \subfloat[Case II]{\includegraphics[width=0.3\linewidth]{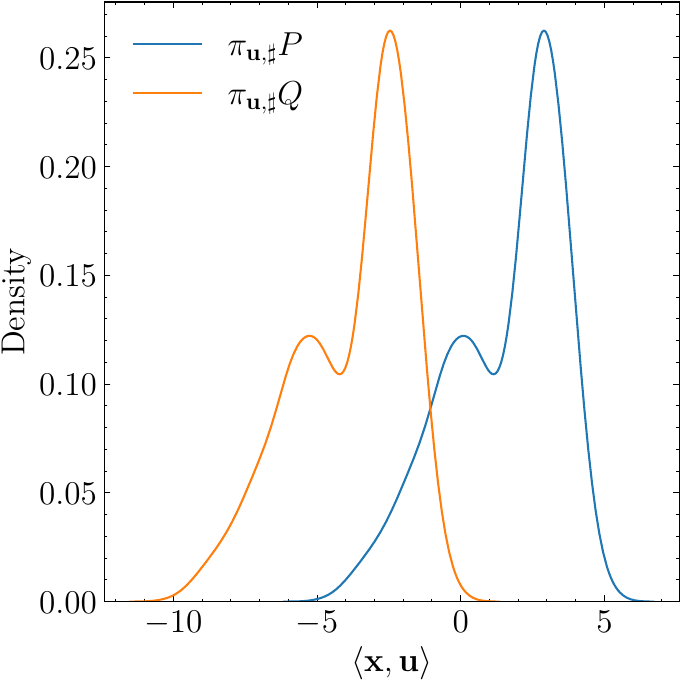}}\hfill
    \subfloat[Case III]{\includegraphics[width=0.3\linewidth]{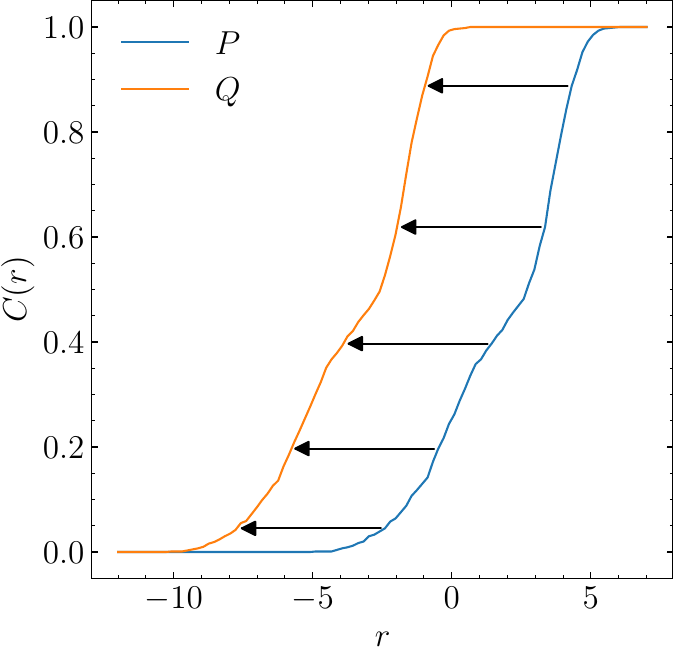}}\\
    \subfloat[Case IV]{\includegraphics[width=0.3\linewidth]{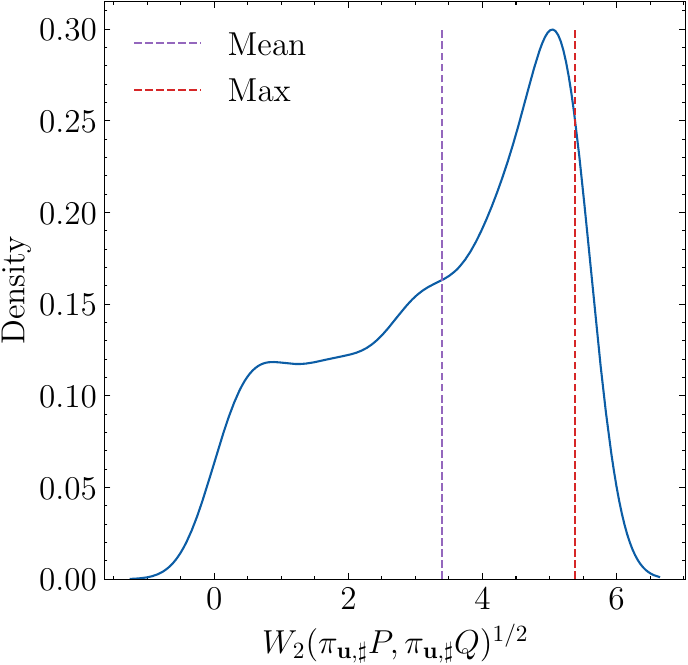}}\hfill
    \subfloat[Case V]{\includegraphics[width=0.3\linewidth]{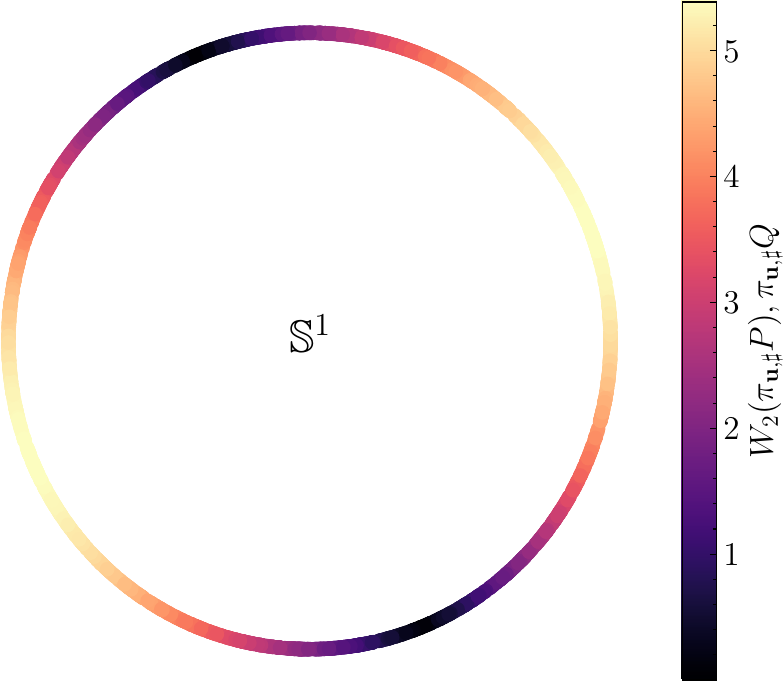}}\hfill
    \subfloat[Case VI]{\includegraphics[width=0.3\linewidth]{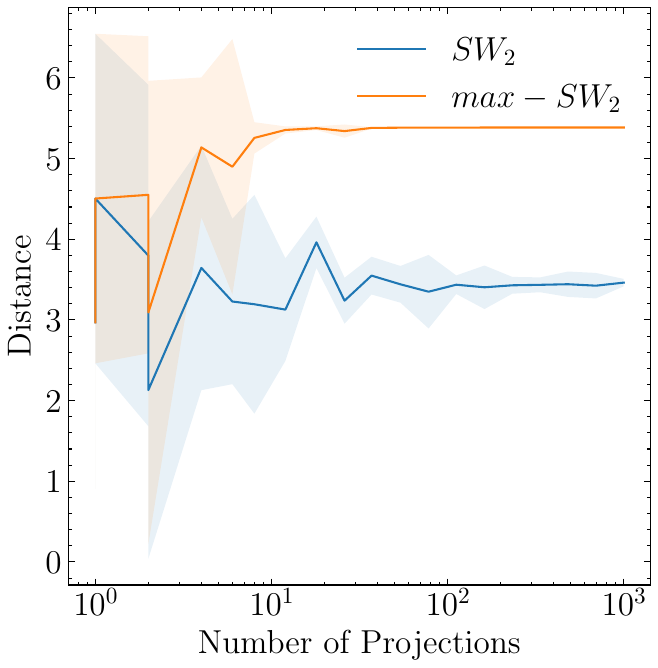}}
    \caption{An illustration of the sliced and max-sliced Wasserstein distances over 2-D distributions (a). In (b), we show the densities of $P$ and $Q$ after a projection by $\mathbf{u}$. In (c), we illustrate the computation of the 1-D Wasserstein distance for $p=1$, as the horizontal difference between the cumulative distributions of $P$ and $Q$. In (d), we show the distribution of the Wasserstein distance over $\mathbf{u} \sim \mathbb{S}^{1}$, alongside the mean (purple) and max (red) values. In (e), we show the Wasserstein distances over $\mathbf{u} \in \mathbb{S}^{1}$. Finally, (f) shows the estimation of the $\text{SW}_{2}$ and max-SW$_{2}$ as a function of the number of projections $L$. Shaded regions show a 95\% confidence interval around the average value.}
    \label{fig:sw2_example}
\end{figure}

Projection-based \gls{ot} relies on projecting data $\mathbf{x} \in \mathbb{R}^{d}$ into sub-spaces $\mathbb{R}^{k}$, $k < d$. A natural choice is $k = 1$, for which computing \gls{ot} can be done by sorting~\cite[chapter 2]{santambrogio2015optimal}. This is called \gls{sw} distance~\cite{rabin2011wasserstein,bonneel2015sliced}. Let $\mathbb{S}^{d-1} = \{\mathbf{u} \in \mathbb{R}^{d}: \lVert \mathbf{u} \rVert_{2}^{2} = 1\}$ denote the unit-sphere in $\mathbb{R}^{d}$, and $\pi_{\mathbf{u}}:\mathbb{R}^{d}\rightarrow\mathbb{R}$ denote $\pi_{\mathbf{u}}(\mathbf{x}) = \langle \mathbf{u}, \mathbf{x} \rangle$. The Sliced-Wasserstein distance is,
\begin{align}
    \text{SW}_{p}(P,Q)^{p} = \int_{\mathbb{S}^{d-1}}W_{p}^{p}(\pi_{\mathbf{u},\sharp}\hat{P}, \pi_{\mathbf{u},\sharp}\hat{Q})d\mathbf{u}.\label{eq:sliced_wasserstein}
\end{align}
We highlight a few advantages. First, $W_{p}(\pi_{\mathbf{u},\sharp}P, \pi_{\mathbf{u},\sharp}Q)$ can be computed in $\mathcal{O}(n \log n)$~\cite{peyre2019computational}. Second, the integration in equation~\ref{eq:sliced_wasserstein} can be computed using Monte Carlo estimation. For samples $\{\mathbf{u}_{\ell}\}_{\ell=1}^{L}$, $\mathbf{u}_{\ell} \in \mathbb{S}^{d-1}$ uniformly,
\begin{align}
    \text{SW}_{p}(\hat{P},\hat{Q})^{p} = \dfrac{1}{L}\sum_{\ell=1}^{L}W_{p}^{p}(\pi_{\mathbf{u}_{\ell},\sharp}\hat{P}, \pi_{\mathbf{u}_{\ell},\sharp}\hat{Q}),
\end{align}
which implies that $\text{SW}(P,Q)$ has $\mathcal{O}(Lnd + Ln\log n)$ time complexity. As shown in~\cite{kolouri2015radon}, $SW(P,Q)$ is indeed a metric. In addition, \cite{deshpande2019max} and~\cite{kolouri2019generalized} proposed variants of $SW$, namely, the max-\gls{sw} distance and the generalized \gls{sw} distance respectively. Contrary to the averaging procedure in eq.~\ref{eq:sliced_wasserstein}, the max-\gls{sw} of~\cite{deshpande2019max} takes the direction with maximum distance between $P$ and $Q$,
\begin{align}
    \text{max-SW}_{p}^{p}(P, Q) = \max{\mathbf{u} \in \mathbb{S}^{d-1}}W_{p}^{p}(\pi_{\mathbf{u,\sharp}}P, \pi_{\mathbf{u,\sharp}}Q).\label{eq:max_SW}
\end{align}
This metric has the same advantage in sample complexity as the \gls{sw} distance, while being easier to compute. We illustrate these concepts in Figure~\ref{fig:sw2_example} for $P, Q \in \mathbb{P}(\mathbb{R}^{2})$.

With respect to figure~\ref{fig:sw2_example}, note that projection directions are not equally important. This is illustrated in~\ref{fig:sw2_example} (d), in which some directions have higher distance than others. This phenomenon was analyzed by~\cite{nguyen2021distributional}, who proposed the so-called \gls{dsw} distance,
\begin{align*}
    \text{DSW}_{p}(P,Q;C) &= \sup{\sigma \in \mathbb{M}_{C}}\biggr(\expectation{\mathbf{u} \sim \sigma}[W_{p}^{p}(\pi_{\mathbf{u},\sharp}P,\pi_{\mathbf{u},\sharp}Q)]\biggr)^{1/p},
\end{align*}
for a family $\mathbb{M}_{C}=\{\sigma \in \mathbb{P}(\mathbb{S}^{d-1}): \expectation{\mathbf{u},\mathbf{u}'\sim \sigma}[|\mathbf{u}^{T}\mathbf{u}'|] \leq C\}$. The intuition behind this metric is that $\sigma$ weights the directions sampled from $\mathbb{S}^{d-1}$. 

In addition, one can project samples on a sub-space $1 < k < d$. For instance, \cite{paty2019subspace} proposed the \gls{srw} distances,
\begin{align*}
    \text{SRW}_{k}(P,Q)^{2} = \inf{\gamma\in\Gamma(P,Q)}\sup{E\in \mathcal{G}_{k}} \expectation{(\mathbf{x}_{1},\mathbf{x}_{2}) \sim \gamma}[\lVert \pi_{E}(\mathbf{x}_{1}-\mathbf{x}_{2}) \rVert^{2}_{2}],
\end{align*}
where $\mathcal{G}_{k} = \{E\subset\mathbb{R}^{d}:dim(E) = k\}$ is the Grassmannian manifold of $k-$dimensional subspaces of $\mathbb{R}^{d}$, and $\pi_{E}$ denote the orthogonal projector onto $E$. This can be equivalently formulated through a projection matrix $\mathbf{U} \in \mathbb{R}^{k \times d}$, that is,
\begin{align*}
    \text{SRW}_{k}(P, Q)^{2} = \inf{\Gamma(P,Q)}\max{\substack{\mathbf{U}\in\mathbb{R}^{k\times d}\\\mathbf{U}\mathbf{U}^{T}=\mathbf{I}_{k}}}\expectation{(\mathbf{x}_{1},\mathbf{x}_{2})\sim\gamma}[\lVert \mathbf{Ux}_{1}-\mathbf{Ux}_{2} \rVert_{2}^{2}]
\end{align*}
In practice, SRW$_{k}$ is based on a projected super gradient method~\cite[Algorithm 1]{paty2019subspace} that updates $\Omega = \mathbf{UU}^{T}$, for a fixed $\gamma$, until convergence. These updates are computationally complex, as they rely on eigendecomposition. Further developments using Riemannian~\cite{lin2020projection} and \gls{bcd}~\cite{huang2021riemannian} circumvent this issue by optimizing over $\text{St}(d,k) = \{\mathbf{U}\in\mathbb{R}^{k\times d}:\mathbf{U}\mathbf{U}^{T} = \mathbf{I}_{k}\}$.

\subsection{Structured Optimal Transport}\label{sec:structured_ot}

In some cases, it is desirable for the \gls{ot} plan to have additional structure, e.g., in color transfer~\cite{ferradans2014regularized} and domain adaptation~\cite{courty2017otda}. Based on this problem,~\cite{alvarez2018structured} introduced a principled way to compute structured \gls{ot} through sub-modular costs.

As defined in~\cite{alvarez2018structured}, a set function $F:2^{V} \rightarrow \mathbb{R}$ is sub-modular if $\forall S \subset T \subset V$ and $\forall v \in V \text{\textbackslash} T$,
\begin{align*}
    F(S \cup \{v\}) - F(S) \geq F(T \cup \{v\}) - F(T).
\end{align*}

These types of functions arise in combinatorial optimization, and further \gls{ot} since \gls{ot} mappings and plans can be seen as a matching between 2 sets, namely, samples from $P$ and $Q$. In addition, $F$ defines a \emph{base polytope},
\begin{align*}
    \mathcal{B}_{F} = \{G\in\mathbb{R}^{|V|}:G(V)=F(V);G(S) \leq F(S)\text{ }\forall S \subset V\}.
\end{align*}

Based on these concepts, note that \gls{ot} can be formulated in terms of set functions. Indeed, suppose $\mathbf{X}^{(P)} \in \mathbb{R}^{n\times d}$ and $\mathbf{X}^{(Q)} \in \mathbb{R}^{m\times d}$. In this case, $\gamma^{\star}$ or $T^{\star}$ can be interpreted as a graph with edge set $E = \{(u_{\ell},v_{\ell})\}_{\ell=1}^{k}$, where $u_{\ell}$ represents a sample in $P$ and $v_{\ell}$, the corresponding sample (through $\gamma$) in $Q$. In this case, the cost of transportation is represented by $F(S) = \sum_{(u,v)\in S}c_{uv}$. Hence, adding a new $(u,v)$ to $S$ is the same regardless of the elements of $S$. 

The insight of~\cite{alvarez2018structured} is using the sub-modular property for acquiring structured \gls{ot} plans. Through Lovász extension~\cite{lovasz1983mathematical}, this leads to,
\begin{align*}
    (\gamma^{\star},\kappa^{\star}) = \argmin{\gamma\in \Gamma(\mathbf{p},\mathbf{q})}\argmax{\kappa\in\mathcal{B}_{F}}\langle \gamma, \kappa \rangle_{F}.
\end{align*}

A similar direction was explored by~\cite{forrow2019statistical}, who proposed to impose a low rank structure on \gls{ot} plans. This was done with the purpose of tackling the curse of dimensionality in \gls{ot}. They introduce the transport rank of $\gamma \in \Gamma(P,Q)$ defined as the smallest integer $K$ for which,
\begin{align*}
    \gamma = \sum_{k=1}^{K}\lambda_{k}(P_{k}\otimes Q_{k}),
\end{align*}
where $P_{k},Q_{k}, k=1,\cdots,K$ are distributions over $\mathbb{R}^{d}$, and $P_{k} \otimes Q_{k}$ denotes the (independent) joint distribution with marginals $P_{k}$ and $Q_{k}$, i.e. $(P_{k} \otimes Q_{k})(\mathbf{x},\mathbf{y}) = P_{k}(\mathbf{x})Q_{k}(\mathbf{y})$. For empirical $\hat{P}$ and $\hat{Q}$, $K$ coincides with the non-negative rank~\cite{cohen1993nonnegative} of $\gamma \in \mathbb{R}^{n\times m}$. As follows,~\cite{forrow2019statistical} denotes the set of $\gamma$ with transport rank at most $K$ as $\Gamma_{K}(P,Q)$ ($\Gamma_{K}(\mathbf{p},\mathbf{q})$ for empirical $\hat{P}$ and $\hat{Q}$). The robust Wasserstein distance is thus,
\begin{align*}
    \text{FW}_{K,2}(P,Q)^{2} = \inf{\gamma\in\Gamma_{K}(P,Q)}\expectation{(\mathbf{x}_{1},\mathbf{x})\sim\gamma}[\lVert \mathbf{x}_{1}-\mathbf{x}_{2} \rVert_{2}^{2}].
\end{align*}
In practice, this optimization problem is difficult due to the constraint $\gamma \in \Gamma_{K}$. As follows, \cite{forrow2019statistical} propose estimating it using the Wasserstein barycenter $B = \mathcal{B}([\nicefrac{1}{2},\nicefrac{1}{2}];\{P, Q\})$ supported on $K$ points, that is $\{\mathbf{x}_{k}^{(B)}\}_{k=1}^{K}$, also called \emph{hubs}. As follows, the authors show how to construct a transport plan in $\Gamma_{k}(P,Q)$, by exploiting the transport plans $\gamma_{1} \in \Gamma(P,B)$ and $\gamma_{2} \in \Gamma(B, Q)$. This establishes a link between the robustness of the Wasserstein distance, Wasserstein barycenters and clustering. Let $\lambda_{k} = \sum_{i=1}^{n}\gamma_{ki}^{(BP)}$ and $\mu_{k}^{(P)} = \nicefrac{1}{\lambda_{k}}\sum_{i=1}^{n}\gamma_{ki}^{(BP)}\mathbf{x}_{i}^{(P)}$, the authors in~\cite{forrow2019statistical} propose the following proxy for the Wasserstein distance,
 \begin{align*}
    \text{FW}_{k,2}^{2}(\hat{P},\hat{Q})^{2} &= \sum_{k=1}^{K}\lambda_{k}\lVert \mu_{k}^{(P)}-\mu_{k}^{(Q)} \rVert_{2}.
\end{align*}

\subsection{Neural Network-based Solvers}\label{sec:icnn_ot}

In \gls{ml}, different works estimate \gls{ot} through \glspl{nn}~\cite{arjovsky2017wasserstein,seguy2018large,makkuva2020optimal,korotin2023neural}. For instance, as we cover in section~\ref{sec:gm},~\cite{arjovsky2017wasserstein} proposes to estimate the Kantorovich-Rubinstein distance in eq.~\ref{eq:kantorovich_rubinstein} by parametrizing $\varphi$ through a \gls{nn}. In the following we cover how different works approximate \gls{ot} plans and maps through \glspl{nn}.

\noindent\textbf{Neural OT Plans.} ~\cite{seguy2018large} was the first to propose to approximate \gls{ot} plans through \glspl{nn}. Their approach relies on solving the entropic regularized dual Kantorovich problem in eq.~\ref{eq:dual_kantorovich}, by parametrizing $\varphi$ and $\psi$ through \glspl{nn} $(u_{\xi}, v_{\eta})$. The optimization procedure consists on maximizing,
\begin{align*}
    \sup{\xi, \eta} \expectation{\mathbf{x}_{1}\sim P, \mathbf{x}_{2}\sim Q}[u_{\xi}(\mathbf{x}_{1})+v_{\eta}(\mathbf{x}_{2})-\epsilon \gamma_{\epsilon}(\mathbf{x}_{1},\mathbf{x}_{2})],
\end{align*}
where, in analogy with eq.~\ref{eq:sinkhorn_optimality},
\begin{align*}
    \gamma_{\epsilon}(\mathbf{x}_{1},\mathbf{x}_{2}) = e^{-(u_{\xi}(\mathbf{x}_{1})+v_{\eta}(\mathbf{x}_{2})-c(\mathbf{x}_{1},\mathbf{x}_{2}))/\epsilon}.
\end{align*}
As a result, for optimal $(\xi, \eta)$, one can estimate the \gls{ot} plan for pairs $\mathbf{x}_{1} \sim P$ and $\mathbf{x}_{2} \sim Q$.

\noindent\textbf{Neural OT Maps.} As we discussed in the previous topic,~\cite{seguy2018large} proposed a way to approximate entropic regularized Kantorovich potentials. Based on this optimization procedure, the authors propose approximating the so-called barycentric mapping through a \gls{nn} $f_{\theta}(\mathbf{x})$,
\begin{align*}
    \min{\theta} \expectation{\mathbf{x}_{1} \sim P,\mathbf{x}_{2}\sim Q}[d(f_{\theta}(\mathbf{x}_{1}),\mathbf{x}_{2})\gamma_{\epsilon}(\mathbf{x}_{1},\mathbf{x}_{2})],
\end{align*}

\begin{figure}[ht]
    \centering
    \resizebox{\linewidth}{!}{
	\begin{tikzpicture}[auto, node distance=1cm,>=latex']
	\node[rectangle,thick,draw,minimum height=1cm,minimum width=.3cm] (x) at (0, 0) {$\mathbf{x}$};
	\node[rectangle,thick,draw,minimum height=1cm,minimum width=.3cm] (z1) at (2, 0) {$\sigma_{1}$};
	\node[rectangle,thick,draw,minimum height=1cm,minimum width=.3cm] (z2) at (4, 0) {$\sigma_{2}$};
	\node[rectangle,] (dots) at (6, 0) {$\cdots$};
	\node[rectangle,thick,draw,minimum height=1cm,minimum width=.3cm] (zk) at (8, 0) {$\sigma_{L-1}$};
	\node[rectangle] (output) at (10, 0) {$f(\mathbf{x};\theta)$};
	
	\draw[->, thick] (x.east) -- (z1.west) node[pos=0.5] {$\mathbf{A}_{0}$};
	\draw[->, thick] (z1.east) -- (z2.west) node[pos=0.5] {$\mathbf{W}_{1}$};
	\draw[->, thick] (z2.east) -- (dots.west) node[pos=0.5] {$\mathbf{W}_{2}$};
	\draw[->, thick] (dots.east) -- (zk.west) node[pos=0.5] {$\mathbf{W}_{L-1}$};
	\draw[->, thick] (x.south) -- (0, -2) -- (4, -2) -- (z2.south) node[pos=0.5] {$\mathbf{A}_{1}$};
	\draw[->, thick] (4, -2) -- (6, -2) -- (dots.south) node[pos=0.5] {};
	\draw[->, thick] (6, -2) -- (8, -2) -- (zk.south) node[pos=0.5] {$\mathbf{A}_{L-1}$};
	\draw[->, thick] (zk.east) -- (output.west);
	\end{tikzpicture}}
    \caption{\gls{icnn} architecture proposed by~\cite{amos2017input}, which implements a convex function $f(\mathbf{x};\theta)$ with respect inputs $\mathbf{x}$.}
    \label{fig:icnn}
\end{figure}

A further development, when $c(\mathbf{x}_{1},\mathbf{x}_{2}) = \lVert \mathbf{x}_{1} - \mathbf{x}_{2} \rVert_{2}^{2}$, was proposed by~\cite{makkuva2020optimal}, who relies on convex analysis~\cite{villani2021topics} and \glspl{icnn}~\cite{amos2017input}. Formally, an \gls{icnn} implements a function $f:\mathbb{R}^{d}\rightarrow\mathbb{R}$ such that $\mathbf{x} \mapsto f_{\theta}(\mathbf{x})$ is convex. This is achieved through a special structure, shown in figure~\ref{fig:icnn}. An $L-$layer \gls{icnn} is defined through the operations,
\begin{align*}
   \mathbf{z}_{\ell+1} = \sigma_{\ell}(\mathbf{W}_{\ell}\mathbf{z}_{\ell}+\mathbf{A}_{\ell}\mathbf{x}+\mathbf{b}_{\ell})\text{ and }f_{\theta}(\mathbf{x}) = \mathbf{z}_{L},
\end{align*}
where all entries of $\mathbf{W}_{\ell}$ are non-negative, $\sigma_{0}$ is convex and $\sigma_{\ell}, \ell=1,\cdots,L-1$ is convex and non-decreasing. The \gls{icnn} architecture is shown in Figure~\ref{fig:icnn}.

The insight from~\cite{makkuva2020optimal} comes from rewriting eq.~\ref{eq:dual_kantorovich} as a mini-max problem involving two convex functions $f_{\theta} \in \text{CVX}(P)$ and $g_{\eta} \in \text{CVX}(Q)$,
\begin{align}
     W_{2}(P,Q)^{2} &= \sup{\theta}\inf{\eta}\mathcal{V}(\theta,\eta)+\mathcal{E},\label{eq:makkuva}\\
     \mathcal{V}(\theta, \eta) &= -\expectation{\mathbf{x}\sim P}[f_{\theta}(\mathbf{x})] - \expectation{\mathbf{x}\sim Q}[\langle \mathbf{x}, \nabla g_{\eta}(\mathbf{x})\rangle - f_{\theta}(\nabla g_{\eta}(\mathbf{x}))]\nonumber,\\
     \mathcal{E} &= \mathbb{E}_{\mathbf{x}\sim P}[\nicefrac{\lVert \mathbf{x} \rVert_{2}^{2}}{2}] + \mathbb{E}_{\mathbf{x} \sim Q}[\nicefrac{\lVert \mathbf{x} \rVert_{2}^{2}}{2}].\nonumber
\end{align}
As shown by the authors, this mini-max problem can be approximated empirically from samples $\mathbf{x}_{i}^{(P)} \sim P$ and $\mathbf{x}_{i}^{(Q)} \sim Q$. Furthermore, due to Brenier's theorem~\cite{brenier1991polar}, $T = \nabla_{\mathbf{x}} g_{\eta}$. As a consequence, one is able to find an \gls{ot} map by taking the gradient $\nabla_{\mathbf{x}} g_{\eta}(\mathbf{x})$.

\subsection{Mini-batch Optimal Transport}\label{sec:minibatch_ot}

A major challenge in \gls{ot} is its time complexity. This motivated different authors~\cite{montavon2016wasserstein,genevay2018learning,sommerfeld2019optimal} to compute the Wasserstein distance between mini-batches rather than complete datasets. For a dataset with $n$ samples, this strategy leads to a dramatic speed-up, since for $K$ mini-batches of size $m \ll n$, one reduces the time complexity of \gls{ot} from $\mathcal{O}(n^{3}\log n)$ to $\mathcal{O}(Km^{3}\log m)$~\cite{sommerfeld2019optimal}. This choice is key when using \gls{ot} as a loss in learning~\cite{damodaran2018deepjdot} and inference~\cite{bernton2019parameter}. Henceforth we describe the mini-batch framework of~\cite{fatras2020learning}, for using \gls{ot} as a loss.

\begin{figure}[ht]
    \centering
    \subfloat[Case I]{\includegraphics[width=0.3\linewidth]{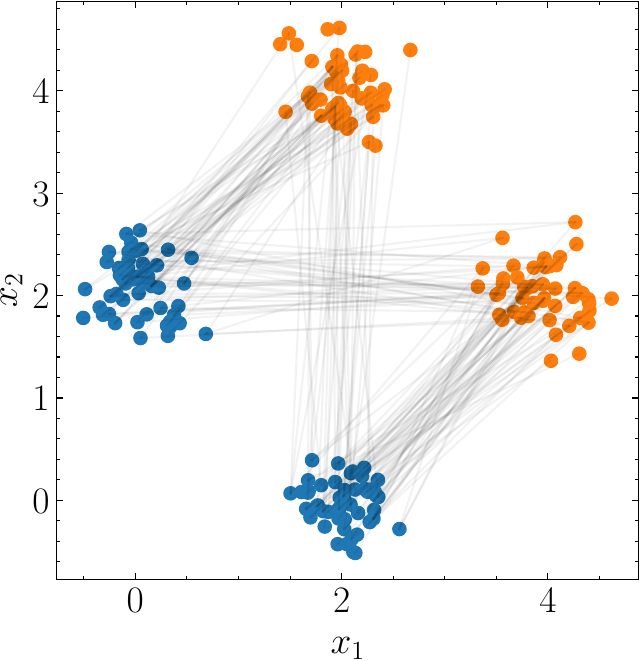}}\hfill
    \subfloat[Case II]{\includegraphics[width=0.3\linewidth]{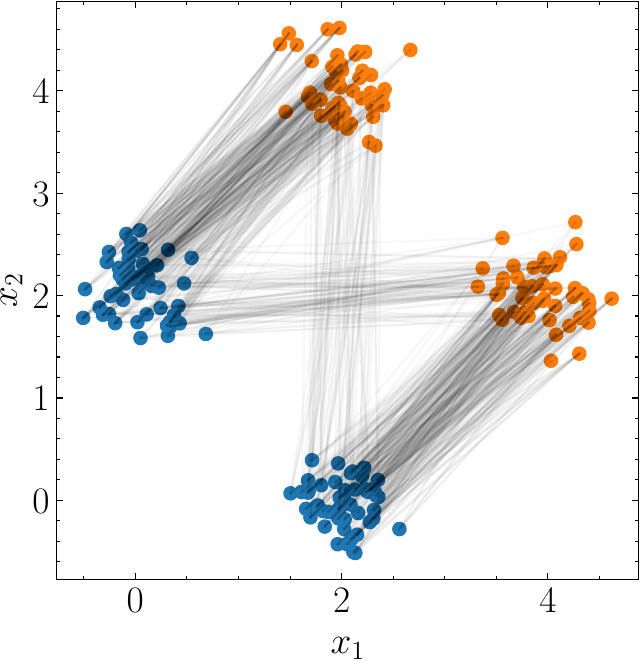}}\hfill
    \subfloat[Case III]{\includegraphics[width=0.3\linewidth]{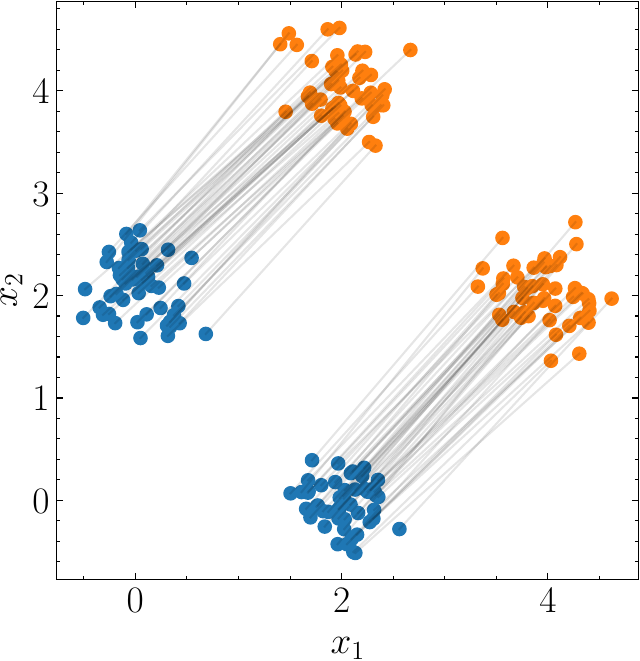}}
    \caption{Mini-batch \gls{ot} between distributions $P$ (in blue) and $Q$ (in orange). As follows, an \gls{ot} plan is calculated with mini-batches of 2 (a), 10 (b) and 100 (c) samples. (c) corresponds to the original \gls{ot} problem. Overall, in mini-batch \gls{ot} the plans become less sparse, due to \gls{ot} being forced to transport all mass between mini-batches.}
    \label{fig:minibatch_ot}
\end{figure}

Let $\mathcal{L}_{OT}$ denote an \gls{ot} loss (e.g., $W_{p}$ or $S_{p,\epsilon}$). Assuming continuous distributions $P$ and $Q$, the \gls{mbot} loss is given by,
\begin{align*}
    \mathcal{L}_{\text{MBOT}}(P, Q) &= \expectation{(\mathbf{X}^{(P)},\mathbf{X}^{(Q)}) \sim P^{\otimes m}\otimes Q^{ \otimes m}}[\mathcal{L}_{OT}(\mathbf{X}^{(P)},\mathbf{X}^{(Q)})],
\end{align*}
where $\mathbf{X}^{(P)} \sim P^{\otimes m}$ indicates $\mathbf{x}_{i}^{(P)} \sim P$, $i=1,\cdots,m$. This loss inherits some properties from \gls{ot}, i.e., it is positive and symmetric, but $\mathcal{L}_{\text{MBOT}}(P, P) > 0$. In practice, let $\{\mathbf{x}_{i}^{(P)}\}_{i=1}^{n_{P}}$ and $\{\mathbf{x}_{j}^{(Q)}\}_{i=1}^{n_{Q}}$ be iid samples from $P$ and $Q$ respectively. Let $\mathcal{I}_{m} \subset \{1,\cdots, n_{P}\}^{m}$ denote a set of $m$ indices. We denote by $\hat{P}_{\mathcal{I}_{m}}$ to the empirical approximation of $P$ with a single mini-batch: $\mathbf{X}^{(P)} = \{\mathbf{x}_{i}^{(P)}: i \in \mathcal{I}_{m}\}$. Therefore,
\begin{align}
\mathcal{L}_{\text{MBOT}}^{(k,m)}(\hat{P},\hat{Q}) = \dfrac{1}{k}\sum_{(\mathcal{I}_{b},\mathcal{I}_{b}') \in \mathbb{I}_{k}}\mathcal{L}_{OT}(\hat{P}_{\mathcal{I}_{b}},\hat{Q}_{\mathcal{I}_{b}'}),\label{eq:mbot}
\end{align}
where $\mathbb{I}_{k}$ is a random set of $k$ mini-batches of size $m$ from $P$ and $Q$. This constitutes an estimator for $\mathcal{L}_{MBOT}(P, Q)$, which converges as $n$ and $k \rightarrow \infty$. We highlight 3 advantages that favor \gls{mbot} for \gls{ml}: (i) it is faster to compute and computationally scalable; (ii) the deviation bound between $\mathcal{L}_{\text{MBOT}}(P, Q)$ and $\mathcal{L}_{\text{MBOT}}^{(k,m)}$ does not depend on the dimensionality of the space; (iii) it has unbiased gradients, i.e., the expected gradient of the sample loss equals the gradient of the true loss. Nonetheless, mini-batch \gls{ot} brings new challenges. As~\cite{fatras2021unbalanced} studies, the use of mini-batches introduces artifacts in \gls{ot} plans, as they become less sparse. This issue is shown in Figure~\ref{fig:minibatch_ot}, which shows the \gls{ot} plan in mini-batch \gls{ot}. We provide further discussion in the next section.

\subsection{Extensions to Optimal Transport}\label{sec:ot_ext}

In this section, we cover 3 extensions to the \gls{ot} problem, namely: (i) unbalanced \gls{ot} (ii) partial \gls{ot} and (iii) \gls{ot} between incomparable spaces.

\noindent\textbf{Unbalanced OT} is an extension to the original Kantorovich problem, which relaxes the mass conservation constraint~\cite{fatras2020learning}. The idea, as discussed in~\cite{liero2018optimal}, is to replace the hard constraint $\gamma \in \Gamma(\mathbf{p}, \mathbf{q})$ by soft constraints in terms of a f-divergence (cf., eq.~\ref{eq:f_div}),
\begin{equation}
\begin{aligned}
    \hat{\gamma}_{\epsilon,\tau} = \argmin{\gamma} \sum_{i=1}^{n}\sum_{j=1}^{m}&\gamma_{ij}c(\mathbf{x}_{i}^{(P)},\mathbf{x}_{j}^{(Q)}) + \epsilon H(\gamma) +\\ &\tau (D_{f}(\gamma_{1}|\mathbf{p}) + D_{f}(\gamma_{2}|\mathbf{q})).
\end{aligned}
\end{equation}
In analogy with the Sinkhorn divergence, unbalanced \gls{ot} defines a divergence $S_{\epsilon, \tau}$ as well. This extension has a few advantages. First, it can be easily implemented on top of the Sinkhorn algorithm~\cite{chizat2018scaling}. Second, it is well defined for positive vectors $\mathbf{p} \in \mathbb{R}^{n}_{+}$, $\mathbf{q} \in \mathbb{R}^{m}_{+}$. Third, it is robust to outliers~\cite{fatras2021unbalanced}, which favors its application to mini-batch \gls{ot}.

\noindent\textbf{Partial OT} defines an \gls{ot} problem in which the transportation plan do not transport a fraction, $0 \leq s \leq 1$, of the total mass. This defines a new set,
\begin{align*}
    \Gamma_{s}(\mathbf{p},\mathbf{q}) = \{\gamma:\sum_{i}\gamma_{ij} \leq q_{j},\sum_{j}\gamma_{ij} \leq p_{i},\sum_{i,j}\gamma_{ij}=s\},
\end{align*}
which substitutes $\Gamma$ in eq.~\ref{eq:KantorovichProblem}. As~\cite{chapel2020partial} proposes, partial \gls{ot} can be solved by adding dummy sink points to which the mass that is not transported, $1 - s$, will be sent to. Similarly to unbalanced \gls{ot}, the partial extension is used to enhance mini-batch \gls{ot}~\cite{nguyen2022improving}.
\begin{figure}[ht]
    \centering
    \includegraphics[width=\linewidth]{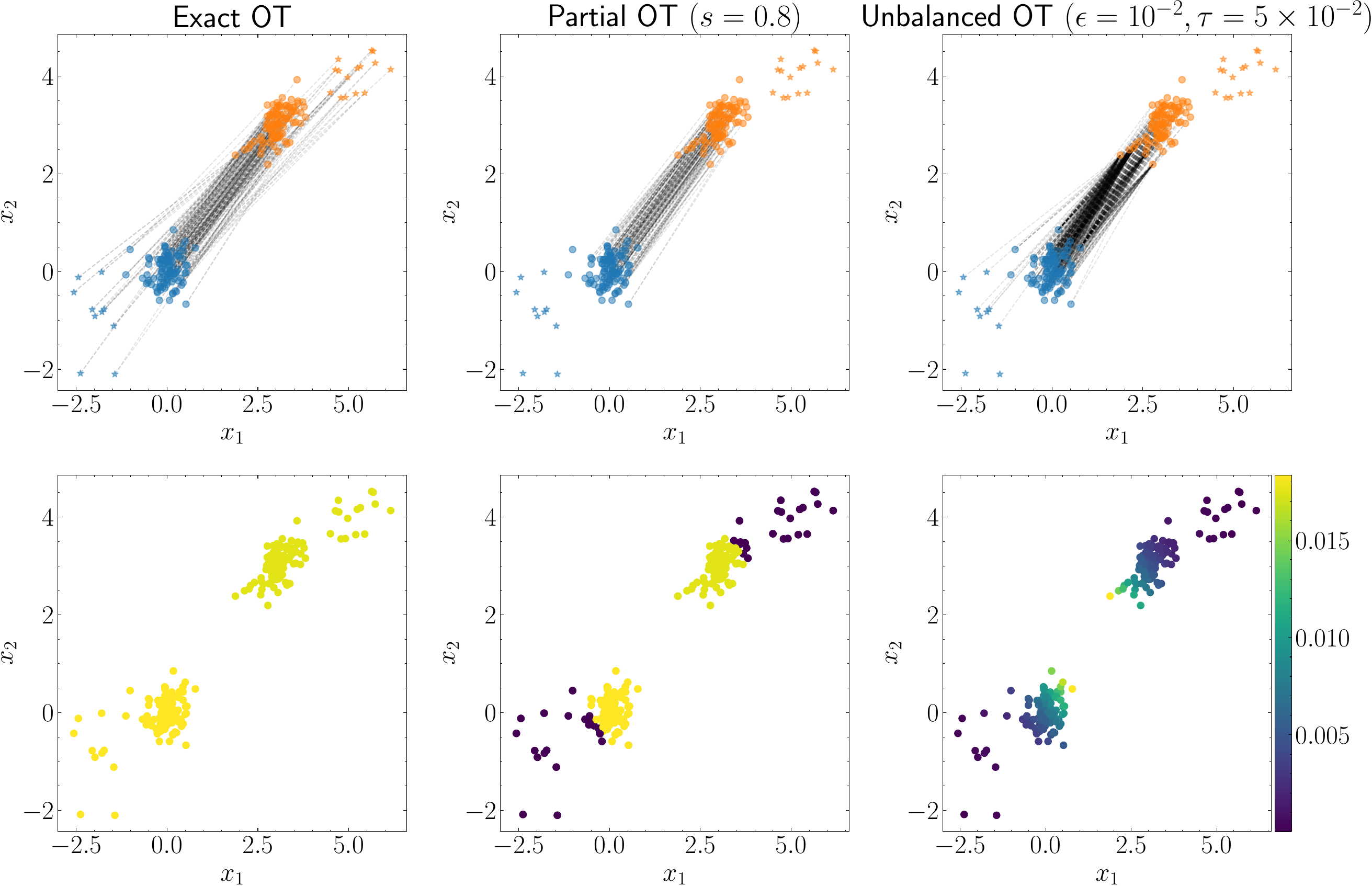}
    \caption{Comparison between exact, partial and unbalanced \gls{ot}. On top, we visualize the \gls{ot} plans as lines joining points transporting mass. On bottom, we color samples by how much mass they send (source distribution) and receive (target distribution), which shows that most outliers in the distributions do not participate in partial nor unbalanced \gls{ot}. This phenomenon highlights the advantage of these extensions for handling datasets with outliers.}
    \label{fig:ot-comparison}
\end{figure}

\noindent\textbf{Gromov-Wasserstein OT} constitutes a series of theoretical extensions to the \gls{ot} problem, when $P$ and $Q$ are probability distributions over \emph{different spaces}. As a consequence, one cannot compute $C_{ij} = c(\mathbf{x}_{i}^{(P)},\mathbf{x}_{j}^{(Q)})$. To that end, one can introduce the \gls{gw} distance~\cite{memoli2011gromov},
\begin{align}
    \text{GW}(P, Q) = \min{\gamma \in \Gamma(\mathbf{p}, \mathbf{q})}\sum_{i,j,k,l}\mathcal{L}(S^{(P)}_{i,k},S_{j,l}^{(Q)})\gamma_{i,k}\gamma_{j,l},\label{eq:gw_dist}
\end{align}
where $S_{i,j}^{(P)}$ quantify the similarity between objects $i$ and $j$ in $P$ (resp. $Q$). We can illustrate this idea with graphs. Formally, a graph $G = (V, E)$ is a pair of a set of vertices $E$ and a set of edges $E$. A histogram over a graph is a pair $P=(V^{(P)}, E^{(P)}, \mathbf{p})$, in which $p_{i}$ is the weight of vertex $i$. An example of Gromov-Wasserstein distance over graph histograms is considering $S_{i,j}^{(P)}$ as the shortest-path distance between vertices $i$ and $j$. 

Additionally, one can define structured objects over graphs, by assigning features to vertices~\cite{titouan2019optimal}, i.e., $\hat{P} = (V^{(P)}, E^{(P)}, \mathbf{X}^{(P)}, \mathbf{p})$, where $\mathbf{X}^{(P)} \in \mathbb{R}^{|V| \times d}$. In this context,~\cite{titouan2019optimal} proposed the \gls{fgw} distance, which interpolates between the standard Wasserstein distance between $(\mathbf{p},\mathbf{X}^{(P)})$ and $(\mathbf{q},\mathbf{X}^{(Q)})$, and the \gls{gw} distance between $(V^{(P)}, E^{(P)}, \mathbf{p})$ and $(V^{(Q)}, E^{(Q)}, \mathbf{q})$, i.e.,
\begin{equation}
\begin{aligned}
    &\text{FGW}_{\alpha}(\hat{P},\hat{Q}) = \min{\gamma\in \Gamma(\mathbf{p}, \mathbf{q})}\sum_{i,j,k,l}L_{i,j,k,l}\gamma_{i,k}\gamma_{j,l},\\
    &\text{for}\,\, L_{i,j,k,l} = (1-\alpha)c(\mathbf{x}^{(P)}_{i},\mathbf{x}^{(Q)}_{j})+\alpha \mathcal{L}(S_{i,k}^{(P)},S_{j,l}^{(Q)}).
\end{aligned}\label{eq:fgw}
\end{equation}

\begin{figure}[ht]
    \centering
    \includegraphics[width=\linewidth]{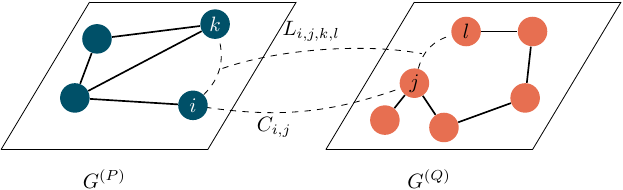}
    \caption{Comparison between two graphs $G^{(P)}$ and $G^{(Q)}$ which lie in different spaces. The Gromov extension of \gls{ot} consists on comparing links $(i, k)$ and $(k, l)$ in both graphs. When features are available for vertices, one can add a cost $C_{ij} = c(\mathbf{x}^{(P)}_{i},\mathbf{x}_{j}^{(Q)})$, leading to the \gls{fgw} distance of~\cite{titouan2019optimal}.}
    \label{fig:example_fgw}
\end{figure}

The ideas behind eqs.~\ref{eq:gw_dist} and~\ref{eq:fgw} are illustrated in~\ref{fig:example_fgw}, in which two graphs are compared with the \gls{gw} and \gls{fgw} distances. In contrast with standard \gls{ot}, the Gromov extension of \gls{ot} is notoriously harder to solve, as problems in eqs.~\ref{eq:gw_dist} and~\ref{eq:fgw} are non-convex. Especially, eq.~\ref{eq:gw_dist} is equivalent to a quadratic assignment problem~\cite{loiola2007survey}, which has NP-hard complexity for arbitrary inputs~\cite[Section 10.6]{peyre2019computational}. Like standard \gls{ot}, one can add entropic regularization to eq.~\ref{eq:gw_dist}~\cite{peyre2016gromov,solomon2016entropic}, alleviating its computational complexity.

\noindent\textbf{Weak Optimal Transport}~\cite{gozlan2017kantorovich} is a generalization of \gls{ot}, in which the ground-cost $c:\mathbb{R}^{d}\times\mathbb{P}(\mathbb{R}^{d})\rightarrow\mathbb{R}$ measures the effort of transportation between a point $\mathbf{x}_{1} \sim P$, and a distribution conditional $\gamma(\cdot|\mathbf{x}_{1})$. In analogy with eq.~\ref{eq:KantorovichProblem},
\begin{align}
    \mathcal{T}_{c}(P, Q) &= \inf{\gamma \in \Gamma(P, Q)}\expectation{\mathbf{x}_{1} \sim P}[c(\mathbf{x}_{1},\gamma(\cdot|\mathbf{x}_{1}))].\label{eq:dual_weak_ot}
\end{align}
The original (strong) \gls{ot} formulation may be retrieved by considering the cost $c(\mathbf{x}_{1}, Q) = \mathbb{E}_{\mathbf{x}_{2}\sim Q}[c(\mathbf{x}_{1},\mathbf{x}_{2})]$. In a similar way to Kantorovich duality (eq.~\ref{eq:dual_kantorovich}), weak-\gls{ot} has a dual formulation as well,
\begin{align*}
    \mathcal{T}_{c}(P, Q) &= \sup{f}\expectation{\mathbf{x}_{1} \sim P}[\varphi^{c}(\mathbf{x}_{1})] + \expectation{\mathbf{x}_{2} \sim Q}[\varphi(\mathbf{x}_{2})],
\end{align*}
where $\varphi^{c}$ is called weak $c-$transform,
\begin{align}
    \varphi^{c}(\mathbf{x}_{1}) &= \inf{P\in \mathbb{P}(\mathbb{R}^{d})}c(\mathbf{x}_{1},P) - \expectation{\mathbf{x}_{2} \sim P}[\varphi(\mathbf{x}_{2})].\label{eq:weak_c_transform}
\end{align}

Based on the weak-\gls{ot} formulation,~\cite{korotin2023neural} proposes a max-min reformulation using convex analysis results~\cite{rockafellar2006integral} for the dual problem in eq.~\ref{eq:dual_weak_ot}. The authors introduce a mapping $T:\mathbb{R}^{d}\times \mathcal{Z} \rightarrow \mathbb{R}^{d}$, and a distribution $S$ (e.g., $S = \mathcal{N}(0, 1$)) for parametrizing $P$ in eq.~\ref{eq:weak_c_transform}, i.e., $P = T(\mathbf{x},\cdot)_{\sharp}S$. Hence,
\begin{align}
    \resizebox{0.885\linewidth}{!}{$\inf{f,T}\expectation{\mathbf{x} \sim Q}[\varphi(\mathbf{x})] + \expectation{\mathbf{x} \sim P}[c(\mathbf{x},T(\mathbf{x},\cdot)_{\sharp}S) - \expectation{z \sim S}[\varphi(T(\mathbf{x},z))]]$}.\label{eq:neural_ot}
\end{align}

The intuition behind the mapping $T$ is as follows. Mappings of the kind $T:\mathbb{R}^{d}\rightarrow\mathbb{R}^{d}$ are called deterministic, i.e., they map $\mathbf{x}_{1} \sim P$ into $\mathbf{x}_{2} \sim Q$. As we discussed in sections~\ref{sec:background} and~\ref{sec:computational_ot}, such a mapping may not exist. As a result, mappings $T(\mathbf{x}_{1}, z)$ are called \emph{stochastic}, as $\mathbf{x}_{2} = T(\mathbf{x}_{1}, z)$ depends on $z \sim S$. In practice,~\cite{korotin2023neural} proposes to parametrize $T$ and $\varphi$ by \glspl{nn} with parameters $\theta$ and $\xi$. The optimization in eq.~\ref{eq:neural_ot} is then carried out by sampling mini-batches from $P$, $Q$ and $S$ respectively.

\begin{figure}[ht]
    \centering
    \subfloat[Case I]{\includegraphics[width=0.45\linewidth]{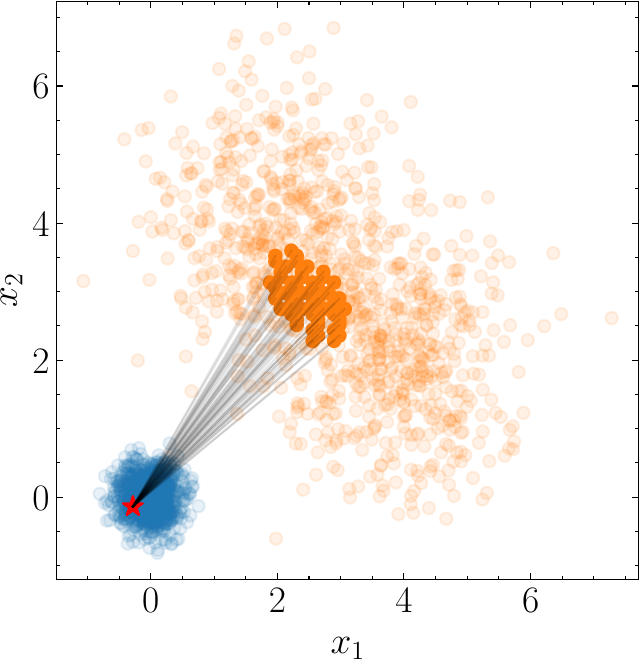}}\hfill
    \subfloat[Case II]{\includegraphics[width=0.45\linewidth]{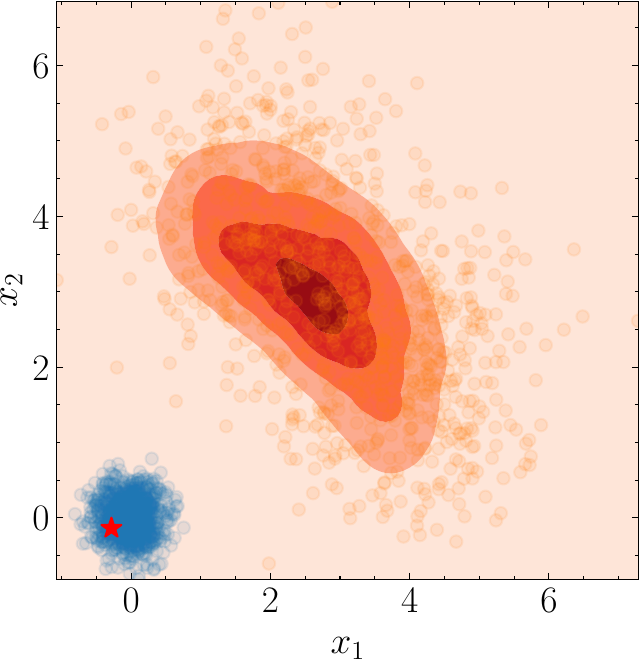}}
    \caption{Comparison between strong (a) and weak (b) \gls{ot}. In (a), a point in distribution $P$ is matched, through an \gls{ot} plan, to points in $Q$. In (b), for $\mathbf{x}_{1}$ in $P$ (red star), one has a distribution $\gamma(\cdot|\mathbf{x}_{1})$ over points in $Q$.}
    \label{fig:Weak-Strong-OT}
\end{figure}
\section{Supervised Learning}\label{sec:supervised-learning}

In this section we review applications of \gls{ot} for supervised learning in two directions. First, in section~\ref{sec:ot_loss}, we review the use of \gls{ot} as a loss in classification. Second, in section~\ref{sec:fairness}, we review \gls{ot} for fairness.

\subsection{OT as a loss}\label{sec:ot_loss}

\noindent\textbf{Empirical Risk Minimization.} Let $\mathcal{X} = \mathbb{R}^{d}$ be a feature space, and $\mathcal{Y} = \{1,\cdots,n_{c}\}$ be a label space. For a distribution $P \in \mathbb{P}(\mathcal{X})$, a ground truth $h_{0}:\mathcal{X}\rightarrow\mathcal{Y}$, and a loss function $\mathcal{L}:\mathcal{Y}\times\mathcal{Y}\rightarrow\mathbb{R}$, the risk of $h:\mathcal{X}\rightarrow\mathcal{Y}$ is,
\begin{align}
    \mathcal{R}_{P}(h) = \expectation{\mathbf{x} \sim P}[\mathcal{L}(h(\mathbf{x}), h_{0}(\mathbf{x}))].\label{eq:true_risk}
\end{align}

In classification, one defines a family of functions $\mathcal{H} \subset \mathcal{X}^{\mathcal{Y}}$, where one searches for $h^{\star} \in \mathcal{H}$ that minimizes eq.~\ref{eq:true_risk}. In practice, one does not have access to $P$, but rather to samples $\mathbf{x}_{i}^{(P)} \sim P$ and $y_{i}^{(P)} = h_{0}(\mathbf{x}_{i}^{(P)})$, which leads to,
\begin{align}
    \hat{\mathcal{R}}_{P}(h) = \dfrac{1}{n}\sum_{i=1}^{n}\mathcal{L}(h(\mathbf{x}_{i}^{(P)}), y_{i}^{(P)}).\label{eq:empirical_risk}
\end{align}
In analogy to $h^{\star}$, one can minimize the empirical risk over $h \in \mathcal{H}$. This strategy is known as \gls{erm}~\cite{vapnik2013nature}. In the context of \glspl{nn}, $h = h_{\theta}$ is parametrized by the weights $\theta$ of the network. An usual choice in \gls{ml} is $\mathcal{L}(\hat{y}_{i},y_{i}) = \sum_{k=1}^{n_{c}}y_{ik}\log\hat{y}_{ik}$ where $y_{ik} = 1$ if and only if the sample $i$ belongs to class $k$, and $\hat{y}_{ik} \in [0, 1]$ is the predicted probability of sample $i$ belonging to $k$. This choice of loss is related to \gls{mle}, and hence to the \gls{kl} divergence (see~\cite[Chapter 4]{bishop2006pattern}) between $P(Y|X)$ and $P_{\theta}(Y|X)$.

\begin{figure}[ht]
    \centering
    \subfloat[Case I]{\includegraphics[scale=0.3]{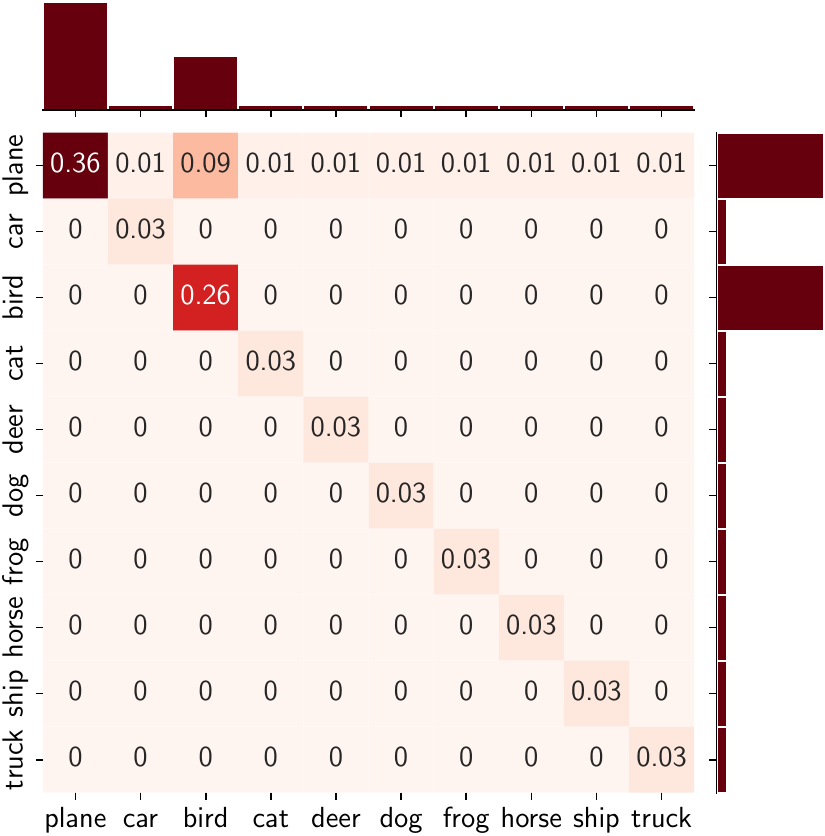}}\hfill
    \subfloat[Case II]{\includegraphics[scale=0.37]{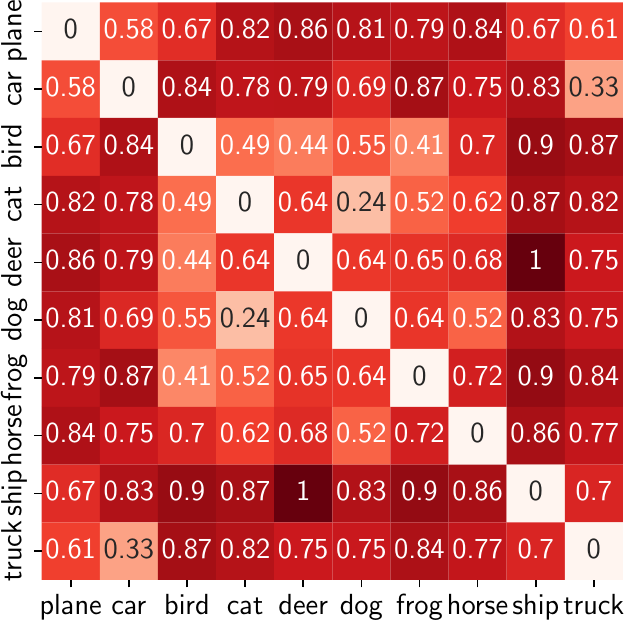}}
    \caption{(a) \gls{ot} plan between two probability vectors over classes in CIFAR10~\cite{krizhevsky2009learning}. (b) Pairwise distances between Word2Vec~\cite{mikolov2013distributed} embeddings of class labels. The Wasserstein loss proposed by~\cite{frogner2015learning} corresponds to the Frobenius inner product between those two matrices.}
    \label{fig:ot_loss_kusner}
\end{figure}

As remarked by~\cite{frogner2015learning}, this choice disregards similarities in the label space. As such, a prediction of semantically close classes (e.g., a car and a truck) yields the same loss as semantically distant ones (e.g., a dog and a ship). To remedy this issue,~\cite{frogner2015learning} proposes using the Wasserstein distance between the network's predictions and the ground-truth vector, $\mathcal{L}(\hat{y},y) = \text{min}_{\gamma \in U(\hat{y},y)}\langle \gamma, \mathbf{C} \rangle_{F}$, where $\mathbf{C} \in \mathbb{R}^{n_{c}\times n_{c}}$ is a metric between labels. This is illustrated in Figure~\ref{fig:ot_loss_kusner}, in the context of the CIFAR10 dataset~\cite{krizhevsky2009learning}.

With respect~\cite{frogner2015learning}, similar principles were applied by~\cite{liu2020importance,liu2020severity} in the context of semantic segmentation, in which the authors use a pre-defined ground-metric corresponding to the severity of wrong classification. Likewise~\cite{fatras2022wasserstein} employed a Wasserstein term between probability vectors as a regularizer term in the context of learning under noisy labels.

\noindent\textbf{Distributionally Robust Optimization.} Besides its contributions to the \gls{erm} framework, \gls{ot} can be used as a tool in \gls{dro}~\cite{rahimian2019distributionally, kuhn2019wasserstein}. \gls{dro} can be used as a generalization of the \gls{erm} when data is noisy. The idea consists on regularizing the problem by robustifying it,
\begin{align}
    \hat{\mathcal{R}}_{P,\epsilon}(h) = \sup{Q \in \mathbb{B}(\epsilon, 1, \hat{P})}\expectation{\mathbf{x} \sim Q}[\mathcal{L}(h(\mathbf{x}), h_{0}(\mathbf{x}))],\label{eq:wdro1}
\end{align}
where $\mathbb{B}_{\epsilon, 1}(\hat{P})$ is the $1-$Wasserstein ball with radius $\epsilon$ around $\hat{P}$. As shown in~\cite{shafieezadeh2019regularization}, minimizing $\hat{\mathcal{R}}_{P,\epsilon}$ over $h \in \mathcal{H}$ is equivalent to a regularized \gls{erm} problem,
\begin{align*}
    \hat{h}_{WDRO} &= \inf{h \in \mathcal{H}}\hat{\mathcal{R}}_{P}(h) + \epsilon \text{Lip}(\mathcal{L}) \lVert h \rVert_{*},
\end{align*}
where $\text{Lip}(\mathcal{L})$ is the Lipschitz constant of the loss $\mathcal{L}$ and $\lVert h \rVert_{*} = \text{sup}\{|h(\mathbf{x})|:\mathbf{x}\in\mathcal{X}\}$ is the dual norm of $h$.

\subsection{Fairness}\label{sec:fairness}

The analysis of biases in \gls{ml} algorithms has received increasing attention, due to the now widespread use of these systems for automatizing decision-making processes. In this context, fairness~\cite{zafar2017fairness} is a sub-field of \gls{ml} concerned with achieving fair treatment to individuals in a population. Throughout this section, we focus on fairness in binary classification.

There are mainly three approaches in the fairness literature~\cite{caton2020fairness}: pre, in and post-processing. First, pre-processing corresponds to transforming the input data so that models trained on it achieve some notion of fairness. Second, in-processing incorporates a metric of fairness in the learning process of a model. Finally, post-processing correspond to applying transformations to a model's outputs, or performing model selection. In this section, we focus on the first two strategies. In this context \gls{ot} is used to enforce statistical parity~\cite{dwork2012fairness}, which is expressed in probabilistic terms as,
\begin{equation}
    P(h(X)=1|S=0) = P(h(X)=1|S=1)\label{eq:statistical_parity}
\end{equation}
\noindent\textbf{Pre-processing.} Based on equation~\ref{eq:statistical_parity},~\cite{gordaliza2019obtaining} proposes a pre-processing technique, based on \gls{ot}, to enforce statistical parity. Let $P_{s} = P(X|S=s)$, $s=0, 1$. Their idea is to devise a transformation $T$ with $T_{\sharp}P_{0} = T_{\sharp}P_{1}$. In addition, $T$ should not destroy information, i.e. $P_{s}$ and $T_{\sharp}P_{s}$ should be as close as possible. As a result,~\cite{gordaliza2019obtaining} chooses to realize both of these constraints with the Wasserstein barycenter $\hat{P}_{t} = \mathcal{B}([1 - t,t];\{\hat{P}_{0},\hat{P}_{1}\})$ (see eq.~\ref{eq:wbary}),
\begin{align*}
    \hat{P}_{t}(\mathbf{x}) &= \sum_{i=1}^{n_{0}}\sum_{j=1}^{n_{1}}\gamma_{ij}\delta(\mathbf{x}-\pi_{t}(\mathbf{x}_{i}^{(P_{0})}, \mathbf{x}_{j}^{(P_{1})})),\\
    \pi_{t}(\mathbf{x}_{i}^{(P_{0})}, \mathbf{x}_{j}^{(P_{1})}) &= (1-t)\mathbf{x}_{i}^{(P_{0})}+t\mathbf{x}_{j}^{(P_{1})}.
\end{align*}
In~\cite{gordaliza2019obtaining}, the authors use $t=\nicefrac{1}{2}$, so as to avoid disparate impact. Furthermore, trying to build $T$ s.t. $T_{\sharp}P_{0} = T_{\sharp}P_{1}$ may compromise too much accuracy. As a result,~\cite{gordaliza2019obtaining} introduce the concept of \emph{random repair}, which corresponds to drawing $b_{1},\cdots,b_{n_{0}+n_{1}} \sim Be(\lambda)$, where $Be(\lambda)$ is a Bernoulli distribution with parameter $\lambda$. For $P_{0}$ (resp. $P_{1}$),
\begin{align*}
    \mathbf{X}^{(P_{0})} &= \bigcup_{i=1}^{n_{0}}\begin{cases}
        \{\mathbf{x}_{i}^{(P_{0})}\}&\text{ if }b_{i} = 0\\
        \{\mathbf{x}_{t,ij}:\gamma_{ij} > 0\}&\text{ if }b_{i} = 1
    \end{cases},
\end{align*}
where the \emph{amount of repair} $\lambda$ regulates the trade-off between fairness ($\lambda = 1$) and classification performance ($\lambda = 0$).

\noindent\textbf{In-Processing.} In another direction, one may use \gls{ot} for devising a regularization term to enforce fairness in \gls{erm}. Two works follow this strategy, namely~\cite{oneto2020exploiting} and~\cite{chiappa2020general}. In this sense, let $f = h \circ g$, be the composition of a feature extractor $g$ and a classifier $h$. Therefore~\cite{oneto2020exploiting} proposes minimizing,
\begin{align*}
    (h^{\star},g^{\star}) = \argmin{h,g}\dfrac{1}{n}\sum_{i=1}^{n}\mathcal{L}(h(g(\mathbf{x}_{i})),y_{i})+\beta d(P_{0},P_{1}),
\end{align*}
for $\beta > 0$, $P_{0} = P(X|S=0)$ (resp $S = 1$) and $d$ being either the \gls{mmd} or the Sinkhorn divergence (see section~\ref{sec:background}). On another direction,~\cite{chiappa2020general} considers the \emph{output distribution} $P_{s} = P(h(g(X))|S=s)$. Their approach is more general, as they suppose that attributes can take more than 2 values, namely $s \in \{1,\cdots, N_{S}\}$. Their insight is that the output distribution should be transported to the distribution closest to each group's conditional, namely $P_{k}=P(h(g(X))|S=k)$. This corresponds to $P_{\bar{s}} = \mathcal{B}(\alpha;\{P_{k}\}_{k=1}^{N_{S}})$, for $\alpha_{k} = \nicefrac{n_{k}}{n}$ and $n=\sum_{k}n_{k}$. As in~\cite{oneto2020exploiting}, the authors enforce this condition through regularization, namely,
\begin{align*}
    \min{h, g} \dfrac{1}{n}\sum_{i=1}^{n}\mathcal{L}(h(g(\mathbf{x}_{i})),y) + \beta \sum_{k=1}^{N_{S}}\dfrac{n_{k}}{n}W_{2}(P_{k},\bar{P}).
\end{align*}
\section{Unsupervised Learning}\label{sec:unsupervised-learning}

In this section, we discuss unsupervised learning techniques that leverage the Wasserstein distance as a loss function. We consider three cases: generative modeling (section~\ref{sec:gm}), representation learning (section~\ref{sec:representation_learning}) and clustering (section~\ref{sec:clustering}).

\subsection{Generative Modeling}\label{sec:gm}

There are mainly three types of generative models that benefit from \gls{ot}, namely, \glspl{gan} (section~\ref{sec:gans}), \glspl{vae} (section~\ref{sec:vae}) and normalizing flows (section~\ref{sec:cnf}). As discussed in~\cite{genevay2017gan,lei2019geometric}, \gls{ot} provides an unifying framework for these principles through the \gls{mke} framework, introduced by~\cite{bassetti2006minimum}, which, given samples from $P_{0}$, tries to find $\theta$ that minimizes $W_{2}(P_{\theta}, P_{0})$. In the following we denote by $\mathcal{X}$ the input space (e.g., image space) and $\mathcal{Z}$ to the latent space (e.g., an Euclidean space $\mathbb{R}^{p}$).

\subsubsection{Generative Adversarial Networks}\label{sec:gans}

\begin{figure*}[ht]
    \centering
    \subfloat[Case I]{
        \resizebox{0.3\linewidth}{!}{
    	\begin{tikzpicture}[auto, node distance=1cm,>=latex']
    	\node [] (input) {};
    	\node [block, right=2 of input] (generator) {$g_{\theta}$};
    	\node [below=2 of generator, align=center] (samples) {\huge{\faDatabase}\\Dataset};
    	\node [below=1 of generator] (aux) {};
    	\node [block, right=2 of aux] (discriminator) {$h_{\xi}$};
    	\node [right=2 of discriminator, align=center] (output) {Real or\\Fake?};
    	\draw [->] (input) -- node [name=edge1] {$\mathbf{z}\sim Q$} (generator);
    	\draw [->] (generator) -| node [name=edge1] {$\mathbf{x} \sim P_{\theta}=g_{\theta\sharp}Q$} (discriminator);
    	\draw [->] (samples) -| node [name=edge1, pos=0.6, right] {$\mathbf{x} \sim P_{0}$} (discriminator);
    	\draw [->] (discriminator) -- node [] {} (output);
    	\end{tikzpicture}
        }
    }\hfill
    \subfloat[Case II]{
        \resizebox{0.3\linewidth}{!}{
            \begin{tikzpicture}
                \node[trapezium,thick,draw,rotate=-90,minimum height=1cm] (t1) at (0, 0) {};
                \node[trapezium,thick,draw,rotate=90,minimum height=1cm] (t3) at (4,0) {};
                \node[rectangle,thick,draw,minimum height=1cm,minimum width=.1cm] at (1.32, .3) {};
                \node[rectangle,thick,draw,minimum height=1cm,minimum width=.1cm, fill=white, fill opacity=1.0] (O) at (1.25, 0) {};
                \node[rectangle,thick,draw,minimum height=2cm,minimum width=.1cm] at (5.35, .3) {};
                \node[rectangle,thick,draw,minimum height=2cm,minimum width=.1cm, fill=white, fill opacity=1.0] (O1) at (5.25, 0) {};
                \node[rectangle,thick,draw,minimum height=2cm,minimum width=.1cm, fill=white, fill opacity=1.0] (I) at (-1.25, 0) {};
                \node[rectangle,thick,draw,minimum height=1cm,minimum width=.1cm, fill=white, fill opacity=1.0] (Z) at (2.6, 0) {};
                \node[rectangle] (txt1) at (-1.25, -1.25) {$\mathbf{x}$};
                \node[rectangle] (txt2) at (6.1, 0) {$\hat{\mathbf{x}}$};
                \node[rectangle] (txt3) at (1.25, -0.75) {$\mathbf{m}_{\eta}(\mathbf{x})$};
                \node[rectangle] (txt4) at (1.25, 1.07) {$\sigma_{\eta}(\mathbf{x})^{2}$};
                \node[rectangle] (txt4) at (2.6, -0.75) {$\mathbf{z}$};
                \node[rectangle] (txt6) at (5.25, -1.25) {$\mathbf{m}_{\theta}(\mathbf{z})$};
                \node[rectangle] (txt7) at (5.25, 1.7) {$\sigma_{\theta}(\mathbf{z})$};
                \node[rectangle] (txt8) at (2.0, 0) {$\sim$};
                \node[rectangle] (txt9) at (5.7, 0) {$\sim$};
                \node[rectangle] (txt10) at (0, 0) {$f_{\eta}$};
                \node[rectangle] (txt11) at (4, 0) {$g_{\theta}$};
                
                \draw[->,thick] (I.east) -- (t1.south);
                \draw[->,thick] (t1.north) -- (O.west);
                \draw[->,thick] (Z.east) -- (t3.north);
                \draw[->,thick] (t3.south) -- (O1.west);
            \end{tikzpicture}
        }
    }\hfill
    \subfloat[Case III]{
        \resizebox{0.3\linewidth}{!}{
            \begin{tikzpicture}
                \node[trapezium,thick,draw,rotate=-90,minimum height=1cm] (t1) at (0, 0) {};
                \node[trapezium,thick,draw,rotate=90,minimum height=1cm] (t2) at (3,0) {};
                \node[rectangle,thick,draw,minimum height=2cm,minimum width=.1cm, fill=white, fill opacity=1.0] (O1) at (4.25, 0) {};
                \node[rectangle,thick,draw,minimum height=2cm,minimum width=.1cm, fill=white, fill opacity=1.0] (I) at (-1.25, 0) {};
                \node[rectangle,thick,draw,minimum height=1cm,minimum width=.1cm, fill=white, fill opacity=1.0] (Z) at (1.5, 0) {};
                \node[rectangle,thick,draw,minimum height=2cm, fill=white, fill opacity=1.0] (Dist1) at (-0.75, -3) {\includegraphics[width=1.8cm]{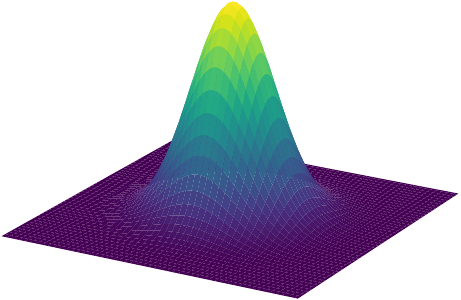}};
                \node[rectangle] (txt1) at (-1.25, -1.25) {$\mathbf{x}$};
                \node[rectangle] (txt2) at (4.25, -1.25) {$\hat{\mathbf{x}}$};
                \node[rectangle] (txt10) at (0, 0) {$f_{\eta}$};
                \node[rectangle] (txt11) at (3, -3) {$h_{\xi}$};
                \node[rectangle] (txt12) at (3, 0) {$g_{\theta}$};
                
                \node[trapezium,thick,draw,rotate=-90,minimum height=.5cm] (t3) at (3, -3) {};
                \node[rectangle, align=center] (txt12) at ([xshift=40pt]t3.north) {Real or\\ Fake?};
                
                \draw[->,thick] (I.east) -- (t1.south);
                \draw[->,thick] (t1.north) -- (Z.west);
                \draw[->,thick] (Z.east) -- (t2.north);
                \draw[->,thick] (t2.south) -- (O1.west);
                \draw[->,thick] (Z.south) |- ([yshift=8pt]t3.south);
                \draw[->,thick] ([yshift=-8pt]Dist1.east) -- ([yshift=-8pt]t3.south);
                \draw[->,thick] (t3.north) -- ([xshift=20pt]t3.north);
            \end{tikzpicture}
        }
    }
    \caption{Different architectures in generative modeling. (a) shows \glspl{gan}, which is composed of a generator of synthetic samples, and a discriminator of real and fake samples. (b) shows \glspl{vae}, which are a stochastic version of auto-encoders, an architecture composed of an encoder and a decoder. (c) shows the architecture of \glspl{aae}, which combine ideas of \glspl{gan} and \glspl{vae}.}
    \label{fig:generative_models}
\end{figure*}

\gls{gan} is a \gls{nn} architecture proposed by~\cite{goodfellow2014generative} for sampling from a complex probability distribution $P_{0}$. This architecture has two networks. First, a generator $g_{\theta}:\mathcal{Z} \rightarrow \mathcal{X}$, that maps $\mathbf{z} \sim Q$ to a sample $\mathbf{x} \in \mathcal{X}$. $Q$ is assumed to be a simple distribution (e.g., $\mathcal{N}(\mathbf{0},\sigma^{2}\mathbf{I})$). Second, a discriminator $h_{\xi}:\mathcal{X}\rightarrow [0,1]$. The \gls{gan} architecture is shown in Figure~\ref{fig:generative_models} (a), and is trained with the following optimization problem,
\begin{align*}
\min{\theta}\max{\xi}\expectation{\mathbf{x}\sim P_{0}}[\log(h_{\xi}(\mathbf{x}))] + \expectation{\mathbf{z} \sim Q}[\log(1-h_{\xi}(g_{\theta}(\mathbf{z})))].
\end{align*}
In other words, $h_{\xi}$ is trained to maximize the probability of assigning the correct label to $\mathbf{x}$ (labeled 1) and $g_{\theta}(\mathbf{z})$ (labeled 0). Conversely, $g_{\theta}$ is trained to minimize $\log(1 - h_{\xi}(g_{\theta}(\mathbf{z})))$. As a consequence it minimizes the probability that $h_{\xi}$ guesses its samples correctly.

As shown in~\cite{goodfellow2014generative}, for an optimal discriminator $h_{\xi^{\star}}$, the generator cost is equivalent to the so-called \gls{js} divergence. In this sense,~\cite{arjovsky2017wasserstein} proposed the \gls{wgan} algorithm, which substitutes the \gls{js} divergence by the \gls{kr} metric (equation~\ref{eq:kantorovich_rubinstein}),
\begin{equation*}
    \min{\theta}\max{h_{\xi} \in \text{Lip}_{1}}\mathcal{L}_{W}(\theta,\xi) = \expectation{\mathbf{x} \sim P_{0}}\biggr[h_{\xi}(\mathbf{x})\biggr] - \expectation{\mathbf{z}\sim Q}\biggr[h_{\xi}(g_{\theta}(\mathbf{z}))\biggr].
\end{equation*}
Nonetheless, the~\gls{wgan} involves maximizing $\mathcal{L}_{W}$ over $h_{\xi} \in \text{Lip}_{1}$, which is not straightforward for \glspl{nn}. Possible solutions are: (i) clipping the weights $\xi$~\cite{arjovsky2017wasserstein}; (ii) penalizing the gradients of $h_{\xi}$~\cite{gulrajani2017improved}; (iii) normalizing the spectrum of the weight matrices~\cite{miyato2018spectral}. Surprisingly, even though (ii) and (iii) improve over (i), several works have confirmed that the \gls{wgan} and its variants \emph{do not estimate the Wasserstein distance}~\cite{pinetz2019estimation,mallasto2019well,stanczuk2021wasserstein}.

In another direction, one can calculate $\nabla_{\theta}W_{p}$ through the primal problem (eq.~\ref{eq:KantorovichProblem}). To alleviate the computational and memory complexity of \gls{ot}, one uses mini-batches~\cite{genevay2018learning,fatras2020learning} and computes the Sinkhorn divergence~\cite{genevay2018learning}. It remains the question on how to calculate the gradients w.r.t. $\theta$. First,~\cite{genevay2018learning} proposes to back-propagate the gradients through the Sinkhorn iterations, but it is commonly time consuming and numerically unstable. Second~\cite{feydy2019interpolating} and~\cite{xie2020fast} advocate to take the gradients at convergence of Sinkhorn iterations, i.e. assuming that the transport plan does not depend on $\theta$. This is justified by the Envelope theorem~\cite{bertsekas1998network}. 

In addition, as discussed in~\cite{genevay2018learning}, an Euclidean ground-cost is likely not meaningful for complex data (e.g., images). The authors thus propose learning a parametric ground cost $(\mathbf{C}_{\eta})_{ij} = \lVert f_{\eta}(\mathbf{x}_{i}^{(P_{\theta})}) - f_{\eta}(\mathbf{x}^{(P_{0})}_{j}) \rVert_{2}^{2}$, where $f_{\eta}:\mathcal{X}\rightarrow\mathcal{Z}$ is a \gls{nn} that learns a representation for $\mathbf{x} \in \mathcal{X}$. Overall the optimization problem proposed by~\cite{genevay2018learning} is,
\begin{align}
    \min{\theta}\max{\eta} S_{c_{\eta},\epsilon}(P_{\theta}, P_{0}).\label{eq:genevay_obj_fn}
\end{align}
We stress the fact that \emph{engineering/learning} the ground-cost $c_{\eta}$ is important for having a meaningful metric between distributions, since it serves to compute distances between samples. For instance, the Euclidean distance is known to not correlate well with perceptual or semantic similarity between images~\cite{wang2004image}.

Finally, as in~\cite{kolouri2018sliced,deshpande2019max,kolouri2019generalized,wu2019sliced}, one can employ sliced Wasserstein metrics (see section~\ref{sec:projbased}). This has two advantages: (i) computing the sliced Wasserstein distances is computationally less complex, (ii) these distances are more robust w.r.t. the curse of dimensionality~\cite{nadjahi2019asymptotic,nadjahi2020statistical}. These properties favour their usage in generative modeling, as data is commonly high dimensional.

\subsubsection{Autoencoders}\label{sec:vae}
In a parallel direction, one can leverage autoencoders for generative modeling. This idea was introduced in~\cite{kingma2014autoencoding}, who proposed using stochastic encoding and decoding functions. Let $f_{\eta}:\mathcal{X}\rightarrow\mathcal{Z}$ be an encoder network. Instead of mapping $\mathbf{z} = f_{\eta}(\mathbf{x})$ deterministically, $f_{\eta}$ predicts a mean $\mathbf{m}_{\eta}(\mathbf{x})$ and a variance $\sigma_{\eta}(\mathbf{x})^{2}$ from which the code $\mathbf{z}$ is sampled, that is, $\mathbf{z} \sim Q_{\eta}(\mathbf{z}|\mathbf{x}) = \mathcal{N}(\mathbf{m}_{\eta}(\mathbf{x}), \sigma_{\eta}(\mathbf{x})^{2})$. The stochastic decoding function $g_{\theta}:\mathcal{Z}\rightarrow\mathcal{X}$ works similarly for reconstructing the input $\hat{\mathbf{x}}$. This is shown in figure~\ref{fig:generative_models} (b). In this framework, the decoder plays the role of generator.

The \gls{vae} framework is built upon variational inference, which is a method for approximating probability densities~\cite{blei2017variational}. Indeed, for a parameter vector $\theta$, generating new samples $\mathbf{x}$ is done in two steps: (i) sample $\mathbf{z}$ from the prior $Q(\mathbf{z})$ (e.g., a Gaussian), then (ii) sample $\mathbf{x}$ from the conditional $P_{\theta}(\mathbf{x}|\mathbf{z})$. The issue comes from calculating the marginal,
\[
    P_{\theta}(\mathbf{x}) = \int Q(\mathbf{z})P_{\theta}(\mathbf{x}|\mathbf{z})d\mathbf{z},
\]
which is intractable. This hinders the \gls{mle}, which relies on $\log P_{\theta}(\mathbf{x})$. \glspl{vae} tackle this problem by first considering an approximation $Q_{\eta}(\mathbf{z}|\mathbf{x})\approx P_{\theta}(\mathbf{z}|\mathbf{x})$. Secondly, one uses the \gls{elbo},
\begin{align*}
    \text{ELBO}(Q_{\eta}) &= \expectation{\mathbf{z}\sim Q_{\eta}}[\log P_{\theta}(\mathbf{x}, \mathbf{z})] - \expectation{\mathbf{z}\sim Q_{\eta}}[\log Q_{\eta}(\mathbf{z})],
\end{align*}
instead of the log-likelihood. Indeed, as shown in~\cite{kingma2014autoencoding}, the \gls{elbo} is a lower bound for the log-likelihood. As follows, one turns to the maximization of the \gls{elbo}, which can be rewritten as,
\begin{equation}
    \mathcal{L}(\theta,\eta) = \expectation{\mathbf{x}\sim P_{0}}\biggr[\expectation{\mathbf{z}\sim Q_{\eta}}[\log P_{\theta}(\mathbf{x}|\mathbf{z})]-\text{KL}(Q_{\eta}\lVert Q)\biggr].
\end{equation}
A somewhat middle ground between the frameworks of~\cite{goodfellow2014generative} and~\cite{kingma2014autoencoding} is the \gls{aae} architecture~\cite{makhzani2015adversarial}, shown in figure~\ref{fig:generative_models} (c). \glspl{aae} are different from \glspl{gan} in two points, (i) an encoder $f_{\eta}$ is added, for mapping $\mathbf{x}\in\mathcal{X}$ into $\mathbf{z}\in\mathcal{Z}$; (ii) the adversarial component is done in the latent space $\mathcal{Z}$. While the first point puts the \gls{aae} framework conceptually near \glspl{vae}, the second shows similarities with \glspl{gan}.

Based on both \glspl{aae} and \glspl{vae},~\cite{tolstikhin2018wasserstein} proposed the \gls{wae} framework, which is a generalization of \glspl{aae}. Their insight is that, when using a deterministic decoding function $g_{\theta}$, one may simplify the \gls{mk} formulation,
\[
\inf{\gamma \in \Gamma}\expectation{(\mathbf{x}_{1},\mathbf{x}_{2})\sim\gamma}[c(\mathbf{x}_{1},\mathbf{x}_{2})] = \inf{Q_{\eta}=Q}\expectation{\mathbf{x}\sim P_{0}}\biggr[\expectation{\mathbf{z}\sim Q_{\eta}}[c(\mathbf{x},g_{\theta}(\mathbf{z}))]\biggr].
\]
As follows,~\cite{tolstikhin2018wasserstein} suggests enforcing the constraint in the infimum through a penalty term $\Omega$, leading to,
\[
\mathcal{L}(\theta, \eta) = \expectation{\mathbf{x}\sim P_{0}}\biggr[\expectation{\mathbf{z}\sim Q_{\eta}}[c(\mathbf{x},g_{\theta}(\mathbf{z}))]\biggr] + \lambda \Omega(Q_{\eta}, Q),
\]
for $\lambda > 0$. This expression is minimized jointly w.r.t. $\theta$ and $\eta$. As choices for the penalty $\Omega$,~\cite{tolstikhin2018wasserstein} proposes using either the \gls{js} divergence, or the \gls{mmd} distance. In the first case the formulation falls back to a mini-max problem similar to the \gls{aae} framework. A third choice was proposed by~\cite{patrini2020sinkhorn}, which relies on the Sinkhorn loss $S_{c,\epsilon}$, thus leveraging the work of~\cite{genevay2018learning}.

\subsubsection{Continuous Normalizing Flows}\label{sec:cnf}

\gls{ot} is linked to \glspl{nf}, especially \glspl{cnf}, through its dynamical formulation. We thus focus on this class of algorithms. For a broader review, we refer the readers to~\cite{kobyzev2020normalizing}. \glspl{nf} rely on the chain rule for computing changes in probability distributions. Let $\mathbf{T}:\mathbb{R}^{d}\rightarrow\mathbb{R}^{d}$ be a diffeomorphism. If, for $\mathbf{z} \sim Q$ and $\mathbf{x} \sim P$, $\mathbf{z} = \mathbf{T}(\mathbf{x})$, one has,
\begin{align*}
    \log P(\mathbf{x}) &= \log Q(\mathbf{z}) - \log |\det\nabla\mathbf{T}^{-1}|.
\end{align*}

Following~\cite{kobyzev2020normalizing}, a \gls{nf} is characterized by a sequence of diffeomorphisms $\{T_{i}\}_{i=1}^{N}$ that transform a simple probability distribution (e.g., a Gaussian distribution) into a complex one. If one understands this sequence \emph{continuously}, they can be modeled through dynamic systems. In this context, \glspl{cnf} are a class of generative models based on dynamic systems,
\begin{align}
    \mathbf{z}(0,\mathbf{x}) = \mathbf{x}\text{, and, }
    \dot{\mathbf{z}}(t,\mathbf{x}) = \mathbf{F}_{\theta}(\mathbf{z}(t,\mathbf{x})).\label{eq:ode}
\end{align}
which is an \gls{ode}. Let $\mathbf{z}_{t}(\mathbf{x}) = \mathbf{z}(t,\mathbf{x})$. By discretizing the time variable, equation~\ref{eq:ode} becomes $\mathbf{z}_{n+1} = \mathbf{z}_{n} + \tau \mathbf{F}_{\theta}(\mathbf{z}_{n})$, for a step-size $\tau > 0$. This equation is similar to the ResNet structure~\cite{he2016deep}. Indeed, this is the insight behind the seminal work of~\cite{chen2018node}, who proposed the \gls{node} algorithm for parametrizing \glspl{ode} through \glspl{nn}. \glspl{cnf} are thus the intersection of \glspl{nf} with \gls{node}. The advantage of the approach of~\cite{chen2018node} is that, associated with eq.~\ref{eq:ode}, the log-probability follows an \gls{ode} as well,
\begin{align}
    \dfrac{\partial}{\partial t}\log|\det\nabla\mathbf{z}| &= \text{Tr}(\nabla_{\mathbf{z}}\mathbf{F}_{\theta}) =  \text{div}(\mathbf{F}_{\theta}).\label{eq:dynamics_log_likelihood}
\end{align}
With this workaround, minimizing the negative log-likelihood amounts to minimizing,
\begin{align}
    \sum_{i=1}^{n}-\log Q(\mathbf{z}_{T}(\mathbf{x}_{i}))-\int_{0}^{T}\text{div}(\mathbf{F}_{\theta}(\mathbf{z}_{t}(\mathbf{x}_{i})))dt,\label{eq:loss_ffjord}
\end{align}
which can be solved using the automatic differentiation~\cite{chen2018node}. One limitation of \glspl{cnf} is that the generic flow can be highly irregular, having unnecessarily fluctuating dynamics. As remarked by~\cite{finlay2020train}, this can pose difficulties to numerical integration. These issues motivated the proposition of two regularization terms for simpler dynamics. The first term is based dynamic \gls{ot}. Let,
\begin{align*}
    \rho_{t} = \mathbf{z}_{t,\sharp}P = \dfrac{1}{n}\sum_{i=1}^{n}\delta_{\mathbf{z}_{t}(\mathbf{x}_{i})},
\end{align*}
which simplifies eq.~\ref{eq:benamou_brenier_formulation} to,
\begin{equation}
\begin{aligned}
& \min{\theta}
& & \mathcal{L}(\theta) = \dfrac{1}{n}\sum_{i=1}^{n}\int_{0}^{T}\lVert\mathbf{F}_{\theta}(\mathbf{z}_{t}(\mathbf{x}_{i}))\rVert^{2}dt, \\
& \text{subject to}
& & \mathbf{z}_{0} = \mathbf{x}\text{, and } \mathbf{z}_{T,\sharp}P = Q,\\
\end{aligned}\label{eq:rnode}
\end{equation}
where $\lVert \mathbf{F}_{\theta}(\mathbf{z}_{t}(\mathbf{x}_{i})) \rVert^{2}$ is the kinetic energy of the particle $\mathbf{x}_{i}$ at time $t$, and $\mathcal{L}$ to the whole kinetic energy. In addition, $\rho_{t}$ has the properties of an \gls{ot} map. Among these, \gls{ot} enforces that: (i) the flow trajectories do not intersect, and (ii) the particles follow geodesics w.r.t. the ground cost (e.g., straight lines for an Euclidean cost), thus enforcing the regularity of the flow. Nonetheless, as~\cite{finlay2020train} remarks, these properties are enforced only on training data. To effectively generalize them, the authors propose a second regularization term, that consists on the Frobenius norm of the Jacobian,
\begin{align}
    \Omega(\mathbf{F}) = \dfrac{1}{n}\sum_{i=1}^{n}\lVert \nabla \mathbf{F}_{\theta}(\mathbf{z}_{T}(\mathbf{x}_{i}))\rVert_{F}^{2}.\label{eq:reg_jacobian}
\end{align}
Thus, the \gls{rnode} algorithm combines equation~\ref{eq:dynamics_log_likelihood} with the penalties in equations~\ref{eq:rnode} and~\ref{eq:reg_jacobian}. Ultimately, this is equivalent to minimizing the \gls{kl} divergence KL$(\mathbf{z}_{T,\sharp}P \lVert Q)$ with the said constraints.

\subsection{Dictionary Learning}\label{sec:representation_learning}

Dictionary learning~\cite{olshausen1997sparse} represents data $\mathbf{X} \in \mathbb{R}^{n \times d}$ through a set of $K$ atoms $\mathbf{D} \in \mathbb{R}^{k\times d}$ and $n$ representations $\mathbf{A} \in \mathbb{R}^{n\times k}$. For a loss $\mathcal{L}$ and a regularization term $\Omega$,
\begin{align}
    \argmin{\mathbf{D},\mathbf{A}}\dfrac{1}{n}\sum_{i=1}^{n}\mathcal{L}(\mathbf{x}_{i}, \mathbf{D}^{T}\mathbf{a}_{i}) + \lambda \Omega(\mathbf{D},\mathbf{A}).\label{eq:dictionary_learning}
\end{align}
In this setting, the data points $\mathbf{x}_{i}$ are approximated linearly through the matrix-vector product $\mathbf{D}^{T}\mathbf{a}_{i}$. In other words, $\mathbf{X} \approx \mathbf{AD}$. The practical interest is finding a faithful and sparse representation for $\mathbf{X}$. In this sense, $\mathcal{L}$ is often the Euclidean distance, and $\Omega(\mathbf{A})$ the $\ell_{1}$ or $\ell_{0}$ norm of vectors.

When the elements of $\mathbf{X}$ are non-negative, the dictionary learning problem is known as \gls{nmf}~\cite{paatero1994positive}. This scenario is particularly useful when data are histograms, that is, $\mathbf{x}_{i} \in \Delta_{d}$. As such, one needs to choose an appropriate loss function for histograms. \gls{ot} provides such a loss, through the Wasserstein distance~\cite{sandler2009nonnegative}, in which case the problem can be solved using \gls{bcd}: for fixed $\mathbf{D}$, solve for $\mathbf{A}$, and vice-versa.

Nonetheless, the \gls{bcd} iterations scale poorly since \gls{nmf} is equivalent to solving $n$ linear programs with $d^{2}$ variable. To circumvent this issue,~\cite{rolet2016fast} propose using the Sinkhorn algorithm~\cite{cuturi2013sinkhorn} for \gls{nmf}. Besides reducing complexity, using the Sinkhorn distance $W_{c, \epsilon}$ makes \gls{nmf} smooth, thus gradient descent methods can be applied successfully. The optimization problem of \gls{wnmf} is,
\begin{equation}
\argmin{\mathbf{D},\mathbf{A}} \dfrac{1}{N}\sum_{i=1}^{N}W_{p,\epsilon}(\mathbf{x}_{i},\mathbf{D}^{T}\mathbf{a}_{i})+\lambda\Omega(\mathbf{D},\mathbf{A}),\label{eq:wdl_nmf}
\end{equation}
subject to $\mathbf{AD}\in(\Delta_{d})^{N}$. Assuming $\mathbf{a}_{i} \in \Delta_{k}$ implies that $\mathbf{D}^{T}\mathbf{a}_{i}$ represents a weighted average of dictionary elements. This strategy can be understood as a barycenter under the Euclidean metric. Alternatively,~\cite{schmitz2018wasserstein} proposes to calculates barycenters on the Wasserstein space, that is,
\begin{align*}
\argmin{\mathbf{D},\mathbf{A}}\sum_{i=1}^{N}W_{p,\epsilon}(\mathcal{B}(\mathbf{a}_{i},\mathbf{D}), \mathbf{x}_{i}) + \lambda\Omega(\mathbf{D},\mathbf{A}).
\end{align*}
As a result,~\cite{schmitz2018wasserstein} perform a non-linear aggregation of atoms.

\begin{figure}[ht]
    \centering
    \includegraphics[width=\linewidth]{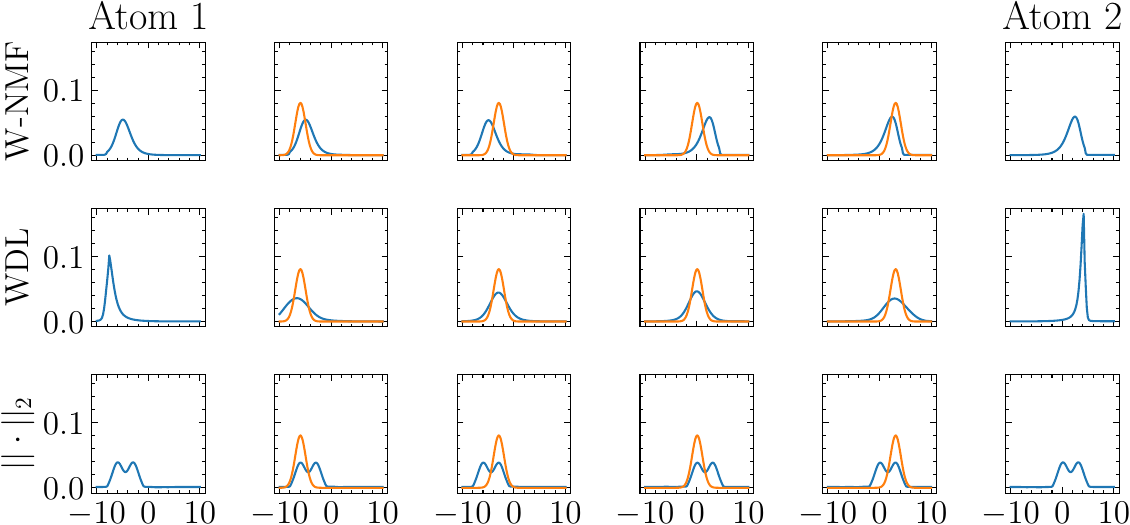}
    \caption{Comparison of \gls{wnmf}~\cite{rolet2016fast} (1st row), \gls{wdl}~\cite{schmitz2018wasserstein} (2nd row) and \gls{nmf} (3rd row) over histograms (1st row).}
    \label{fig:wdl}
\end{figure}

We show a comparison of these strategies in Figure~\ref{fig:wdl}, for a problem in which the histograms are Gaussian distributions $\mathbf{x}_{\ell} = \mathcal{N}(m_{\ell}, 1)$, $m_{\ell} = (1 - \ell/3)m_{0} + (\ell/3)m_{1}$, with $m_{0} = -6$, $m_{1} = 6$, and $\ell=0,\cdots,3$. As a result, the underlying set of histograms can be conveniently expressed in a Wasserstein space (see our discussion in sec.~\ref{sec:background}), which illustrates the advantage of \gls{wdl}. Generally, the advantage of \gls{wdl} is non-linearly interpolating atom distributions, allowing the learned dictionary to have more expressivity.

\noindent\textbf{Graph Dictionary Learning.} An interesting use case of this framework is \gls{gdl}. Let $\{G_{\ell}:(\mathbf{p}_{\ell},\mathbf{S}_{\ell}\}_{\ell=1}^{N}$ denote a dataset of $N$ graphs encoded by their node similarity matrices $\mathbf{S}_{\ell}\in\mathbb{R}^{n_{\ell}\times n_{\ell}}$, and histograms over nodes $\mathbf{p}_{\ell} \in \Delta_{n_{\ell}}$.~\cite{vincent2021online} proposes,
\begin{align*}
\argmin{\{\mathbf{S}_{k}\}_{k=1}^{K},\{\mathbf{a}_{\ell}\}_{\ell=1}^{N}}\sum_{\ell=1}^{N}\text{GW}_{2}\biggr(\mathbf{S}_{\ell},\sum_{k=1}^{K}a_{\ell,k}\mathbf{S}_{k}\biggr)^{2} - \lambda\lVert \mathbf{a}_{\ell} \rVert_{2}^{2},
\end{align*}
in analogy to eq.~\ref{eq:dictionary_learning}, using the \gls{gw} distance (see eq.~\ref{eq:gw_dist}) as loss function. Parallel to this line of work,~\cite{xu2020gromov} proposes \gls{gwf}, which approximates $\mathbf{S}^{(\ell)}$ non-linearly through \gls{gw}-barycenters~\cite{peyre2016gromov}.

\noindent\textbf{Topic Modeling.} Dictionary learning can be used for topic modeling~\cite{blei2003latent}, in which a set of documents $\{\hat{P}_{i}\}_{i=1}^{N}$ is expressed in terms of a dictionary of topics $\{\hat{Q}_{k}\}_{k=1}^{K}$ weighted by mixture coefficients. Indeed, documents can be interpreted as probability distributions over words in a vocabulary $V=\{w_{1},\cdots,w_{n}\}$~\cite{kusner2015word}. In this context~\cite{huynh2020otlda} proposes an \gls{ot}-inspired algorithm for learning a set of distributions $\mathcal{Q} = \{\hat{Q}_{k}\}_{k=1}^{K}$ over words, which represent \emph{a topic}. Learning is based on a hierarchical \gls{ot} problem,
\begin{align*}
    \min{\{\hat{Q}_{k}\}_{k=1}^{K}, \mathbf{b}, \gamma}&\sum_{k=1}^{K}\sum_{i=1}^{N} \gamma_{ik}W_{2,\epsilon}(\hat{Q}_{k},\hat{P}_{i}) - H(\gamma),
\end{align*}
where $\hat{Q}_{k} = \sum_{v=1}^{n}q_{k,v}\delta_{\mathbf{x}_{v}^{(Q_{k})}}$, and $\mathbf{b} = [b_{1}, \cdots, b_{K}]$ is the coefficient vector of topics. In this sense $\sum_{i}\gamma_{ik} = b_{k}$, and $\sum_{k}\gamma_{ik}=p_{i}$, with $p_{i}=\nicefrac{n_{i}}{n}$, i.e., the proportion of words in document $\hat{P}_{i}$. Note that, when topics $\mathcal{Q}$ are computed, one can compute hierarchical distances between documents $P_{i}$ and $P_{j}$ through the \gls{hott} distance~\cite{yurochkin2019hierarchical},
\begin{align*}
    \text{HOTT}(\hat{P}_{i},\hat{P}_{j}) &= W_{1}\biggr(\sum_{k=1}^{K}d_{ik}\delta_{\hat{Q}_{k}},\sum_{k=1}^{K}d_{jk}\delta_{\hat{Q}_{k}}\biggr),
\end{align*}
where $\mathbf{d}_{i},\mathbf{d}_{j} \in \Delta_{K}$ are document distributions over topics and the ground-cost is the \gls{wmd}~\cite{kusner2015word}, hence the term \emph{hierarchical}.

\subsection{Clustering}\label{sec:clustering}

Clustering is a problem within unsupervised learning that deals with the aggregation of features into groups~\cite{bishp2006pattern}. From the perspective of \gls{ot}, this is linked to the notion of quantization~\cite{pollard1982quantization,canas2012learning}; Indeed, from a distributional viewpoint, given $\mathbf{X}^{(P)} \in \mathbb{R}^{n\times d}$, quantization corresponds to finding the matrix $\mathbf{X}^{(Q)}\in\mathbb{R}^{K\times d}$ minimizing $W_{2}(P, Q)$. This has been explored in a number of works, such as~\cite{gruber2004optimum,graf2007foundations} and~\cite{cuturi2014fast}. In this sense, \gls{ot} serves as a framework for quantization/clustering, thus allowing for theoretical results such as convergence bounds for the $K$-means algorithm~\cite{canas2012learning}, as well as extensions~\cite{del2019robust}.

\noindent\textbf{Co-Clustering.} While standard clustering can be seen as a method for grouping the rows of a matrix $\mathbf{X} \in \mathbb{R}^{n \times d}$, co-clustering seeks to group rows and columns simultaneously. \gls{ot} contributed to this setting in 2 ways~\cite{laclau2017coclustering}. First, the \gls{ccot} strategy relies on row and column distributions,
\begin{align*}
    \hat{P} = \dfrac{1}{n}\sum_{i=1}^{n}\delta_{\mathbf{r}_{i}^{(P)}}\text{, and, }\hat{Q} = \dfrac{1}{d}\sum_{j=1}^{d}\delta_{\mathbf{c}_{j}^{(Q)}},
\end{align*}
where $\mathbf{r}_{i}^{(P)} = [x_{1,i},\cdots,x_{n,i}] \in \mathbb{R}^{n}$ and $\mathbf{c}_{j}^{(Q)} = [x_{j,1},\cdots,x_{j,d}] \in \mathbb{R}^{d}$. Note that, as discussed in~\cite[Section 3.6]{laclau2017coclustering}, \gls{ot} is only defined for $n \geq d$, in which case one can sub-sample the rows of $\mathbf{X}$ for defining a $d \times d$ \gls{ot} problem. Clustering over rows and columns is done by jump detection~\cite{matei2012nonlinear} on vectors $(\mathbf{f},\mathbf{g})$ defined by the Sinkhorn algorithm (cf. eq.~\ref{eq:sinkhorn_optimality}).

Second, the \gls{ccot}-\gls{gw} strategy uses the Gromov extension of \gls{ot} (cf. eq.~\ref{eq:gw_dist}) between kernel matrices $K_{r} \in \mathbb{R}^{n \times n}$, $K_{c} \in \mathbb{R}^{d \times d}$ and $K \in \mathbb{R}^{p \times p}$, for an hyper-parameter $p \in \mathbb{N}$. The matrix $K$ is the \gls{gw} barycenter~\cite{peyre2016gromov} of $K_{r}$ and $K_{c}$, and $K_{r}, K_{c}$ capture the intra-rows and intra-columns similarities. Clustering is done similarly to \gls{ccot}, i.e., one solves 2 entropic regularized Gromov \gls{ot} problems, between $K_{r}$ and $K$, and between $K$ and $K_{c}$.

\noindent\textbf{\glspl{gmm}} are a kind of probabilistic model defined by $\theta = \{\beta_{k},\mathbf{m}_{k},\mathbf{S}_{k}\}_{k=1}^{K}$, where,
\begin{align}
    P_{\theta}(\mathbf{x}) = \sum_{k=1}^{K}\beta_{k}\mathcal{N}(\mathbf{x}|\mathbf{m}_{k},\mathbf{S}_{k}),\label{eq:gmm}
\end{align}
where $\sum\beta_{k}=1$, $\beta_{k} \geq 0$. From the perspective of clustering, each $(\beta_{k},\mathbf{m}_{k},\mathbf{S}_{k})$ represents a group of data points. From eq.~\ref{eq:gmm} one can get soft-assignments for $\mathbf{x}$ using,
\begin{align*}
    \alpha_{k}(\mathbf{x}) = \dfrac{\beta_{k} \mathcal{N}(\mathbf{x}|\mathbf{m}_{k},\mathbf{S}_{k})}{\sum_{j}\beta_{j} \mathcal{N}(\mathbf{x}|\mathbf{m}_{j},\mathbf{S}_{j})},
\end{align*}
which draws a parallel between clustering and generative modeling. Learning a \gls{gmm} is usually done via maximum likelihood, which ultimately amounts to minimizing $\text{KL}(P_{0}|P_{\theta})$, where $P_{0}$ represents the data distribution In this context~\cite{kolouri2018sliced} proposed learning $P_{\theta}$ by minimizing $\text{SW}_{p}(P_{0}, P_{\theta})$ (see sec~\ref{sec:projbased}). This problem can be nicely written in closed form, due to the properties of Gaussian distributions. Let $P = \mathcal{N}(\mathbf{m},\mathbf{S})$, then~\cite{kolouri2018sliced} shows $\pi_{\mathbf{u},\sharp}P = \mathcal{N}(\langle \mathbf{u}, \mathbf{m} \rangle, \mathbf{u}^{T}\mathbf{S}\mathbf{u})$. As shown in~\cite{kolouri2018sliced} \gls{sw}-based \glspl{gmm} are robust to parameter initialization.
\section{Transfer Learning}\label{sec:transfer-learning}

\gls{tl} is a \gls{ml} framework, concerned with learning scenarios in which data follows different probability distributions. Following~\cite{pan2009survey}, \gls{tl} can be formalized through the notions of domain and task. In the first case, \emph{a domain} is a pair $\mathcal{D} = (\mathcal{X},P(X))$ of a feature space (e.g., $\mathbb{R}^{d}$) and a feature marginal distribution. In the second case, \emph{a task} is a pair $\mathcal{T} = (\mathcal{Y},P(Y|X))$ of a label space (e.g., $\{1,\cdots,n_{c}\}$) and a conditional distribution $P(Y|X)$. Given a source $(\mathcal{D}_{S},\mathcal{T}_{S})$ and a target $(\mathcal{D}_{T},\mathcal{T}_{T})$, the goal of \gls{tl} is \emph{improving performance on the target, given knowledge from the source}.

The \emph{distributional shift} occurring in \gls{tl} can be modeled via $P_{S}(X,Y) \neq P_{T}(X,Y)$ for a source $S$ and a target $T$. Since $P(X, Y) = P(X)P(Y|X) = P(Y)P(X|Y)$, three types of shift may occur~\cite{redko2019advances}: (i) \emph{covariate shift}, that is, $P_{S}(X) \neq P_{T}(X)$, (ii) \emph{concept shift}, namely, $P_{S}(Y|X)\neq P_{T}(Y|X)$ or $P_{S}(X|Y) \neq P_{T}(X|Y)$, and (iii) \emph{target shift}, for which $P_{S}(Y)\neq P_{T}(Y)$. In the following, we focus on \gls{uda}, a setting in which one has access to labeled data from the source domain, and unlabeled data from the target domain.

\subsection{Shallow Domain Adaptation}\label{sec:shallow_da}

\begin{figure}[ht]
    \centering
    \includegraphics[width=0.8\linewidth]{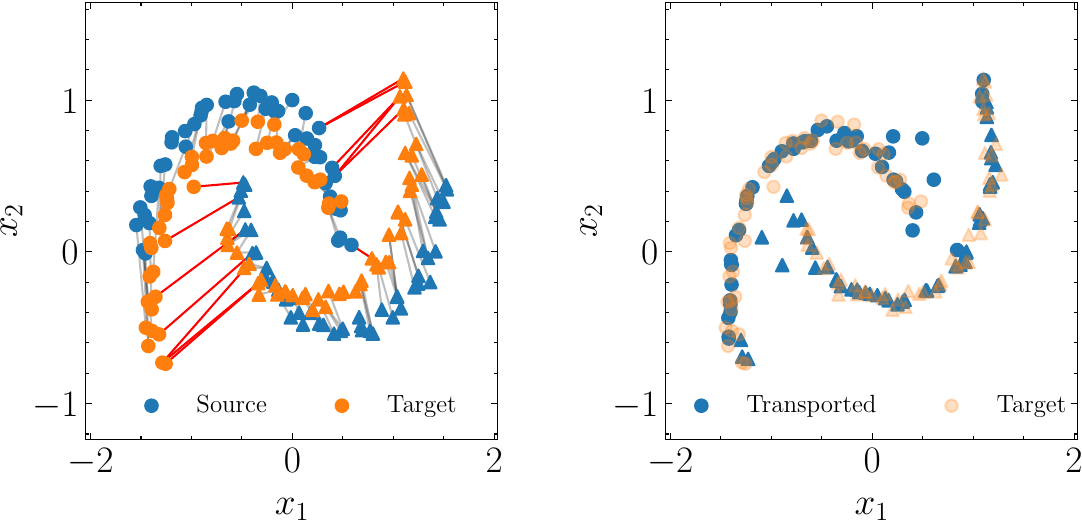}
    \caption{\gls{otda} methodology: (a) Source (red) and target (blue) samples follow different distributions; A classifier (dashed black line) fit on source is not adequate for target; (b) $T_{\gamma}$ then transports source domain samples to the target domain.}
    \label{fig:otda}
\end{figure}

Under covariate shift, \gls{ot} can be used for matching $P_{S}(X)$ and $P_{T}(X)$. This is straightforward with the Monge problem, in which $T_{\sharp}P_{S}=P_{T}$, but its solution may not exist in the discrete setting (e.g., section~\ref{sec:computational_ot}). Given this shortcoming,~\cite{courty2017otda} uses the barycentric mapping,
\begin{align}
    T_{\gamma}(\mathbf{x}^{(P_{S})}_{i}) = \argmin{\mathbf{x}\in\mathbb{R}^{d}}\sum_{j=1}^{n_{T}}\gamma_{ij}c(\mathbf{x},\mathbf{x}^{(P_{T})}_{j}),\label{eq:barycentric_projection}
\end{align}
where $\gamma = \text{OT}(\mathbf{p}_{S},\mathbf{p}_{T},\mathbf{C})$. The case where $c(\mathbf{x}_{1},\mathbf{x}_{2})=\lVert \mathbf{x}_{1} - \mathbf{x}_{2} \rVert^{2}_{2}$ is particularly interesting, as $\mathbf{T}_{\gamma}$ has closed-form in terms of the support $\mathbf{X}^{(P_{T})} \in \mathbb{R}^{n_{T}\times d}$ of $\hat{P}_{T}$, $\hat{\mathbf{X}}^{(P_{S})} = \text{diag}(\mathbf{p}_{S})^{-1}\gamma\mathbf{X}^{(P_{T})}$. As follows, each point $\mathbf{x}_{i}^{(P_{S})}$ is mapped into $\hat{\mathbf{x}}_{i}^{(P_{S})}$, which is a convex combination of the points $\mathbf{x}_{j}^{(P_{T})}$ that receives mass from $\mathbf{x}_{i}^{(P_{S})}$, namely, $\gamma_{ij} > 0$. This effectively generates a new training dataset $\{(\hat{\mathbf{x}}_{i}^{(P_{S})}, y_{i}^{(P_{S})})\}_{i=1}^{n_{S}}$, which hopefully leads to a classifier $\hat{h}$ that works well on $\mathcal{D}_{T}$. An illustration is shown in Figure~\ref{fig:otda}.

\begin{wrapfigure}{r}{0.4\linewidth}
    \centering
    \includegraphics[width=\linewidth]{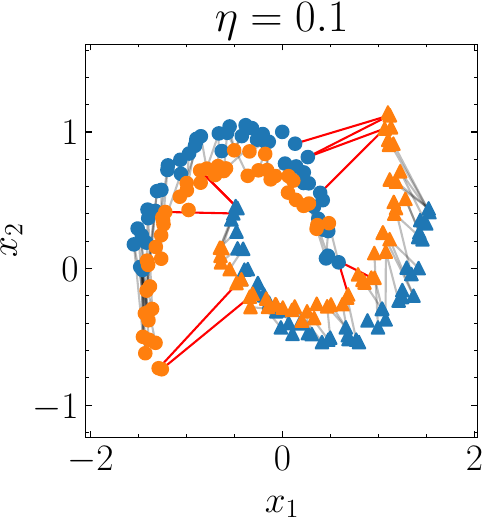}
    \caption{Optimal transport with class regularization induces a match between $\hat{P}_{S}$ and $\hat{P}_{T}$ that tends to be more class sparse.}
    \label{fig:ot-plan-regularized}
\end{wrapfigure}

The barycentric mapping has two limitations. First, it may map points to the wrong side of the decision boundary. As noted by~\cite{courty2017otda}, this happens when $\gamma_{ij}$ moves mass between different classes. To avoid that,~\cite{courty2017otda} proposes to further regularize \gls{ot} plans:
\begin{align*}
    \gamma^{\star} = \argmin{\gamma\in U(\mathbf{p}_{S},\mathbf{p}_{T})}\langle \mathbf{C}, \gamma \rangle_{F} - \epsilon H(\gamma) + \eta \Omega(\gamma;\mathbf{y}^{(P_{S})},\mathbf{y}^{(P_{T})}),
\end{align*}

\noindent where $\Omega$ is a class-based regularizer. This additional regularization enforces that $\gamma_{ij} > 0$ if and only if $\mathbf{x}_{i}^{(P_{S})}$ and $\mathbf{x}_{j}^{(P_{T})}$ have the same class. We highlight the wrong pairings done by \gls{ot} in Figs.~\ref{fig:otda} and~\ref{fig:ot-plan-regularized}. The regularization strategies are divided into semi-supervised and unsupervised. In the former case, one uses a few labeled samples in the target domain to penalize $\{\gamma_{ij} > 0,y_{i}^{(P_{S})} \neq y_{j}^{(P_{T})}\}$. In the latter case, one penalizes transport plans such that $\mathbf{x}_{j}^{(P_{T})}$ receives mass from $\mathbf{x}_{i}^{(P_{S})}$ with different classes. This is achieved through group-sparsity~\cite[Eq. 17]{courty2017otda}.

Second, through eq.~\ref{eq:barycentric_projection}, $T_{\gamma}$ it is only defined on the samples $\mathbf{x}_{i}^{(P_{S})}$ in $\mathbf{X}^{(P_{S})}$. It is thus undefined for new samples coming from $P_{S}$. There are three strategies to solve this issue. First,~\cite{ferradans2014regularized} proposes to define $T_{\gamma}$ for new points $\mathbf{x}^{(P_{S})}$ through its nearest neighbors. This, however, introduces quantization effects on the mapping. This issue motivates~\cite{perrot2016mapping} to parametrize $T_{\gamma}$ as a linear, or a kernel mapping. Finally,~\cite{seguy2018large} proposes a large scale strategy in which the barycentric mapping is parametrized through a neural net (see section~\ref{sec:icnn_ot}).

In a more general direction, one may assume that \emph{the shift occurs at the level of joint distributions}. In \gls{uda} this is not directly possible, since the labels from $P_{T}(X, Y)$ are not available. A workaround was introduced in~\cite{courty2017joint}, where the authors propose a proxy distribution $P_{T}^{h}(X,h(X))$:
\begin{align*}
    \hat{P}_{T}^{h} = \dfrac{1}{n_{T}}\sum_{j=1}^{n_{T}}\delta_{\mathbf{x}^{(P_{T})}_{j},h(\mathbf{x}^{(P_{T})}_{j})}.
\end{align*}
where $h\in\mathcal{H}$ is a classifier. As follows,~\cite{courty2017joint} propose minimizing the Wasserstein distance $W_{c}(\hat{P}_{S}, \hat{P}_{T}^{h})$ over possible classifiers $h \in \mathcal{H}$. This yields a joint minimization problem,
\begin{align*}
    \min{\substack{h\in\mathcal{H}\\ \gamma\in U(\mathbf{p}_{S},\mathbf{p}_{T})}}\sum_{i=1}^{n_{S}}\sum_{j=1}^{n_{T}}c(\mathbf{x}^{(P_{S})}_{i},y^{(P_{S})}_{i},\mathbf{x}^{(P_{T})}_{j})\gamma_{ij} + \lambda\Omega(h).
\end{align*}
The cost $c$ can be designed in terms of two factors, a feature loss $c_{f}(\mathbf{x}^{(P_{S})}_{i},\mathbf{x}^{(P_{T})}_{j})$ (e.g., the Euclidean distance), and a label loss $c_{\ell}(y^{(P_{S})}_{i},h(\mathbf{x}^{(P_{T})}_{j}))$ (e.g., the Hinge loss). As proposed in~\cite{courty2017joint}, the importance of $c_{f}$ and $c_{\ell}$ can be controlled by a parameter $\alpha$,
\begin{align*}
    C_{ij} &= \alpha c_{f}(\mathbf{x}_{i}^{(P_{S})},\mathbf{x}_{j}^{(P_{T})}) + c_{\ell}(y_{i}^{(P_{S})},h(\mathbf{x}_{j}^{(P_{T})})).
\end{align*}

\subsection{Deep Domain Adaptation}\label{sec:deep_da}

In this section, we assume that a deep \gls{nn} is a composition $f = h_{\xi} \circ g_{\theta}$ of a feature extractor $g$ with parameters $\theta$, and a classifier $h_{\xi}$ with parameters $\xi$. The intuition behind deep \gls{da} is to force the feature extractor to \emph{learn domain invariant features}. Given $\mathbf{x}_{i}^{(P_{S})} \sim P_{S}$ and $\mathbf{x}_{j}^{(P_{T})} \sim P_{T}$, an invariant feature extractor $g_{\theta}$ should suffice,
\begin{align*}
    (g_{\theta})_{\sharp}\hat{P}_{S} = \dfrac{1}{n_{S}}\sum_{i=1}^{n_{S}}\delta_{\mathbf{z}_{i}^{(P_{S})}} \overset{\mathcal{D}}{\approx} \dfrac{1}{n_{T}}\sum_{j=1}^{n_{T}}\delta_{\mathbf{z}_{j}^{(P_{T})}} = (g_{\theta})_{\sharp}\hat{P}_{T},
\end{align*}
where $\mathbf{z}_{i}^{(P_{S})} = g_{\theta}(\mathbf{x}_{i}^{(P_{S})})$ (resp. $\hat{P}_{T}$), and $\hat{P}_{S} \overset{\mathcal{D}}{\approx} \hat{P}_{T}$ means $\mathcal{D}(\hat{P}_{S},\hat{P}_{T}) \approx 0$ for a given divergence or distance between distributions. This implies that, after the application of $g_{\theta}$, distributions $\hat{P}_{S}$ and $\hat{P}_{T}$ are close. In this sense, this condition is enforced by adding an additional term to the classification loss function (e.g., the \gls{mmd} as in~\cite{ghifary2014domain}). In this context,~\cite{ganin2016domain} proposes the \gls{dann} algorithm, based on the loss function,
\begin{align*}
    \mathcal{L}_{\text{DANN}}(\theta,\xi,\eta) &= \hat{\mathcal{R}}_{P_{S}}(h_{\xi} \circ g_{\theta}) - \lambda \mathcal{L}_{\mathcal{H}}(\theta,\eta),
\end{align*}
where $\mathcal{R}_{P_{S}}$ denotes the empirical risk (see eq.~\ref{eq:empirical_risk}) and $\mathcal{L}_{\mathcal{H}}$,
\begin{align*}
\dfrac{1}{n_{S}}\sum_{i=1}^{n_{S}}\log h_{\eta}(\mathbf{z}_{i}^{(P_{S})}))+\dfrac{1}{n_{T}}\sum_{j=1}^{n_{T}}\log (1-h_{\eta}(\mathbf{z}_{j}^{(P_{T})})),
\end{align*}
where $h_{\eta}$ is a supplementary classifier, that \emph{discriminates} between source (labeled $0$) and target (labeled $1$). The \gls{dann} algorithm is a mini-max optimization problem,
\begin{align*}
\min{\theta,\xi}\max{\eta}\mathcal{L}_{\text{DANN}}(\theta,\xi,\eta),
\end{align*}
so as to minimize classification error and \emph{maximize domain confusion}. This draws a parallel with the \gls{gan} algorithm of~\cite{goodfellow2014generative}, presented in section~\ref{sec:gans}. This remark motivated~\cite{shen2018wasserstein} for using the Wasserstein distance instead of $\mathcal{L}_{\mathcal{H}}$. Their algorithm is called \gls{wdgrl}, which uses as loss:
\begin{align*}
    \mathcal{L}_{W}(\theta,\eta) &= \dfrac{1}{n_{S}}\sum_{i=1}^{n_{S}}h_{\eta}(g_{\theta}(\mathbf{x}^{(P_{S})}_{i})) - \dfrac{1}{n_{T}}\sum_{j=1}^{n_{T}}h_{\eta}(g_{\theta}(\mathbf{x}^{(P_{T})}_{j})).
\end{align*}
Following our discussion on \gls{kr} duality (see section~\ref{sec:background}) as well as the Wasserstein \gls{gan} (see section~\ref{sec:gans}), one needs to maximize $\mathcal{L}_{W}$ over $h_{\eta}\in\text{Lip}_{1}$.~\cite{shen2018wasserstein} proposes doing so using the gradient penalty term of~\cite{gulrajani2017improved},
\begin{align*}
    \min{\theta}\min{\xi} \hat{\mathcal{R}}_{P_{S}}(h_{\xi} \circ g_{\theta})+\lambda_{2}\max{\eta}\mathcal{L}_{W}(\theta,\eta)-\lambda_{1}\Omega(h_{\eta}),
\end{align*}
where $\lambda_{1}$ controls the gradient penalty term, and $\lambda_{2}$ controls the importance of the domain loss term.

Finally, we highlight that the \gls{jdot} framework can be extended to deep \gls{da}. First, one includes the feature extractor $g_{\theta}$ in the ground-cost. For $\mathbf{z}_{i}^{(P_{S})} = g_{\theta}(\mathbf{x}_{i}^{(P_{S})})$ (resp. $P_{T}$),
\begin{align*}
    C_{ij}(\theta,\xi) &= \alpha \lVert \mathbf{z}_{i}^{(P_{S})} - \mathbf{z}_{j}^{(P_{T})} \rVert^{2} + c_{\ell}(y_{i}^{(P_{S})},h_{\xi}(\mathbf{z}_{j}^{(P_{T})})),
\end{align*}
second, the objective function includes a classification loss $\mathcal{R}_{P_{S}}$, and the \gls{ot} loss, that is,
\begin{align}
    \mathcal{L}_{JDOT}(\theta,\xi) &= \hat{\mathcal{R}}_{P_{S}}(h_{\xi} \circ g_{\theta}) + \sum_{i=1}^{n_{S}}\sum_{j=1}^{n_{T}}\gamma_{ij}C_{ij}(\theta,\xi),\label{eq:loss_jdot}
\end{align}
which is jointly minimized w.r.t. $\gamma \in \mathbb{R}^{n_{S} \times n_{T}}$, $\theta$ and $\xi$. \cite{damodaran2018deepjdot} proposes minimizing $\mathcal{L}_{\text{JDOT}}$ jointly w.r.t. $\theta$ and $\xi$, using mini-batches (see section~\ref{sec:minibatch_ot}). Nonetheless, the authors noted that one needs large mini-batches for a stable training. To circumvent this issue~\cite{fatras2021unbalanced} proposed using \emph{unbalanced \gls{ot}} (see sec.~\ref{sec:ot_ext}), allowing for smaller batch sizes and improved performance.

\begin{figure}
    \centering
    \subfloat[Case I]{\includegraphics[width=0.23\linewidth]{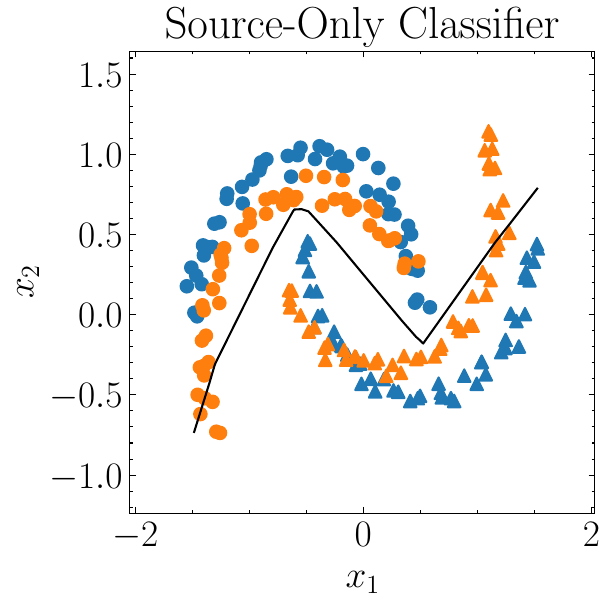}}
    \subfloat[Case II]{\includegraphics[width=0.23\linewidth]{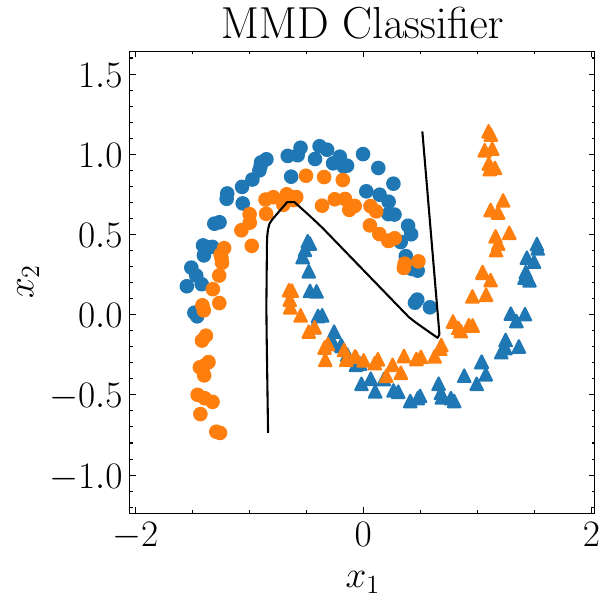}}
    \subfloat[Case III]{\includegraphics[width=0.23\linewidth]{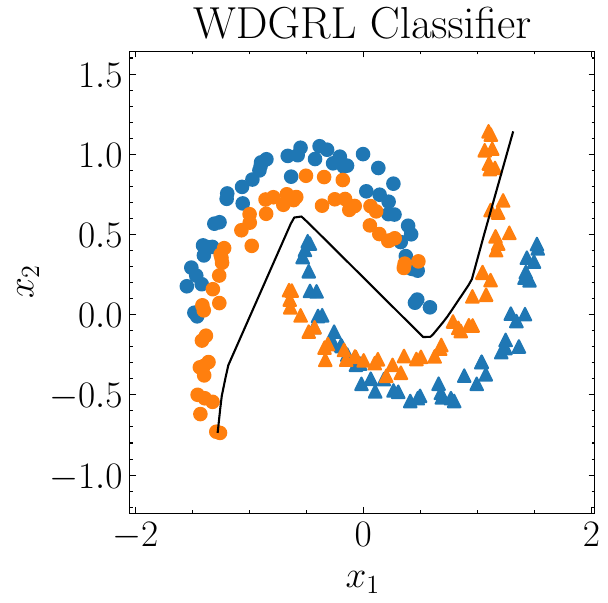}}
    \subfloat[Case IV]{\includegraphics[width=0.23\linewidth]{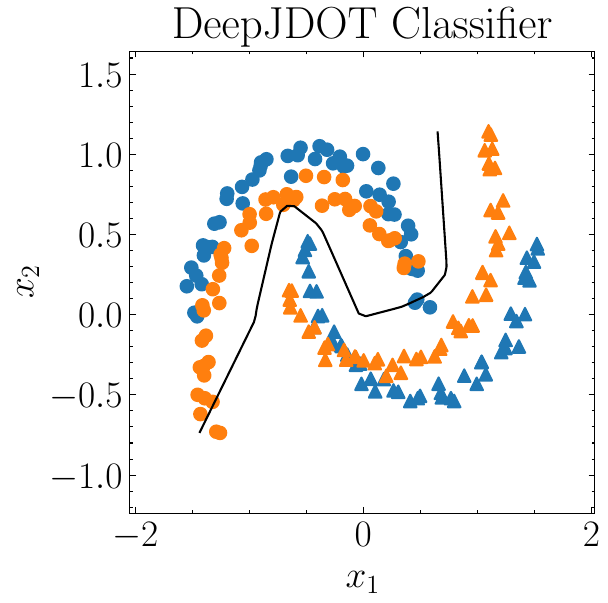}}\\
    \subfloat[Case I]{\includegraphics[width=0.23\linewidth]{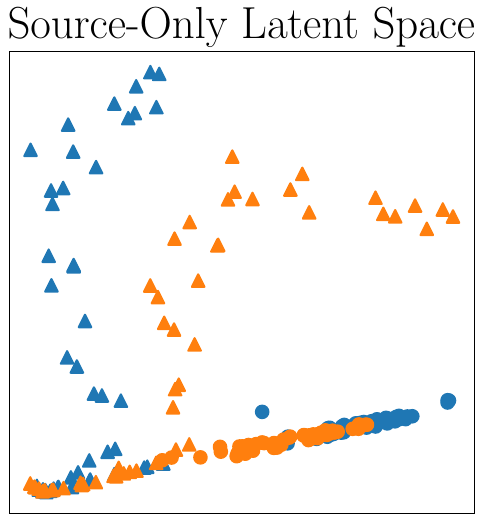}}
    \subfloat[Case II]{\includegraphics[width=0.23\linewidth]{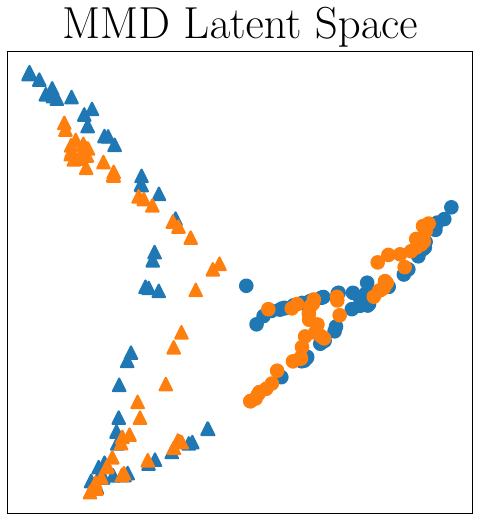}}
    \subfloat[Case III]{\includegraphics[width=0.23\linewidth]{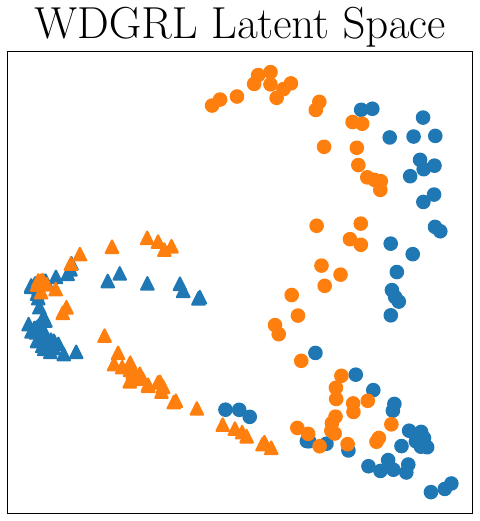}}
    \subfloat[Case IV]{\includegraphics[width=0.23\linewidth]{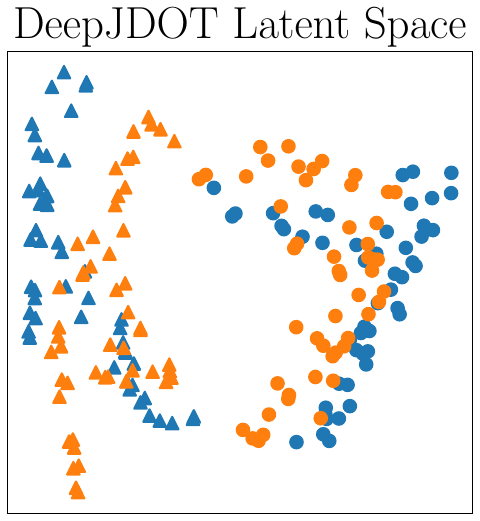}}
    \caption{Comparison of Deep \gls{da} strategies, based on the \gls{mmd} (b, f) and \gls{ot} (c, d, g, h). Overall, (e -- h) show the PCA (2 components) of the latent space of a neural net. Note that deep \gls{da} match $\hat{P}_{S}$ and $\hat{P}_{T}$ in such space. As a result, they are able to find a classifier that correctly predicts on target domain samples.}
    \label{fig:deep-da-example}
\end{figure}

\subsection{Extensions to Domain Adaptation}\label{sec:general_da}

\noindent\textbf{\gls{msda}} considers the problem of \gls{da} when source data comes from multiple, distributionally heterogeneous domains, namely $P_{S_{1}},\cdots,P_{S_{K}}$. In this sense,~\cite{montesuma2021wbt} proposes using the Wasserstein barycenter for building an intermediate domain $\hat{P}_{B} = \mathcal{B}(\alpha;\{\hat{P}_{S_{k}}\}_{k=1}^{N_{S}})$ (see equation~\ref{eq:wbary}), for $\alpha_{k} = \nicefrac{1}{N_{S}}$. Since $\hat{P}_{B} \neq \hat{P}_{T}$, the authors propose using an additional adaptation step for transporting the barycenter towards the target. Furthermore,~\cite{turrisi2022multi} proposes the Weighted \gls{jdot} algorithm, which generalizes the work of~\cite{courty2017joint} for \gls{msda},
\begin{align*}
    \min{\alpha,g_{\theta},h_{\xi}}\dfrac{1}{k}\sum_{k=1}^{K}\hat{\mathcal{R}}_{P_{S_{k}}}(h_{\xi} \circ g_{\theta})+\mathcal{T}_{c_{\theta,\xi}}(\hat{P}_{T}^{h_{\xi}},\sum_{k=1}^{K}\alpha_{k}\hat{P}_{S_{k}}).
\end{align*}

\noindent\textbf{\gls{hda}} is a \gls{da} problem in which source and target domains are incomparable, that is, $\mathbf{X}^{(P_{S})} \in \mathbb{R}^{n_{S}\times d_{S}}$ and $\mathbf{X}^{(P_{T})} \in \mathbb{R}^{n_{T}\times d_{T}}$, $n_{S} \neq n_{T}$ and $d_{S} \neq d_{T}$. To that end,~\cite{titouan2020co} relies on the \gls{gw} \gls{ot} formalism (see sec.~\ref{sec:ot_ext}),
\begin{align*}
    (\gamma^{(s)},\gamma^{(f)}) = \min{\substack{\gamma^{(s)}\in \Gamma(\mathbf{w}_{S},\mathbf{w}_{T})\\\gamma^{(f)}\in \Gamma(\mathbf{v}_{S},\mathbf{v}_{T})}}\sum_{i,j,k,l}L(\mathbf{X}_{i,k}^{(P_{S})},\mathbf{X}_{j,l}^{(P_{T})})\gamma_{ij}^{(s)}\gamma_{ij}^{(f)},
\end{align*}
where $\gamma^{(s)} \in \mathbb{R}^{n_{S} \times n_{T}}$ and $\gamma^{(f)} \in \mathbb{R}^{d_{S}\times d_{T}}$ are \gls{ot} plans between \emph{samples} and \emph{features}, respectively. Using $\gamma^{(s)}$, one can estimate labels on the target domain using label propagation~\cite{redko2019optimal},
$\hat{\mathbf{Y}}^{(P_{T})}=\text{diag}(\mathbf{p}_{T})^{-1}(\gamma^{(s)})^{T}\mathbf{Y}^{(P_{S})}$.

\noindent\textbf{Transferability} concerns the task of predicting whether \gls{tl} will be successful. From a theoretical perspective, different \glspl{ipm} bound the target risk $\mathcal{R}_{P_{T}}$ by the source risk $\mathcal{R}_{P_{S}}$, such as the $\mathcal{H}$, Wasserstein and \gls{mmd} distances (see e.g.,~\cite{ben2010theory,redko2017theoretical,redko2019advances}). Furthermore there is a practical interest in accurately measuring transferability \emph{a priori}, before training or fine-tuning a network. A candidate measure coming from \gls{ot} is the Wasserstein distance, but in its original formulation it only takes features into account.~\cite{alvarez2020geometric} propose the \gls{otdd}, which relies on a Wasserstein-distance based metric on the label space,
\begin{align*}
    \text{OTDD}(P_{S},P_{T}) &= \argmin{\gamma \in \Gamma(P_{S},P_{T})}\expectation{(\mathbf{z}_{1},\mathbf{z}_{2}) \sim \gamma}[c(\mathbf{z}_{1},\mathbf{z}_{2})],\\
    c(\mathbf{z}_{1},\mathbf{z}_{2}) &= \lVert \mathbf{x}_{1}-\mathbf{x}_{2} \rVert_{2}^{2}+W_{2}(P_{S,y_{1}},P_{T,y_{2}})^{2},
\end{align*}
where $\mathbf{z}=(\mathbf{x},y)$, and $P_{S,y}$ is the conditional distribution $P_{S}(X|Y=y)$. As the authors show in~\cite{alvarez2020geometric}, this distance is highly correlated with transferability and can even be used to interpolate between datasets with different classes~\cite{alvarez2021dataset}.

\section{Reinforcement Learning}\label{sec:reinf_learning}

\gls{rl} deals with dynamic learning scenarios and sequential decision-making. Following~\cite{sutton2018reinforcement}, one assumes an environment modeled through a \gls{mdp}, which is a 5-tuple $(\mathcal{S}, \mathcal{A}, P, R, \rho_{0}, \lambda)$ of a state space $\mathcal{S}$, an action space $\mathcal{A}$, a transition distribution $P:\mathcal{S}\times\mathcal{A}\times\mathcal{S}\rightarrow\mathbb{R}$, a reward function $R:\mathcal{S}\times\mathcal{A}\rightarrow\mathbb{R}$, a distribution over the initial state $\rho_{0}$, and a discount factor $\lambda \in (0,1)$.

\gls{rl} revolves around an agent acting on the state space, changing the state of the environment. The actions are chosen according to a policy $\pi:\mathcal{S}\rightarrow\mathbb{R}$, which is a distribution $\pi(\cdot|s)$ over actions $a \in \mathcal{A}$, for $s \in \mathcal{S}$. Policies can be evaluated according their average returns,
\begin{align}
    \eta(\pi) &= \expectation{P,\pi}\biggr[\sum_{t=0}^{\infty}\lambda^{t}R(s_{t},a_{t})\biggr].\label{eq:utility}
\end{align}
Furthermore, under $\pi$, one can evaluate states and state-action pairs through $V_{\pi}$ and $Q_{\pi}$,
\begin{equation}
\begin{aligned}
    V_{\pi}(s) &= \expectation{P,\pi}\biggr[\sum_{\ell=0}^{\infty}\lambda^{\ell}R(s_{t+\ell},a_{t+\ell})|s_{t}=s\biggr],\\
    Q_{\pi}(s,a) &= \expectation{P,\pi}\biggr[\sum_{\ell=0}^{\infty}\lambda^{\ell}R(s_{t+\ell},a_{t+\ell})|s_{t}=s,a_{t}=a\biggr],
\end{aligned}\label{eq:rl_def}
\end{equation}
where the expectation over $P$ and $\pi$ corresponds to $s_{0} \sim \rho_{0} $, $a_{t} \sim \pi(\cdot|s_{t})$, and $s_{t+1} \sim P(\cdot|s_{t},a_{t})$. These quantities are related through Bellman's equation~\cite{bellman1952theory},
\begin{equation}
\begin{aligned}
    \mathcal{T}_{\pi}V_{\pi}(s_{t}) &= \expectation{P, \pi}[R(s_{t},a_{t})] + \gamma \expectation{P, \pi}[V_{\pi}(s)],\\
    \mathcal{T}_{\pi}Q_{\pi}(s_{t}, a_{t}) &= \expectation{P,\pi}[R(s_{t},a_{t})] + \gamma \expectation{P, \pi}[Q_{\pi}(s, a)],
\end{aligned}\label{eq:bellman_eq}
\end{equation}
where $\mathcal{T}_{\pi}$ is called Bellman operator. $Q_{\pi}$ can be learned \emph{from experience} (i.e., tuples $(s,a,r,s')$) through Q-Learning~\cite{watkins1992q}. For a learning rate $\alpha_{t} > 0$, the updates are as follows,
\begin{align}
    Q_{\pi}(s,a) &\leftarrow (1-\alpha_{t})Q_{\pi}(s,a)+\alpha_{t}\biggr(r+\lambda V_{\pi}(s')\biggr).\label{eq:q-learning}
\end{align}

\subsection{Distributional Reinforcement Learning}\label{sec:distributional_rl}

\gls{drl}~\cite{white1988mean,morimura2010parametric,azar2012sample} differs from traditional \gls{rl} by considering uncertainty over returns. In this context,~\cite{bellemare2017distributional} proposed studying the random return $Z_{\pi}$, such that $Q_{\pi}(s_{t},a_{t}) = \expectation{P,\pi}[Z_{\pi}(s_{t},a_{t})]$, where the expectation is taken in the sense of equations~\ref{eq:rl_def}.~\cite{bellemare2017distributional} defines \gls{drl} by analogy with equation~\ref{eq:bellman_eq},
\begin{align*}
    Z_{\pi}(s_{t},a_{t}) \overset{D}{=} R(s_{t},a_{t}) + \lambda Z_{\pi}(s_{t+1}, a_{t+1}),
\end{align*}
where $\overset{D}{=}$ means equality in distribution. Note that $Z_{\pi}$ is a distribution over returns, hence a distribution over $\mathbb{R}$. In \gls{drl}, one has mainly two choices for parametrizing $Z_{\pi}$, which rely on an empirical approximation,
\begin{align*}
    Z_{\theta}(x) = \sum_{i=1}^{n}z_{i}\delta(x-x_{i}),
\end{align*}
where one focuses either on $z_{i}$~\cite{bellemare2017distributional}, or the $x_{i}$~\cite{dabney2018distributional}. The optimal $\theta$ is found by minimizing a notion of discrepancy between $Z_{\theta}$ and $T_{\pi}Z_{\theta}$. \gls{ot} contributes to \gls{drl} precisely at this point: one uses $W_{p}(Z_{\theta}, T_{\pi}Z_{\theta})$.

Concerning the parametrization choice, let $\theta:\mathcal{S}\times\mathcal{A}\rightarrow\mathbb{R}^{n}$.~\cite{bellemare2017distributional} suggests to fix $\{x_{i}\}_{i=1}^{n}$ and estimate $\mathbf{z} = \text{softmax}(\theta(s, a))$. This poses technical challenges, as the stochastic gradients of $W_{p}$ are biased. To tackle this issue,~\cite{dabney2018distributional} proposes fixing $z_{i} := n^{-1}$ and estimating $x_{i} = (\theta(s, a))_{i}$. This choice presents 2 advantages; (i) Discretizing $Z_{\theta}$'s support is more flexible as the support is now free, potentially leading to more accurate predictions; (ii) It allows the use of the Wasserstein distance as loss without suffering from biased gradients.

\subsection{Bayesian Reinforcement Learning}\label{sec:bayesian_rl}

Similarly to \gls{drl}, \gls{brl}~\cite{ghavamzadeh2015bayesian} adopts a distributional view reflecting the uncertainty over a given variable. In the remainder of this section, we discuss the \gls{wql} algorithm of~\cite{metelli2019propagating}, a modification of the Q-Learning algorithm of~\cite{watkins1992q} (eq.~\ref{eq:q-learning}). For a family of distributions $\mathcal{Q}$ (e.g., Gaussian distributions), this strategy considers a distribution $Q(s, a) \in \mathcal{Q}$, called $Q-$posterior, which represents the posterior distribution of the $Q-$function estimate. For each state there is a $V-$posterior $V(s)$, defined in terms of Wasserstein barycenters,
\begin{align}
    V(s) \in \arginf{V \in \mathcal{Q}}\expectation{a\sim \pi(\cdot|s)}[W_{2}(V,Q(s,a))^{2}].\label{eq:v-posterior}
\end{align}
This formulation shows an alternative, continuous view of Wasserstein barycenters, as one has infinitely many distributions $Q(\cdot,a)$ over states, for $a \sim \pi(\cdot|s)$.

Upon a transition $(s,a,s',r)$,~\cite{metelli2019propagating} defines the Wasserstein Temporal Difference, which is the distributional analogous to equation~\ref{eq:q-learning},
\begin{equation}
    Q_{t+1}(s,a)\in \mathcal{B}([1-\alpha_{t},\alpha_{t}];\{ Q_{t}(s,a),\mathcal{T}_{\pi}Q_{t}(s,a) \})\label{eq:q-update}
\end{equation}
where $\alpha_{t} > 0$ is the learning rate, whereas $\mathcal{T}_{\pi}Q_{t} = r + \lambda V_{t}$. For specific families $\mathcal{Q}$ (e.g., Gaussian distributions), eqs.~\ref{eq:v-posterior} and~\ref{eq:q-update} have an analytical solution (see~\cite[Table 1]{metelli2019propagating}).

\subsection{Policy Optimization}\label{sec:ppo}

\gls{po} focuses on searching for a policy $\pi$ that maximizes $\eta(\pi)$. In this section we focus on gradient-based approaches (e.g.,~\cite{sutton1999policy}), who parametrize $\pi = \pi_{\theta}$ so that training consists on maximizing $\eta(\theta) = \eta(\pi_{\theta})$. However, as~\cite{liu2017stein} remarks, these algorithms suffer from high variance and slow convergence. \cite{liu2017stein} thus propose a distributional view. Let $P$ be a distribution over $\theta$, they propose the following optimization problem,
\begin{align}
    P^{\star} &= \argmax{P}\expectation{\theta\sim P}[\eta(\pi_{\theta})] - \alpha \text{KL}(P\lVert P_{0}),\label{eq:liu_variational_rl}
\end{align}
where $P_{0}$ is a prior, and $\alpha > 0$ controls regularization. The minimum of equation~\ref{eq:liu_variational_rl} implies $P(\theta) \propto \text{exp}(\eta(\pi_{\theta})/\alpha)$. From a Bayesian perspective, $P(\theta)$ is the posterior distribution, whereas $\text{exp}(\eta(\pi_{\theta})/\alpha)$ is the likelihood function. As discussed in~\cite{zhang2018policy}, this formulation can be written in terms of gradient flows in the space of probability distributions (see~\cite{ambrosio2008gradient,santambrogio2017euclidean} for further details). First, consider the functional,
\begin{align*}
    F(P) &= -\int P(\theta)\log P_{0}(\theta) d\theta + \int P(\theta)\log P(\theta)d\theta.
\end{align*}
For Itó-diffusion~\cite{jordan1998variational}, optimizing $F$ w.r.t. $P$ becomes,
\begin{align}
    P_{k+1}^{(h)} = \argmin{P} \text{KL}(P\lVert P_{0}) + \dfrac{W_{2}^{2}(P,P_{k}^{(h)})}{2h},\label{eq:WgradFlow}
\end{align}
which corresponds to the \gls{jko} scheme~\cite{jordan1998variational}. In this context,~\cite{zhang2018policy} introduces 2 formulations for learning an optimal policy: indirect and direct learning. In the first case, one defines the gradient flow for equation~\ref{eq:WgradFlow}, in terms of $\theta$. This setting is particularly suited when the parameter space is low dimensional. In the second case, one uses policies directly into the described framework, which is related to~\cite{haarnoja2017reinforcement}.

\section{Final Remarks}\label{sec:conclusion}

\subsection{Challenges}\label{sec:challenges}

\begin{figure}[ht]
    \centering
    \includegraphics[width=\linewidth]{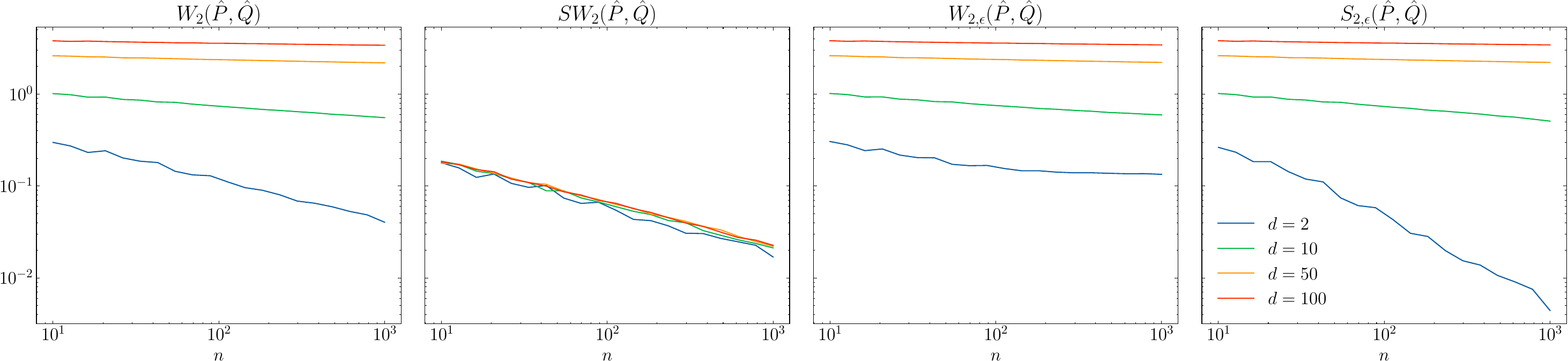}
    \caption{Estimation of Wasserstein distances with finite samples as a function of number of samples $n$, and dimensions $d$. Overall, the plug-in empirical estimator $W_{2}(\hat{P}, \hat{Q})$ suffers from the curse of dimensionality, as estimation becomes harder in high dimensions. This issue   can be alleviated through alternative estimators, such as sliced Wasserstein or entropic \gls{ot}.}
    \label{fig:sample_complexity}
\end{figure}

\noindent\textbf{Curse of Dimensionality.} A known theoretical fact is that \gls{ot} estimation becomes harder in high dimensions~\cite{dudley1969speed,weed2019sharp}, i.e., $|W_{p}(P,Q) - W_{p}(\hat{P},\hat{Q})| \leq \mathcal{O}(n^{-\nicefrac{1}{d}})$. We illustrate this issue by extending the experiments of~\cite{genevay2019sample} in Figure~\ref{fig:sample_complexity}. Especially, we compare $W_{2}$, $W_{2,\epsilon}$, $S_{2,\epsilon}$ and $SW_{2}$ for $P = Q = \mathcal{U}([0,1]^{d})$, i.e., the uniform distribution over the $d-$dimensional hypercube. Naturally, $W_{p}(P, Q) = 0$. We then sample $\mathbf{X}^{(P)} \sim P$ and $\mathbf{X}^{(Q)} \sim Q$ independently. We analyze these values as a function of the number of samples $n$, and the dimension of the data $d$. Note that, except for $SW_{2}$, the rate of estimation decays as we increase $d$. This challenge is key in several areas of \gls{ml}, as data commonly lives in a high-dimensional space.

\noindent\textbf{Computational Complexity.} When computing exact \gls{ot}, one has a $\mathcal{O}(n^{3}\log n)$ complexity over the amount of samples coming from the expensive linear program. This issue has been partially alleviated with the Sinkhorn algorithm of~\cite{cuturi2013sinkhorn}, which have $\mathcal{O}(n^{2})$ complexity \emph{per iteration}, and is amenable on GPU. Another direction consists of breaking probability distributions into $L$ $1-$D slices, which reduces the \gls{ot} problem to $L$ sorting problems, which has $\mathcal{O}(n \log n)$ complexity. Nonetheless, on one hand, for a large number of Sinkhorn iterations, exact \gls{ot} may be more efficient. The same remark can be made concerning the number of projections for the \gls{sw} distance. On the other hand, the Sinkhorn divergence is smoother than the exact Wasserstein distance~\cite{feydy2019interpolating}, which is desirable for optimization.

In table~\ref{tab:OT_complexity} we show a summary of \emph{time} and \emph{sample} complexities for \gls{ot} estimators. While the first concerns computational or time of execution, the second concerns the number of samples needed to accurately estimate the Wasserstein distance $W_{p}$.

\begin{table}[ht]
    \caption{Time and sample complexity of empirical \gls{ot} estimators in terms of the number of samples $n$. $\dagger$ indicates complexity of a single iteration; $^{*}$ indicates that the complexity is affected by samples dimension.}
    \centering
    \resizebox{\linewidth}{!}{
    \begin{tabular}{clll}
         \toprule
         Reference & Estimator & Time Complexity & Sample Complexity \\
         \midrule
         \cite{dudley1969speed} & $W_{p}(\hat{P},\hat{Q})$ & $\mathcal{O}(n^{3}\log n)$ & $\mathcal{O}(n^{-\nicefrac{1}{d}})$\\
         \cite{genevay2019sample} & $S_{p,\epsilon}(\hat{P},\hat{Q})$ & $\mathcal{O}(n^{2})^{\dagger}$ & $\mathcal{O}(n^{-1/2}(1+\epsilon^{\lfloor d/2 \rfloor}))$\\
         \cite{niles2019estimation} & SRW$_{k}(\hat{P},\hat{Q})$ & $\mathcal{O}(n^{2})^{\dagger,*}$ & $\mathcal{O}(n^{-\nicefrac{1}{k}})$\\
         \cite{nadjahi2019asymptotic} & $\text{SW}_{p}(\hat{P},\hat{Q})$ & $\mathcal{O}(n\log n)^{*}$ & $\mathcal{O}(n^{-\nicefrac{1}{2}})$\\
         \cite{forrow2019statistical} & $FW_{K,2}(\hat{P},\hat{Q})$ & $\mathcal{O}(n^{2})^{\dagger}$ & $\mathcal{O}(n^{-\nicefrac{1}{2}})$\\
         \cite{fatras2020learning} & $\mathcal{L}^{(k,m)}_{\text{MBOT}}(\hat{P},\hat{Q})$ & $\mathcal{O}(km^{3}\log m)$ & $\mathcal{O}(n^{-\nicefrac{1}{2}})$\\
         \bottomrule
    \end{tabular}}
    \label{tab:OT_complexity}
\end{table}

\subsection{Future Trends}

\noindent\textbf{ML for OT.} Several works~\cite{arjovsky2017wasserstein,seguy2018large,makkuva2020optimal,korotin2023neural} consider introducing \gls{ml} methodology in \gls{ot} theory, primarily through \glspl{nn}. The main advantage of doing so is scaling \gls{ot} methods to larger datasets and higher dimensions. However, without a careful analysis, the resulting methods may end up not estimating \gls{ot} (e.g.,~\cite{arjovsky2017wasserstein,korotin2021do}), even if the method successfully performs the \gls{ml} task, as in the case of generative modeling~\cite{stanczuk2021wasserstein}, which suggests an apparent separation between \gls{ot} estimation and \gls{ml} tasks. However, recent approaches~\cite{makkuva2020optimal,korotin2023neural} pose \gls{ot} estimation as a learning task. This trend shows an interplay between \gls{ot} and \gls{ml} practice.

\noindent\textbf{Sliced Wasserstein distances.} The \gls{sw} distance has been generalized by a number of works~\cite{kolouri2019generalized,nguyen2022revisiting,bonet2022spherical}. For instance, while~\cite{kolouri2019generalized} proposes a non-linear slicing method through \glspl{nn},~\cite{nguyen2022revisiting} slices distributions through convolutions. Furthermore,~\cite{bonet2022spherical} define the \gls{sw} distance over the sphere $\mathbb{S}^{d-1}$. Further research can focus on the idea of \gls{sw} distances over manifolds.

\noindent\textbf{Decentralized and Private OT.} In recent \gls{ml} practice, researchers considered distributed learning over several devices or clients without directly communicating data~\cite{mcmahan2017communication}, known as \emph{federated learning}. In this context, \emph{differential privacy} a strong guarantee for protecting clients' data~\cite{dwork2008differential}. Recent advances in \gls{ot} devise ways to privately~\cite{le2019differentially,cao2021don} and federated~\cite{rakotomamonjy2023federated} computing the Wasserstein distance, as well as distributed \gls{msda} strategies~\cite{castellon2023federated}. These works serve as a starting point for \gls{ot}-inspired federated strategies in other fields of \gls{ml}.

\noindent\textbf{Supervised Learning.} As discussed in section~\ref{sec:supervised-learning}, \gls{ot} contributes in two ways: defining a loss that considers semantic similarities between classes and defining robust alternatives to \gls{erm}. Future works could consider the construction of label embeddings (vector representations) for classes so that these embeddings capture the geometry of their corresponding distributions, i.e., $P(X|Y=y)$.

\noindent\textbf{Generative Modeling.} This subject has been one of the most active \gls{otml} topics. This phenomenon is evidenced by the impact of papers such as~\cite{arjovsky2017wasserstein} and~\cite{gulrajani2017improved} had on how generative models are understood and trained~\cite{bassetti2006minimum}. Future works can focus on using extensions to \gls{ot} as a loss. For instance, one can devise outlier-robust generative models through unbalanced or partial \gls{ot}. Furthermore, one can devise generative graph models with the \gls{fgw} distance.

\noindent\textbf{Dictionary Learning.} \gls{ot} provides a rich theory for dictionary learning when objects have an underlying probabilistic interpretation, such as histograms. In this sense, one can perform dictionary learning in Wasserstein space by using the Wasserstein distance as a loss function and using Wasserstein barycenters for combining atoms. This idea is recurrent in \gls{ml}, e.g., images and text~\cite{schmitz2018wasserstein}, graphs~\cite{vincent2021online,xu2020gromov} and domain adaptation~\cite{montesuma2023learning}. Future contributions can focus on the manifold of interpolations of atoms in the Wasserstein space and the interpretability of coefficients $\mathbf{a}_{i}$.

\noindent\textbf{Clustering.} As shown in section~\ref{sec:clustering}, \gls{ot} provides a probabilistic framework for clustering algorithms (e.g., k-Means~\cite{cuturi2014fast}). Furthermore, \gls{ot} contributed to two clustering settings: co-clustering and \glspl{gmm}. In the first case,~\cite{laclau2017coclustering} relies on the \gls{gw} formalism for finding clusters over samples and features. In the second case,~\cite{kolouri2018sliced} uses the \gls{sw} distance for learning \glspl{gmm}. This metric works well in high dimensions, is easy to compute, and is robust to initialization. Future works can consider the theoretical analysis of the approach of~\cite{kolouri2018sliced}, such as proving convergence and robustness to initialization.

\noindent\textbf{Transfer Learning.} From a practical perspective, \gls{ot}-based \gls{uda} has shown state-of-the-art performance in various fields, such as image classification~\cite{courty2017otda}, sentiment analysis~\cite{courty2017joint}, fault diagnosis~\cite{cheng2019wasserstein,montesuma2022cross}, and audio classification~\cite{montesuma2021wbt,turrisi2022multi}. Advances in the field include methods and architectures that can handle large-scale datasets, such as \gls{wdgrl}~\cite{shen2018wasserstein} and Deep\gls{jdot}~\cite{damodaran2018deepjdot}. From a theoretical perspective, various works~\cite{redko2017theoretical,redko2019advances} show that \gls{ot} is an essential framework for \emph{formalizing \gls{uda}}. This effectively consolidated \gls{ot} as a valuable toolbox for transferring knowledge between different domains.

One may draw parallels between UDA, fairness, and generative modeling in a broader context. In the first case, the random repair strategy of~\cite{gordaliza2019obtaining} is conceptually similar to \gls{otda}~\cite{courty2017otda}. This similarity suggests that theoretical results on the optimal amount of repaired data can be derived, similarly to~\cite[Theorem 3]{redko2017theoretical}. Furthermore, parallel between (DANN, WDGRL) and (GAN, WGAN), which shows an interesting interplay between the two fields.

\noindent\textbf{Reinforcement Learning.} Reinforcement Learning. The distributional \gls{rl} framework of~\cite{bellemare2017distributional} re-introduced previous ideas in modern \gls{rl}, such as formulating the goal of an agent in terms of distributions over rewards. This idea turned out to be successful, as the authors' method surpassed the state-of-the-art in a variety of game benchmarks. Furthermore,~\cite{dabney2020distributional} studies how this framework relates to dopamine-based \gls{rl}, highlighting this formulation's importance.

In a similar spirit,~\cite{metelli2019propagating} proposes a distributional method for propagating uncertainty of the value function. Here, it is important to highlight that while~\cite{bellemare2017distributional} focuses on the randomness of the reward function,~\cite{metelli2019propagating} studies the stochasticity of value functions. In this sense, the authors propose a variant of Q-Learning, called \gls{wql}, that leverages Wasserstein barycenters for updating the $Q-$function distribution. From a practical perspective, the \gls{wql} algorithm is theoretically grounded and has good performance. Further research can focus on the common points between~\cite{bellemare2017distributional},~\cite{dabney2018distributional}, and~\cite{metelli2019propagating}, either by establishing a common framework or by comparing the 2 methodologies empirically. Finally,~\cite{zhang2018policy} explores policy optimization as gradient flows in the space of distributions. This formalism can be understood as the continuous-time counterpart of gradient descent.

\subsection{Conclusion}

Optimal transport serves as a formal toolkit for probabilistic machine learning. In this survey, we cover applications to supervised, unsupervised, transfer, and reinforcement learning settings, as well as \emph{how to compute it}. In a nutshell, it is used for computing losses or manipulating probability distributions. On the one hand, the main attractive points of this theory are its flexibility and useful theoretical properties. On the other hand, two challenges remain relevant in the context of machine learning: \gls{ot} is difficult to estimate in high dimensions, and it is computationally heavy. Finally, \gls{ot} has positively impacted the probabilistic machine learning landscape. Conversely, solutions coming from machine learning begin to influence \gls{ot}, such as using neural networks to approximate \gls{ot} maps and plans.

\newpage

\ifCLASSOPTIONcaptionsoff
  \newpage
\fi

\bibliographystyle{IEEEtran}
\bibliography{citations.bib}

\end{document}